\def\year{2022}\relax
\documentclass[letterpaper]{article} 
\usepackage{aaai22}  
\usepackage{times}  
\usepackage{helvet}  
\usepackage{courier}  
\usepackage[hyphens]{url}  
\usepackage{graphicx} 
\usepackage{natbib}  
\usepackage{caption} 
\DeclareCaptionStyle{ruled}{labelfont=normalfont,labelsep=colon,strut=off} 
\usepackage{algorithm}
\usepackage{algorithmic}
\usepackage{booktabs}
\usepackage{enumerate}
\usepackage{multirow}
\usepackage{threeparttable}
\usepackage{caption}
\usepackage{epsfig}
\usepackage{amsmath}
\usepackage{amssymb}
\usepackage[switch]{lineno}

\usepackage{color}

\usepackage[hang,flushmargin]{footmisc} 

\usepackage[breaklinks=true,bookmarks=false]{hyperref}

\usepackage{newfloat}
\usepackage{listings}
\floatstyle{ruled}
\newfloat{listing}{tb}{lst}{}
\floatname{listing}{Listing}
\title{Texture Reformer: Towards Fast and Universal Interactive Texture Transfer}

\author{
    Zhizhong Wang, \hspace{0.4cm}Lei Zhao$^*$, \hspace{0.4cm} Haibo Chen, \hspace{0.4cm} Ailin Li, \\
    Zhiwen Zuo, \hspace{0.4cm} Wei Xing$^*$, \hspace{0.4cm}Dongming Lu
}
\affiliations{
    College of Computer Science and Technology, Zhejiang University\\
    \{endywon, cszhl, cshbchen, liailin, zzwcs, wxing, ldm\}@zju.edu.cn

}

\begin{document}
\frenchspacing

\twocolumn[{%
	\renewcommand\twocolumn[1][]{#1}%
	\maketitle
	\vspace{-1cm}
	\begin{center}
		\centering
		\includegraphics[width=1\linewidth]{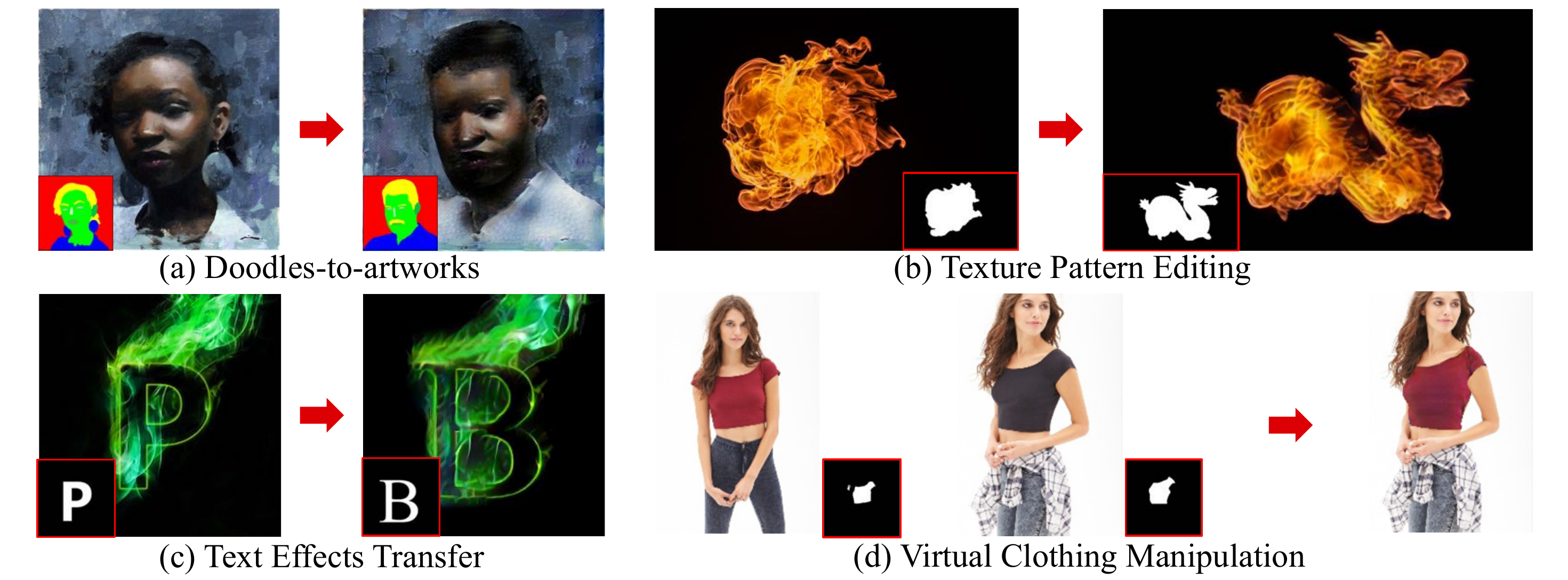}
		\vspace{-1.7em}
		\captionof{figure}{Representative results generated by our interactive texture reformer. The stylized images are synthesized under the guidance of corresponding user-specified semantic maps. Our framework is universal for multiple challenging user-controlled texture transfer tasks, {\em e.g.}, (a) turning doodles into artworks, (b) editing texture patterns, (c) transferring text effects, (d) manipulating clothing textures and distributions. Compared with the state-of-the-art interactive texture transfer algorithms, {\em it not only can achieve higher quality results but, more remarkably, also is 2-5 orders of magnitude faster.}
		}
		\label{fig:teaser}
		
	\end{center} 
}]

\renewcommand{\thefootnote}{}
\footnote{*Corresponding authors.}
\footnote{Copyright \copyright~2022, Association for the Advancement of Artificial Intelligence (www.aaai.org). All rights reserved.}

\renewcommand{\thefootnote}{1}
\begin{abstract}
In this paper, we present the {\em texture reformer}, a fast and universal neural-based framework for interactive texture transfer with user-specified guidance. The challenges lie in three aspects: 1) the diversity of tasks, 2) the simplicity of guidance maps, and 3) the execution efficiency. To address these challenges, our key idea is to use a novel feed-forward multi-view and multi-stage synthesis procedure consisting of I) a global view structure alignment stage, II) a local view texture refinement stage, and III) a holistic effect enhancement stage to synthesize high-quality results with coherent structures and fine texture details in a coarse-to-fine fashion. In addition, we also introduce a novel learning-free {\em view-specific texture reformation (VSTR)} operation with a new semantic map guidance strategy to achieve more accurate semantic-guided and structure-preserved texture transfer. The experimental results on a variety of application scenarios demonstrate the effectiveness and superiority of our framework. And compared with the state-of-the-art interactive texture transfer algorithms, it not only achieves higher quality results but, more remarkably, also is 2-5 orders of magnitude faster. Code is available at \url{https://github.com/EndyWon/Texture-Reformer}.
\end{abstract}

\section{Introduction}
\label{introduction}
As a variant of texture synthesis, texture transfer is a long-standing problem that seeks to transfer the stylized texture from a given sample to the target image \cite{efros2001image}. After the rapid development in recent years, a bunch of conventional \cite{hertzmann2001image} or neural-based \cite{gatys2016image} methods have been proposed and obtained visually appealing results. However, due to the lack of user guidance, general texture transfer methods often produce unsatisfying results against human expectations. To resolve this dilemma, the community resorts to using the user-specified semantic maps to guide the transfer process, which is called {\em interactive texture transfer} \cite{men2018common}. Users can control the shape, scale, and spatial distribution of the objects to be synthesized in the target image via semantic maps.

At first, the interactive texture transfer methods are only designed for specific usage scenarios. \cite{champandard2016semantic} proposed Neural Doodle to turn doodles painted by users into fine artworks with provided samples. Lu~{\em et al.} designed HelpingHand \cite{lu2012helpinghand}, RealBrush \cite{lu2013realbrush}, and DecoBrush \cite{lu2014decobrush} to edit different kinds of texture patterns. \cite{yang2017awesome,yang2019controllable} achieved text effects transfer that can migrate various fantastic text effects of stylized texts onto raw plain texts. \cite{han2018viton} introduced an image-based virtual try-on network to transfer a target clothing item in a product image to the corresponding region of a clothed person. These approaches seem to be isolated, but they all share a common notion of transferring textures under user guidance. 

To unify them, \cite{men2018common} proposed a common framework for interactive texture transfer by incorporating multiple custom channels to dynamically guide the synthesis. This method is capable of handling various challenging tasks and achieves the state of the art. However, it relies on several CPU-based operations and a backward optimization process, thus usually requiring several minutes to generate a result for each interaction, which is prohibitively slow. Therefore, existing algorithms are hard to satisfy the practical requirements due to the limitations of efficiency or application scenarios. A fast and universal framework is eagerly desired, and it will undoubtedly improve the user experience and bring higher application and research value to both industry and academia.

However, achieving such a goal is a rather challenging task. The challenges mainly lie in three aspects: 1) The diversity of tasks: the discrepancies between different tasks make the transfer problem difficult to model uniformly. Besides, for each task, the algorithm should be robust to different input samples. 2) The simplicity of guidance maps: the doodle semantic map as guidance gives few hints on how to place different inner textures and preserve local high-frequency structures \cite{men2018common}. 3) The execution efficiency: the trade-off between efficiency and quality is always an intractable problem. This is particularly important for interactive systems since the insufficient computational speed not only brings inconvenience to users but also hampers the truly exploratory use of these techniques.

To address these challenges, in this paper, we propose the {\em texture reformer}, a fast and universal neural-based framework for interactive texture transfer with user-specified guidance. The key insight is to use a novel feed-forward multi-view and multi-stage synthesis procedure, which consists of three different stages: I) a global view structure alignment stage, II) a local view texture refinement stage, and III) a holistic effect enhancement stage. Specifically,~for stage I and II, we introduce a novel {\em View-Specific Texture Reformation (VSTR)} operation with a new semantic map guidance strategy to achieve more accurate semantic-guided and structure-preserved texture transfer. By specifying a global view for VSTR, our stage I first captures and aligns the inner structures of the source textures as completely as possible. Then, the results are carefully rectified and refined in stage II via specifying a local view for VSTR. Finally, in stage III, we leverage the Statistics-based Enhancement (SE) operations to further enhance the low-level holistic effects ({\em e.g.}, colors, brightness, and contrast). Note that our framework is built upon several auto-encoder networks trained solely for image reconstruction, and the VSTR and SE operations are {\em learning-free}. Therefore, it can achieve interactive texture transfer universally. By cascading the above three stages, our texture reformer can synthesize high-quality results with coherent structures and fine texture details in a coarse-to-fine fashion. We demonstrate the effectiveness and superiority of our framework on a variety of application scenarios, including doodles-to-artworks, texture pattern editing, text effects transfer, and virtual clothing manipulation (see Fig.~\ref{fig:teaser}). The experimental results show that compared with the state-of-the-art algorithms, our texture reformer not only achieves higher quality results but, more remarkably, also is 2-5 orders of magnitude faster. {\em As far as we know, our work is the first to meet the requirements of quality, flexibility, and efficiency at the same time in this task.} 

In summary, our contributions are threefold:
\begin{itemize}
	\setlength{\itemsep}{2pt}
	\setlength{\parsep}{0pt}
	\setlength{\partopsep}{0pt}
	\setlength{\parskip}{0pt}
	\item We propose a novel multi-view and multi-stage neural-based framework, {\em i.e.}, {\em texture reformer}, to achieve fast and universal interactive texture transfer for the first time.
	
	\item We also introduce a novel learning-free {\em view-specific texture reformation (VSTR)} operation with a new semantic map guidance strategy, to realize more accurate semantic-guided and structure-preserved texture transfer.
	
	\item We apply our framework to many challenging interactive texture transfer tasks, and demonstrate its effectiveness and superiority through extensive comparisons with the state-of-the-art (SOTA) algorithms.
\end{itemize}

\vspace{-0.5em}
\section{Related Work}
{\bf Conventional Texture Transfer.} Conventional texture transfer relies on hand-crafted algorithms \cite{haeberli1990paint} or features \cite{kwatra2005texture} to migrate the textures from source samples to target images. The pioneering works of \cite{efros1999texture,efros2001image} sampled similar patches to synthesize and transfer textures. Later, \cite{hertzmann2001image} proposed Image Analogy to generate the stylized result of the target image. \cite{barnes2009patchmatch,barnes2010generalized} proposed PatchMatch to accelerate the nearest-neighbor search process, which was further extended to image melding \cite{darabi2012image}, style transfer \cite{frigo2016split}, and text effects transfer \cite{yang2017awesome},~{\em etc}. However, for interactive texture transfer, these methods fail to synthesize the textures with salient structures and are prone to wash-out effects \cite{men2018common}. To combat the issues, \cite{men2018common} proposed a common framework for interactive texture transfer by utilizing an improved PatchMatch and multiple custom channels to dynamically guide the synthesis, achieving SOTA performance. However, as analyzed in Sec.~\ref{introduction}, this method suffers from rather slow computational speed, thus cannot satisfy the practical requirements.

Unlike the SOTA conventional texture transfer methods \cite{yang2017awesome,men2018common}, our proposed texture reformer is neural-based, and not only can achieve higher quality results but also is several orders of magnitude faster.

{\bf Neural-based Style Transfer.} The seminal works of \cite{gatys2016image,gatys2015texture} have proved the power of Deep Convolutional Neural Networks (DCNNs)~\cite{simonyan2014very} in style transfer and texture synthesis, where the Gram matrices of the features extracted from different layers of DCNNs are used to represent the style of images. Further works improved it in many aspects, including efficiency~\cite{johnson2016perceptual,ulyanov2016texture}, quality~\cite{sheng2018avatar,jing2018stroke,gu2018arbitrary,kolkin2019style,park2019arbitrary,wang2020glstylenet,wang2021evaluate,chen2020creative,chen2021dualast,chen2021artistic,lin2021drafting,cheng2021style,an2021artflow}, generality~\cite{li2017universal,huang2017arbitrary,lu2019closed,zhang2019metastyle,jing2020dynamic}, and diversity~\cite{wang2020diversified,chen2021diverse}. For interactive style transfer, \cite{gatys2017controlling} introduced user spatial control into~\cite{gatys2016image}, which is further accelerated by \cite{lu2017decoder}. However, due to the characteristics of Gram matrix matching, these methods often produce disordered textures, which cannot preserve the local inner structures, as will be demonstrated in later Sec.~\ref{cmp}.

Another line of neural-based style transfer is based on neural patches. \cite{li2016combining,li2016precomputed} first achieved it by combining Markov Random Fields (MRFs) and DCNNs. \cite{liao2017visual} proposed Deep Image Analogy for more accurate semantic-level patch matching. Later, \cite{chen2016fast} leveraged a ``style swap" operation for fast patch-based stylization. To incorporate user control, \cite{champandard2016semantic} augmented \cite{li2016combining} with semantic annotations, leading to higher quality and avoiding common glitches. However, the results usually contain too many low-level noises. Also, efficiency is concerned as it still relies on a time-consuming backward optimization process.

Generative Adversarial Networks (GANs) \cite{goodfellow2014generative} provide another idea to generate textures by training discriminator and generator networks to play an adversarial game. cGANs \cite{mirza2014conditional}, which were further extended to image-to-image translation \cite{isola2017image}, have been applied to many image manipulation tasks, such as interactive image editing \cite{zhu2016generative}, texture synthesis \cite{fruhstuck2019tilegan}, sketch2image \cite{chen2018sketchygan}, and inpainting \cite{zhao2020uctgan}, {\em etc}. However, all of them are trained on class-specific datasets. By contrast, our method only needs one exemplar for generating the target image from a corresponding semantic map. Recently, some single image generative models \cite{shaham2019singan,lin2020tuigan} were also proposed to generate images based on only a single image. Nevertheless, these methods often produce poor results for images with complex texture details and structures, and need several hours to train the model on each pair of image samples.

\begin{figure*}[t]
	\centering
	\includegraphics[width=0.9\linewidth]{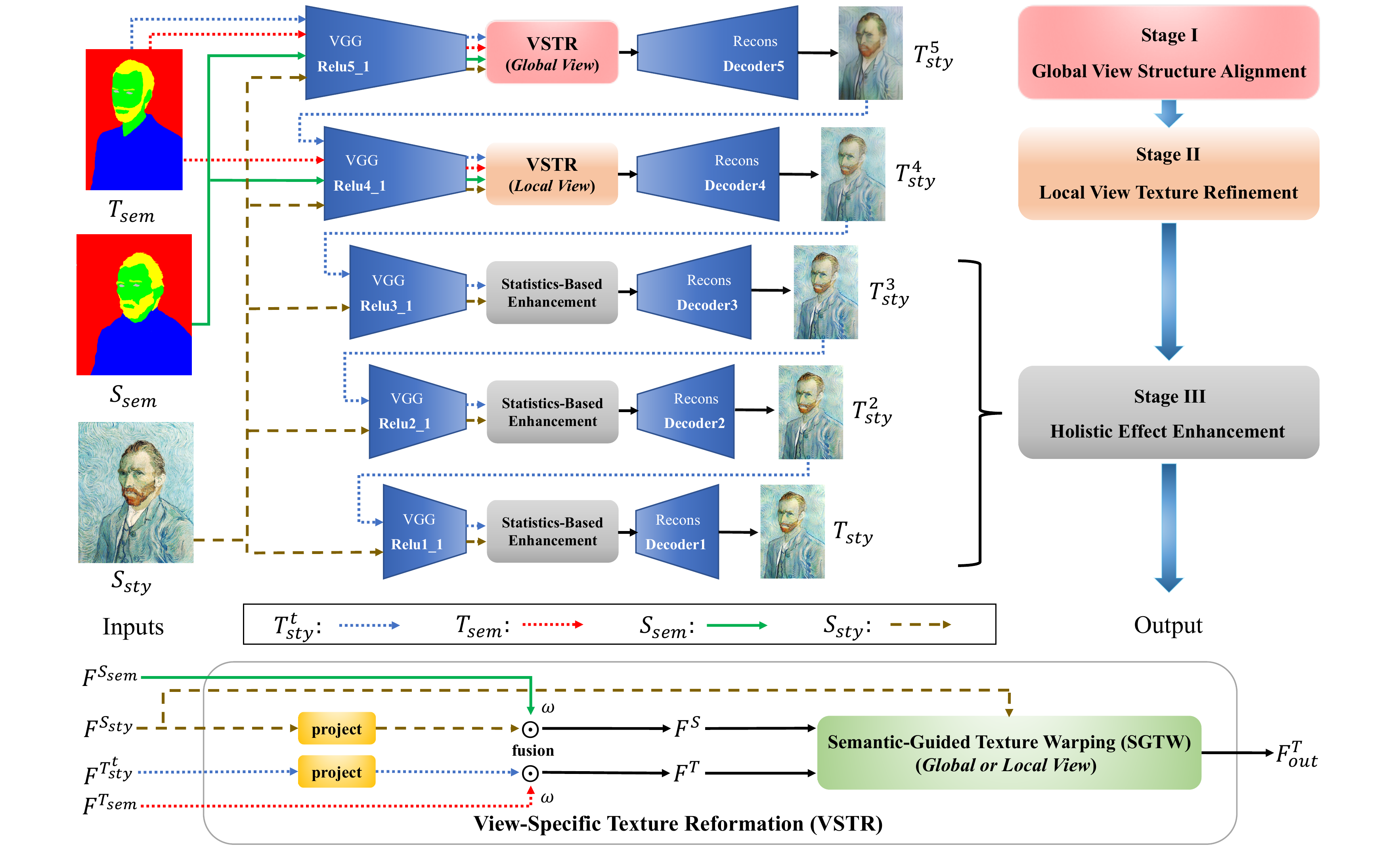}
	\vspace{-0.5em}
	\caption{Overall pipeline of our proposed multi-view and multi-stage texture reformer.
	}
	\label{fig:over}
	\vspace{-0.5em}
\end{figure*}

\renewcommand\arraystretch{0.6}
\begin{figure*}[t!]
	\centering
	\setlength{\tabcolsep}{0.1cm}
	\begin{tabular}{cccccccp{0.15cm}|p{0.15cm}cccc}
		\includegraphics[width=0.1\linewidth]{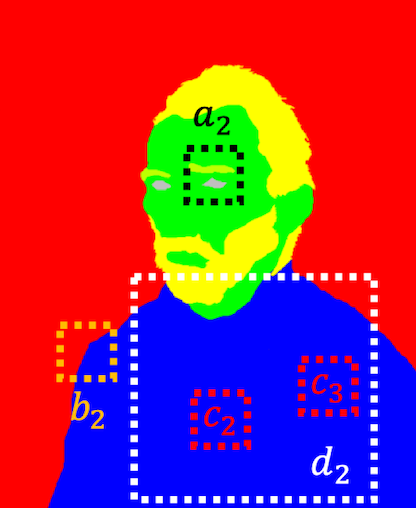}&
		\multirow{2}{*}[0.5in]{\bf :}&
		\includegraphics[width=0.1\linewidth]{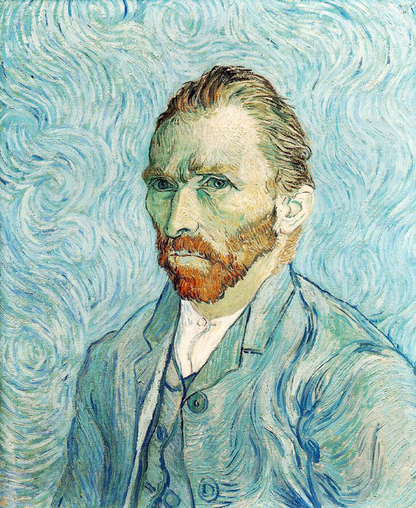}&
		\multirow{2}{*}[0.5in]{\bf ::}&
		\includegraphics[width=0.0835\linewidth]{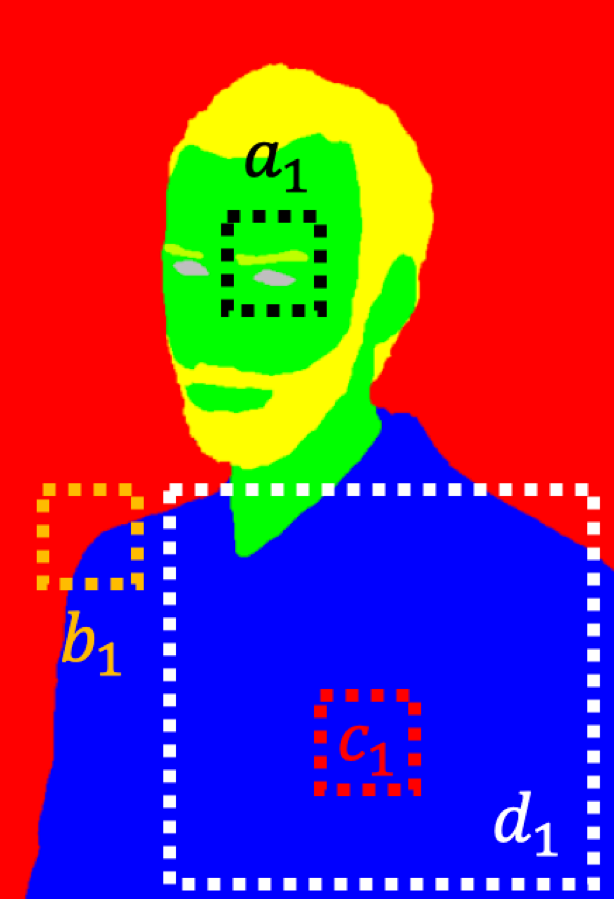}&
		\multirow{2}{*}[0.5in]{\bf :}&
		\includegraphics[width=0.0835\linewidth]{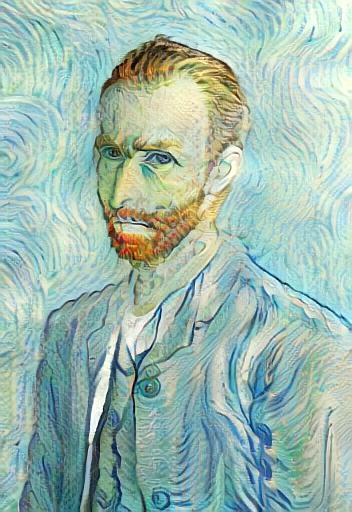}&
		&&
		
		\includegraphics[width=0.0835\linewidth]{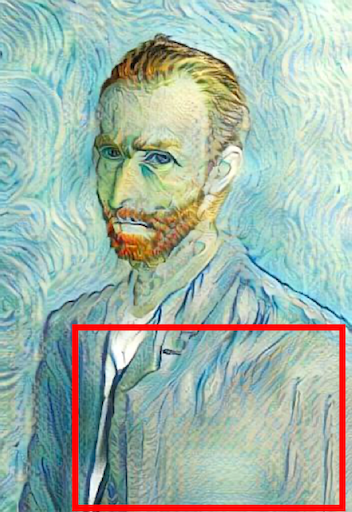}&
		\includegraphics[width=0.0835\linewidth]{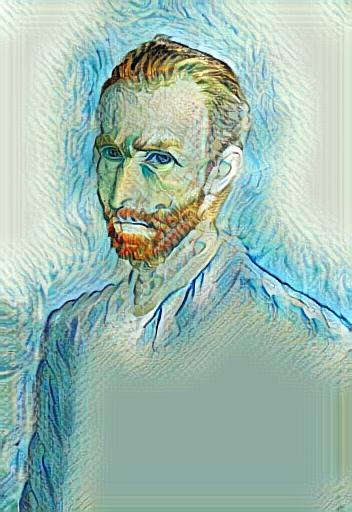}&
		\includegraphics[width=0.0835\linewidth]{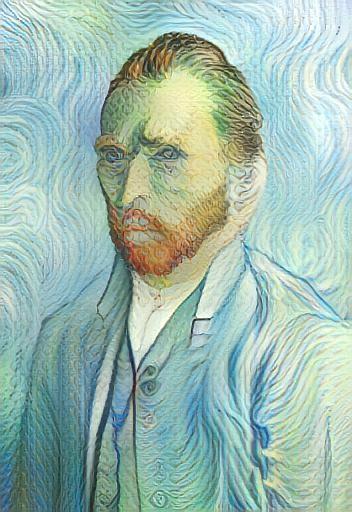}&
		\includegraphics[width=0.0835\linewidth]{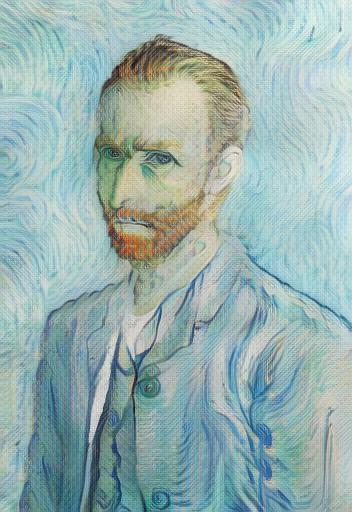}
		\\
		
		$S_{sem}$ \scriptsize (input) && $S_{sty}$ \scriptsize (input) && $T_{sem}$ \scriptsize (input) && $T_{sty}$ \scriptsize (output)&
		&&
		\scriptsize (a) w/o global view & \scriptsize (b) w/o stage I & \scriptsize (c) w/o stage II & \scriptsize (d) w/o stage III
		\\
		\specialrule{0em}{1pt}{1pt}
		\hline
		\specialrule{0em}{1pt}{1pt}
		\multicolumn{6}{c|}{\scriptsize Time/sec (averaged on 100 pairs of 512$\times$512px images):} & \scriptsize 0.956 
		&&& \scriptsize 0.912 & \scriptsize 0.581& \scriptsize 0.569& \scriptsize 0.653
		
	\end{tabular}
    \vspace{-0.5em}
	\caption{ {\bf Left:} Illustration of the interactive texture transfer task. Input three images: $S_{sem}$ (semantic map of source image), $S_{sty}$ (stylized source image aligned to $S_{sem}$), and $T_{sem}$ (user-specified semantic map of target image), the stylized target image $T_{sty}$ with the style of source image $S_{sty}$ can be automatically synthesized such that $S_{sem}:S_{sty}::T_{sem}:T_{sty}$. {\bf Right:} Effects of different critical components and stages in our texture reformer (Fig.~\ref{fig:over}). {\bf Bottom:} Efficiency comparison.
	}
	\vspace{-1em}
	\label{fig:problem}
\end{figure*}

\vspace{-0.5em}
\section{Proposed Approach}
\label{approach}
We first describe the task of interactive texture transfer following the definitions in~\cite{men2018common}. As illustrated in the left part of Fig.~\ref{fig:problem}, given a stylized source image $S_{sty}$ and its corresponding semantic map $S_{sem}$, interactive texture transfer aims to generate the stylized target image $T_{sty}$ with a user-specified target semantic map $T_{sem}$. Users can control the shape, scale, and spatial distribution of the objects to be synthesized in the target image via semantic maps.

Using a semantic map that contains few hints to reproduce the structural image is a challenging task. The key challenge is to preserve the structures of the inner textures of each semantic region, {\em e.g.}, the clothing structures in the blue region of Fig.~\ref{fig:problem}. \cite{men2018common} combat it by introducing structure guidance based on the boundary patches of semantic maps to provide a prior in the synthesis procedure. However, it involves several structure extraction~\cite{goferman2011context} and propagation~\cite{myronenko2010point,bookstein1989principal} processes, which are cumbersome and time-consuming. In a fundamentally different way, we do not use any additional structure guidance but only benefit from the strong representative power of DCNNs to extract the multi-level image features. Based on these features, our key insight is to use a multi-view and multi-stage synthesis procedure to progressively generate structural textures in a coarse-to-fine fashion. In the following sections, we will first depict the overall pipeline and some critical components of our framework (Sec.~\ref{over}), and then introduce each of its stages in detail (Sec.~\ref{stage1}-\ref{stage3}).

\vspace{-0.65em}
\subsection{Overview of Texture Reformer}
\label{over}
The overall pipeline of our framework is depicted in Fig.~\ref{fig:over}, which consists of three stages: I) a global view structure alignment stage, II) a local view texture refinement stage, and III) a holistic effect enhancement stage. Specifically, stage I is similar to a global copy-and-paste, which roughly aligns the spatial positions of the source patterns in $S_{sty}$ to the target positions in the target semantic map $T_{sem}$. This global view alignment can help preserve the inner structures of the source patterns as completely as possible, which is critical to synthesize the structure-preserved textures. The warping and finer alignment is achieved via stage II, which uses a rather small local view, and can rectify and refine the results of stage I to a large extent, thus robust for different deformation requirements. Finally, the low-level holistic effects ({\em e.g.}, colors, brightness, and contrast) are further enhanced in stage III, thereby obtaining high-quality results. The visualizations of the inputs/outputs of each stage are shown in Fig.~\ref{fig:over} ($T_{sty}^5$-$T_{sty}$). These stages are carried out at different levels of VGG~\cite{simonyan2014very} features and are hierarchically cascaded to work in a coarse-to-fine fashion. Uniformly, they share the same workflow that generates the outputs using an AE (auto-encoder)-based image reconstruction process coupled with bottleneck feature operations. We adopt view-specific texture reformation (VSTR) for stage I and II, and statistics-based enhancement (SE) for stage III, which we will introduce in detail.

{\bf AE-based Image Reconstruction.} We construct auto-encoder networks for general image reconstruction. We employ the first parts (up to $Relu{\bf X}\_1$) of a pre-trained VGG-19~\cite{simonyan2014very} as encoders, {\em fix} them and train symmetrical decoder networks with the nearest neighbor interpolation as upsampling layers for inverting the bottleneck features to the original RGB images. As shown in Fig.~\ref{fig:over}, in our framework, we select feature maps at five layers, {\em i.e.}, $Relu{\bf X}\_1$ ({\bf X}=1,2,3,4,5), and train five decoders accordingly with the following loss:
\begin{equation}
\mathcal{L}_{recon}=\parallel I_r-I_i \parallel^2_2+\lambda \parallel \Phi(I_r) - \Phi(I_i) \parallel^2_2
\label{loss}
\end{equation}
where $I_i$ and $I_r$ are the input image and reconstructed output, and $\Phi$ is the VGG encoder that extracts the $Relu{\bf X}\_1$ features. The decoders are trained on the Microsoft COCO dataset~\cite{lin2014microsoft} and $\lambda$ is set to 1. 

{\bf View-Specific Texture Reformation (VSTR).} We propose a novel {\em learning-free} VSTR operation to robustly propagate the thorough texture patterns onto the target features under the guidance of semantic maps. Denote $F^{S_{sty}}$ and $F^{T_{sty}^t}$ as the VGG features ({\em e.g.}, extracted from $Relu5\_1$) of the stylized source image $S_{sty}$ and the {\em temporary} stylized target image $T_{sty}^t$. We first project them into a common space to standardize the data and dispel the domain gap,
\begin{equation}
F^{S_{sty}}_1=\frac{F^{S_{sty}}-\mu(F^{S_{sty}})}{\sigma(F^{S_{sty}})};\ F^{T_{sty}^t}_1=\frac{F^{T_{sty}^t}-\mu(F^{T_{sty}^t})}{\sigma(F^{T_{sty}^t})},
\end{equation}
where $\mu$ and $\sigma$ are the mean and standard deviation.

\begin{figure}[t]
	\centering
	\setlength{\tabcolsep}{0.03cm}
	\begin{tabular}{ccccc}
		\includegraphics[width=0.195\linewidth]{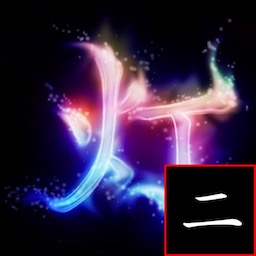}&
		\includegraphics[width=0.195\linewidth]{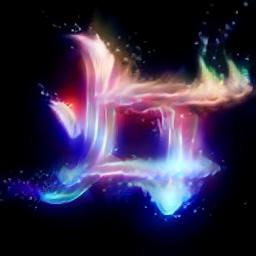}&
		\includegraphics[width=0.195\linewidth]{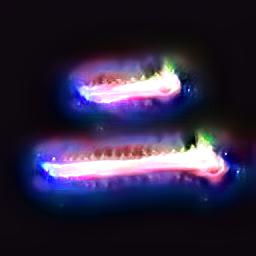}&
		\includegraphics[width=0.195\linewidth]{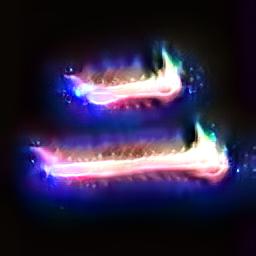}&
		\includegraphics[width=0.195\linewidth]{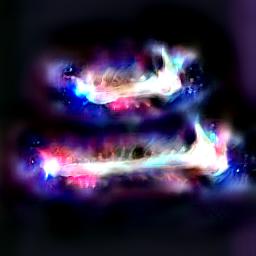}
		\\
		
		\includegraphics[width=0.195\linewidth]{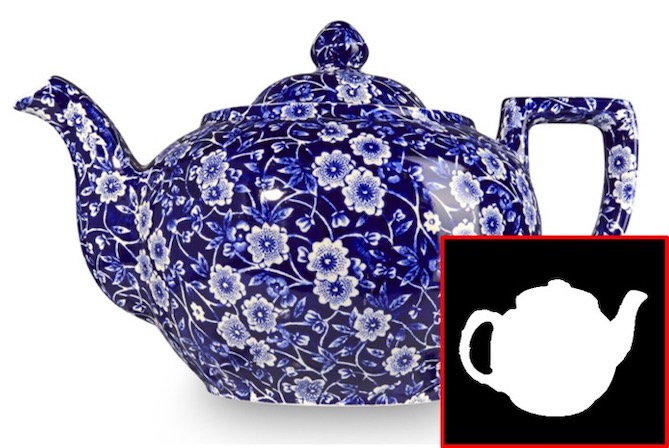}&
		\includegraphics[width=0.195\linewidth]{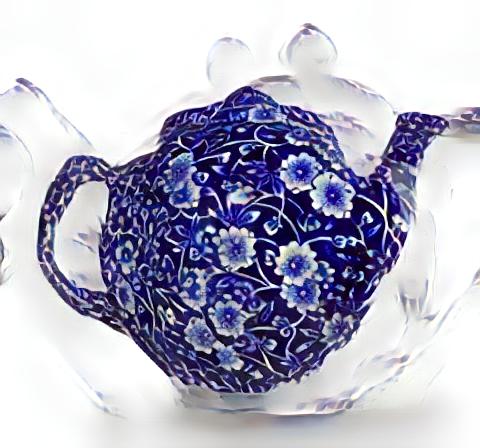}&
		\includegraphics[width=0.195\linewidth]{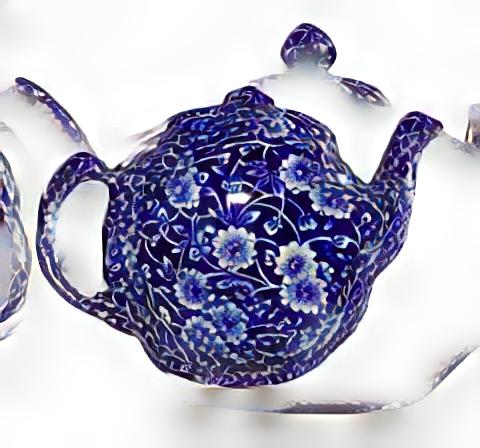}&
		\includegraphics[width=0.195\linewidth]{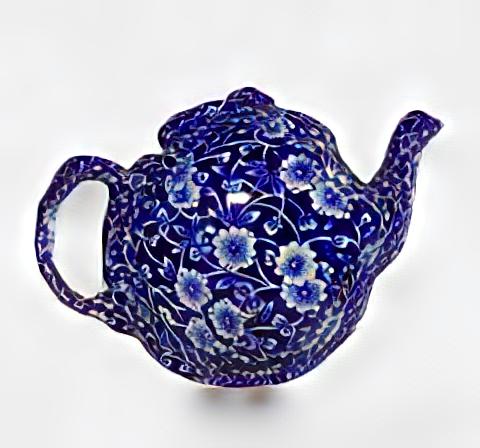}&
		\includegraphics[width=0.195\linewidth]{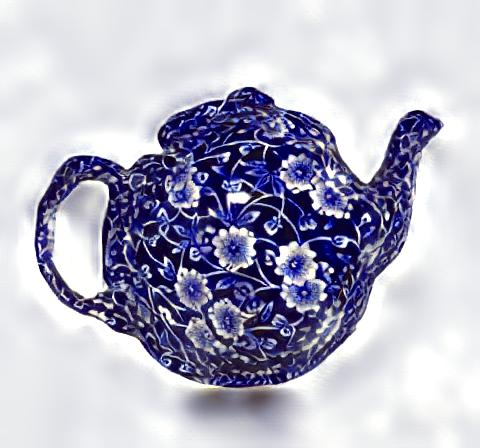}
		\\
		
		\scriptsize ($S_{sty}$, $T_{sem}$)&\scriptsize downsample &\scriptsize add (ours)&\scriptsize concat (ours) &\scriptsize covariance (SE)
		\\
		\specialrule{0em}{1pt}{1pt}
		\hline
		\specialrule{0em}{1pt}{1pt}
		\scriptsize Time/sec: &\scriptsize 0.682 &\scriptsize 0.897 &\scriptsize 0.956 &\scriptsize 1.732
		
	\end{tabular}
	\vspace{-0.5em}
	\caption{Comparison of different semantic map guidance strategies ($2^{nd}$ to $4^{th}$ columns) and different enhancement operations (last column).
	}
	\vspace{-1.5em} 
	\label{fig:ablation}
\end{figure}

Then, we fuse the information from the source and target semantic maps $S_{sem}$ and $T_{sem}$ to guide the propagations between corresponding semantic regions. Existing works~\cite{champandard2016semantic,gatys2017controlling} often directly concatenate the downsampled semantic maps, like follows:
\begin{equation}
F^S=F^{S_{sty}}_1\parallel \omega S_{sem}^l;\  F^T=F^{T_{sty}^t}_1\parallel \omega T_{sem}^l,
\end{equation}
where $\parallel$ denotes channel-wise concatenation. $l$ denotes the downsampling factor. $\omega$ is the hyperparameter that controls the weight of semantic awareness. However, as pointed out by~\cite{gatys2017controlling}, this method has limited capacity to model complex textures and usually produces inaccurate semantic matching ({\em e.g.}, the $2^{nd}$ column in Fig.~\ref{fig:ablation}). This can be attributed to the information discrepancy between the RGB images and deep VGG features. In addition, the difference in the amount of channels may also make it hard to find a good compromise ({\em i.e.}, the proper value of $\omega$) between them.

To resolve this issue, we introduce a new semantic map guidance strategy in our VSTR. That is, first extracting the VGG features $F^{S_{sem}}$ and $F^{T_{sem}}$ for semantic maps $S_{sem}$ and $T_{sem}$, and then conducting the fusion in the VGG embedding space. 
\begin{equation}
F^S=F^{S_{sty}}_1\odot \omega F^{S_{sem}};\  F^T=F^{T_{sty}^t}_1\odot \omega F^{T_{sem}},
\label{semaware}
\end{equation}
where the fusion operation $\odot$ can be channel-wise concatenation or position-wise addition. We find these two operations could perform closely in some cases ({\em e.g.}, the $3^{rd}$ and $4^{th}$ top images in Fig.~\ref{fig:ablation}). But in general, concatenation often achieves more accurate semantic guidance (see the $3^{rd}$ and $4^{th}$ bottom images in Fig.~\ref{fig:ablation}) yet addition can provide faster speed (see the bottom efficiency comparison in Fig.~\ref{fig:ablation}).

After obtaining the fused features $F^S$ and $F^T$, inspired by \cite{chen2016fast}, we introduce a {\em Semantic-Guided Texture Warping (SGTW)} module with a specific field of view ({\em i.e.}, patch size $p$) to warp and transfer the textures. The detailed procedure is as follows:

\renewcommand\arraystretch{0.6}

{\small
\begin{enumerate}
	\setlength{\itemsep}{1pt}
	\setlength{\parsep}{0pt}
	\setlength{\partopsep}{0pt}
	\setlength{\parskip}{0pt}
	\item Extract a set of $p\times p$ {\em original} source patches from the {\em original} source feature $F^{S_{sty}}$, denoted by $\{\phi_i({F^{S_{sty}}})\}_{i\in \{1,\dots,n_s\}}$, where $n_s$ is the number of extracted patches.
	
	\item Extract a set of $p\times p$ {\em fused} source patches from the {\em fused} source feature $F^S$, denoted by $\{\phi_i({F^S})\}_{i\in \{1,\dots,n_s\}}$.
	
	\item Determine the closest-matching {\em fused} source patch for each {\em fused} target patch in $F^T$ by using a convolutional layer with the normalized fused source patches $\{\phi_i({F^S})/\parallel\phi_i({F^S})\parallel\}$ as filters and $F^T$ as input. The computed result $\mathcal{T}$ has $n_s$ feature channels, and each spatial location is a cosine similarity vector between a fused target patch and all fused source patches.
	
	\item Binarizing the scores in $\mathcal{T}$ such that the maximum value along the channel is {\bf 1} and the rest are {\bf 0}. The result is denoted as ${\hat {\mathcal{T}}}$.
	
	\item Generate the output $F^T_{out}$ by a deconvolutional layer with the {\em original} source patches $\{\phi_i(F^{S_{sty}})\}$ as filters and ${\hat {\mathcal{T}}}$ as input.
\end{enumerate}
}

The novel insight behind SGTW is that we exploit the semantic-guided matching relationship between the patches of {\em fused} features $F^S$ and $F^T$ to reassemble and warp the {\em original source} feature $F^{S_{sty}}$. This not only guarantees the accurate alignment with the target semantic map, but also theoretically ensures that the output feature $F^T_{out}$ can preserve the texture details of the original source feature $F^{S_{sty}}$ {\em losslessly}, since all its patches are from $F^{S_{sty}}$. Moreover, by specifying different views for SGTW, we can control the granularity of preserved texture details ({\em e.g.}, the integrity of inner structures) and the alignment accuracy with the target semantic map, as will be shown in later Sec.~\ref{stage1} and~\ref{stage2}. By leveraging SGTW, our VSTR thus can realize more accurate semantic-guided and structure-preserved texture transfer. {\em Note that the matching and reassembling steps actually only add two convolutional layers to the feed-forward networks, and thus their implementation is very efficient.}

{\bf Statistics-based Enhancement (SE).} This operation aims to enhance the holistic effects of the stylized target image based on global statistics matching. Either the first-order statistics ({\em e.g.}, mean and standard deviation)~\cite{huang2017arbitrary} or the second-order statistics ({\em e.g.}, covariance)~\cite{li2017universal} can be adopted. In practice, we find the first-order statistics can work better in our task. As shown in the last column of Fig.~\ref{fig:ablation}, though higher-order statistics can reproduce the surface gloss of ceramic teapot more faithfully, they may produce inferior results with hazy shadows and consume much more time. Thus, we define our SE operation as a simple first-order statistics matching.
\begin{equation}
\begin{aligned}
&SE(F^{S_{sty}}, F^{T_{sty}^t}) = \\
&\sigma(F^{S_{sty}})(\frac{F^{T_{sty}^t}-\mu(F^{T_{sty}^t})}{\sigma(F^{T_{sty}^t})}) + \mu(F^{S_{sty}}),
\end{aligned}
\end{equation}
where $\mu$ and $\sigma$ are the mean and standard deviation.

\vspace{-0.1em}
\subsection{Global View Structure Alignment Stage}
\label{stage1}
As introduced in Sec.~\ref{over}, the goal of this stage is to preserve the inner structures of the source textures as completely as possible so as to provide good structure guidance for subsequent stages. An important intuition we will use is that the inner structures with different scales can be captured from different views, and if we process from the global view, then the complete inner structures can be captured. For example, as we plotted in the semantic maps of Fig.~\ref{fig:problem}, if we match $T_{sem}$ and $S_{sem}$ from a local view ({\em i.e.}, use a small patch size $p$), only the patches covering small or boundary structures in $T_{sem}$ ({\em e.g.}, patch $a_1$ and $b_1$) can find the proper counterparts in $S_{sem}$ (patch $a_2$ and $b_2$). For those in the large plain regions ({\em e.g.}, patch $c_1$), it is hard to choose their best-suited partners among internal source patches ({\em e.g.}, patch $c_2$ and $c_3$), since they are completely identical (both full-blue). Thus, the inner structures in these regions cannot be retained, and the results would show severe wash-out effects, like image (b) in the right part of Fig.~\ref{fig:problem} (where only the local view of stage II (Sec.~\ref{stage2}) is used). However, if we enlarge the view of these hard patches to include some salient structures ({\em e.g.}, patch $d_1$), they can easily find the proper counterparts in $S_{sem}$ again (patch $d_2$). At this point, the complete inner structures can be well captured and preserved.

Following this intuition, we make a global view setting in the VSTR of this stage to handle the feature maps from the global view, {\em i.e.}, using a dynamic global/maximum patch size $p$ to cover the inner structures as completely as possible:
\begin{equation}
p = min [H(F^S), W(F^S), H(F^T), W(F^T)] - 1,
\label{patchsize}
\end{equation}
where $H$ and $W$ denote the height and width of the features. Unfortunately, operating directly on these large patches will severely grow the computation and time cost. To alleviate this issue, we resort to the deepest layer ({\em i.e.}, $Relu5\_1$) of VGG-19 to implement the global alignment, which kills two birds with one stone: (1) The costs can be minimized, since the features at this layer have the smallest size (see the efficiency comparison below $T_{sty}$ and image (a) in Fig.~\ref{fig:problem}). (2) This layer provides the highest-level structure features and the largest receptive field, perfectly suitable for this stage. For validation, we use a small local view ({\em i.e.}, $p=3$) in this stage~to obtain the right image (a) of Fig.~\ref{fig:problem}. As observed in the red rectangle area, the result still suffers from wash-out effects that lose the inner structures, but here the effects are alleviated to some extent compared to the right image (b) (which is aligned at a shallower layer $Relu4\_1$). It indicates that our global view setting can help capture more intact inner structures, and the deeper VGG layer can provide higher-level structure features and larger receptive fields for better global alignment.

\renewcommand\arraystretch{0.6}
\begin{figure}[t]
	\centering
	\setlength{\tabcolsep}{0.05cm}
	\begin{tabular}{ccc}
		\includegraphics[width=0.305\linewidth]{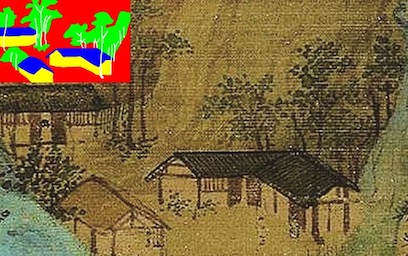}&
		\includegraphics[width=0.305\linewidth]{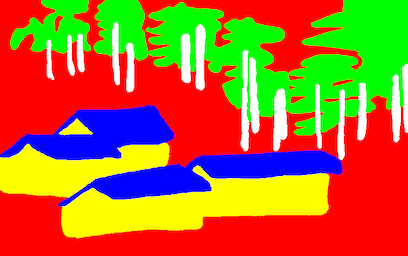}&
		\includegraphics[width=0.305\linewidth]{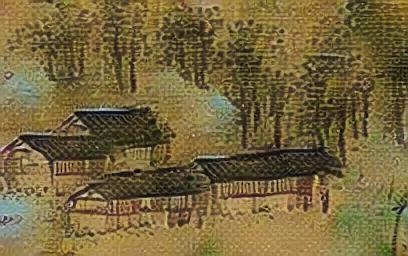}
		\\
		\includegraphics[width=0.205\linewidth]{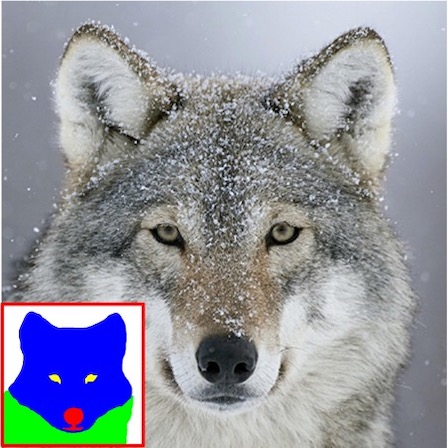}&
		\includegraphics[width=0.205\linewidth]{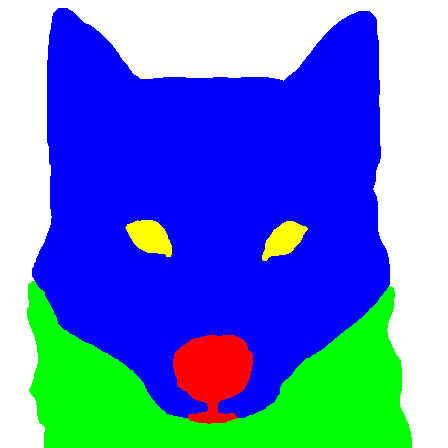}&
		\includegraphics[width=0.205\linewidth]{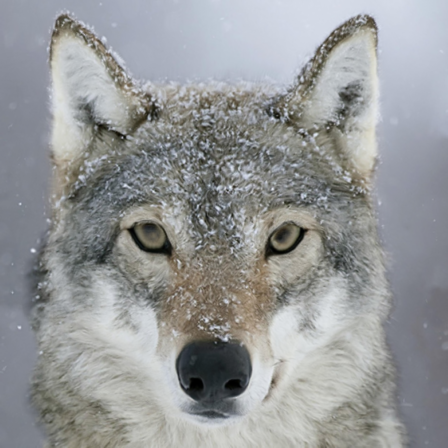}
		\\
		\footnotesize Input (source) & \footnotesize Input (semantics) &\footnotesize Output (target) 
		
	\end{tabular}
    \vspace{-0.5em}
	\caption{Doodles-to-artworks.
	}
	\label{fig:doodle}
	\vspace{-0.5em}
\end{figure}

\renewcommand\arraystretch{0.6}
\begin{figure}[t]
	\centering
	\setlength{\tabcolsep}{0.2cm}
	\begin{tabular}{ccc}
		\includegraphics[width=0.225\linewidth]{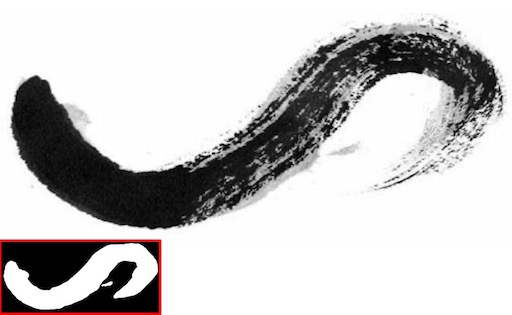}&
		\includegraphics[width=0.205\linewidth]{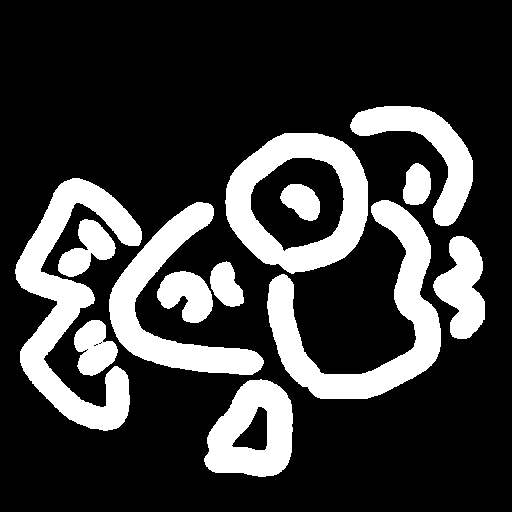}&
		\includegraphics[width=0.205\linewidth]{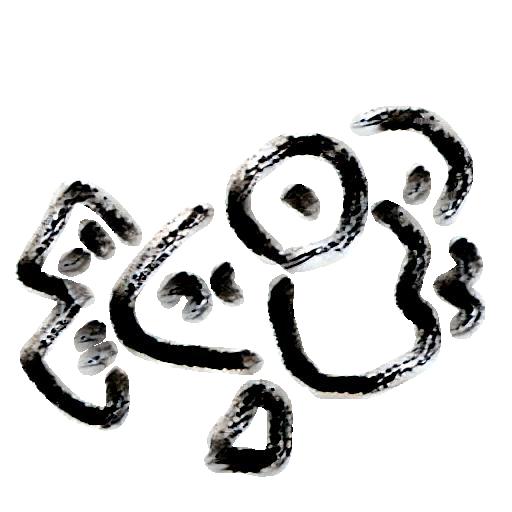}
		\\
		\includegraphics[width=0.183\linewidth]{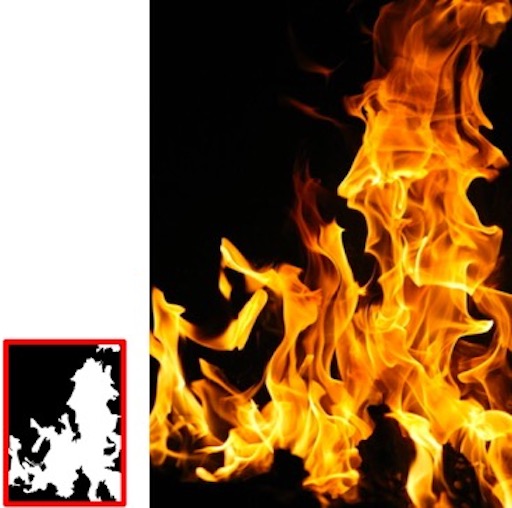}&
		\includegraphics[width=0.275\linewidth]{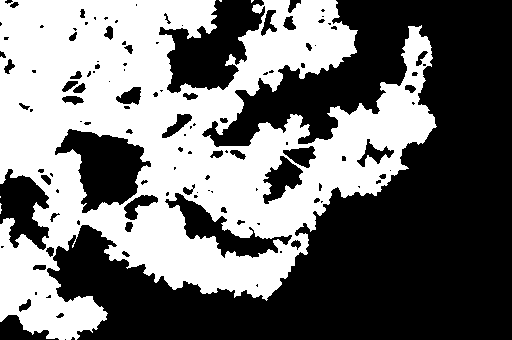}&
		\includegraphics[width=0.275\linewidth]{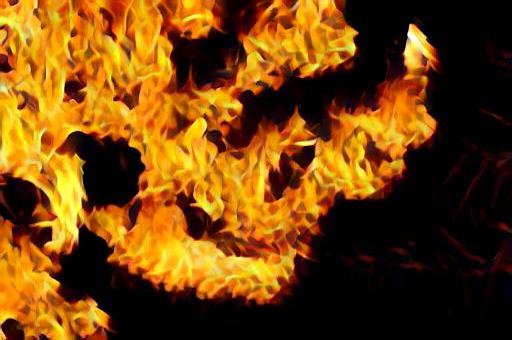}
		\\
		\footnotesize Input (source) & \footnotesize Input (semantics) &\footnotesize Output (target) 
		
	\end{tabular}
     \vspace{-0.5em}
	\caption{Texture Pattern Editing.
	}
	\vspace{-1em} 
	\label{fig:pattern}
\end{figure}

\vspace{-0.5em}
\subsection{Local View Texture Refinement Stage}
\label{stage2}
This stage takes the output of stage I as the input temporary stylized target image $T_{sty}^t$ to guide a more detailed synthesis. Similar to stage I, it also uses VSTR to process the bottleneck features. The difference is, we process the features at a relatively shallower layer $Relu4\_1$, and use a much smaller patch size ({\em i.e.}, $p=3$) to handle the features from only the local view. The effect of this stage can be inferred by comparing $T_{sty}$ with the right image (c) in Fig.~\ref{fig:problem}. The local view helps rectify and refine the local structures and texture details to a large extent, thus achieving more accurate alignment and higher quality.

\vspace{-0.5em}
\subsection{Holistic Effect Enhancement Stage}
\label{stage3}
The former two stages have been able to transfer satisfying inner structures and texture details. However, as they~are based on high-level features, the synthesized images often neglect the low-level holistic effects ({\em e.g.}, colors, brightness, and contrast), as shown in the right image (d) of Fig.~\ref{fig:problem}.~To further enhance these low-level effects, this stage utilizes the statistics-based enhancement (SE) on the low-level features at three shallow layers, {\em i.e.}, $Relu{\bf X}\_1$ ({\bf X}=1,2,3). As such, we can finally synthesize high-quality results which perform well in both high-level structures and low-level effects. Note that though our VSTR can also be used here to enhance the low-level effects, we do not recommend it as it will severely increase the time cost and memory requirement.

\renewcommand\arraystretch{0.6}
\begin{figure}[t]
	\centering
	\setlength{\tabcolsep}{0.02cm}
	\begin{tabular}{ccccc}
		\includegraphics[width=0.195\linewidth]{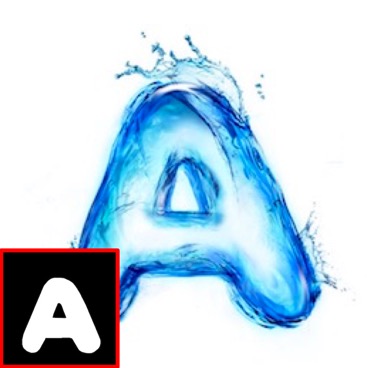}&
		\includegraphics[width=0.195\linewidth]{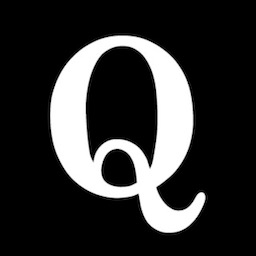}&
		\includegraphics[width=0.195\linewidth]{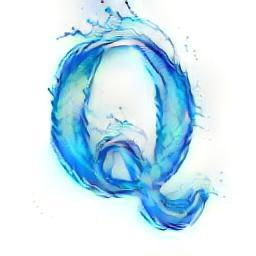}&
		\includegraphics[width=0.195\linewidth]{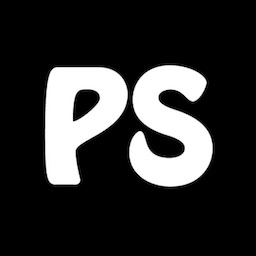}&
		\includegraphics[width=0.195\linewidth]{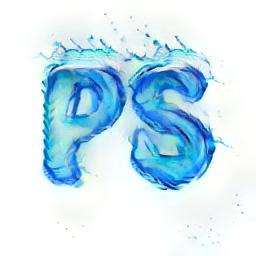}
		\\
		
		\includegraphics[width=0.195\linewidth]{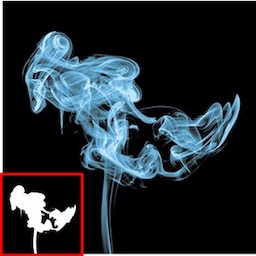}&
		\includegraphics[width=0.195\linewidth]{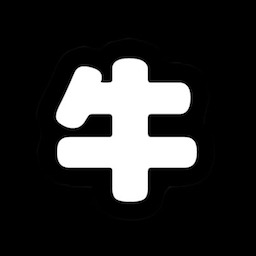}&
		\includegraphics[width=0.195\linewidth]{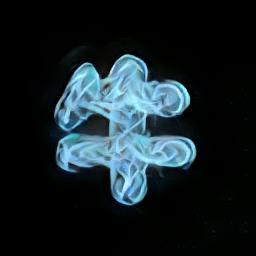}&
		\includegraphics[width=0.195\linewidth]{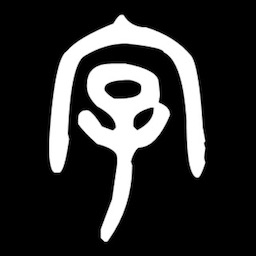}&
		\includegraphics[width=0.195\linewidth]{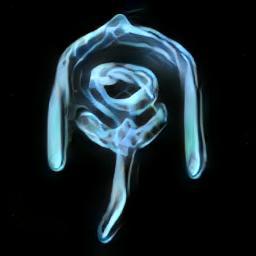}
		\\
		\footnotesize Input & \footnotesize Input &\footnotesize Output & \footnotesize Input &\footnotesize Output
		
	\end{tabular}
    \vspace{-0.5em}
	\caption{Text Effects Transfer.
	}
	\label{fig:text}
	\vspace{-0.5em}
\end{figure}

\renewcommand\arraystretch{0.6}
\begin{figure}[t]
	\centering
	\setlength{\tabcolsep}{0.02cm}
	\begin{tabular}{cccp{0.02cm}|p{0.02cm}ccc}
		\footnotesize Input & \footnotesize Input & \footnotesize Output &&&
		\footnotesize Input & \footnotesize Input &\footnotesize Output
		\\
		\includegraphics[width=0.16\linewidth]{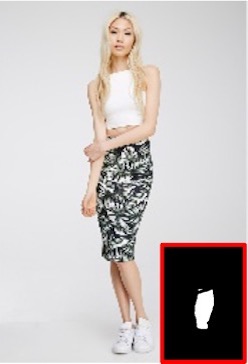}&
		\includegraphics[width=0.16\linewidth]{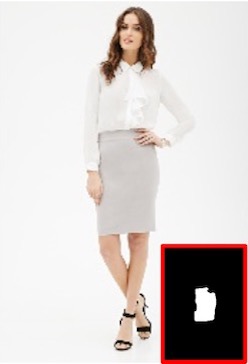}&
		\includegraphics[width=0.16\linewidth]{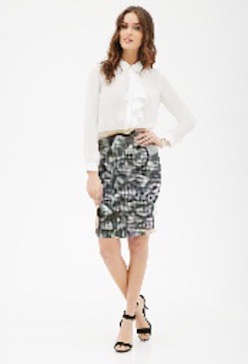}&&&
		\includegraphics[width=0.16\linewidth]{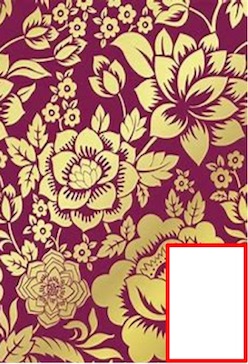}&
		\includegraphics[width=0.16\linewidth]{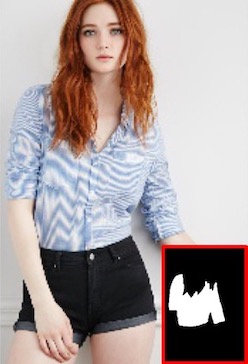}&
		\includegraphics[width=0.16\linewidth]{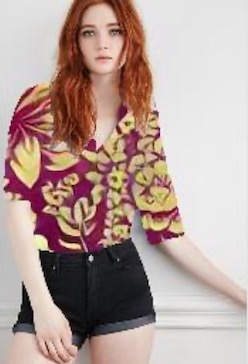}
		\\
		\multicolumn{3}{c}{\footnotesize (a) Clothing Texture $\rightarrow$ Clothing}&&&\multicolumn{3}{c}{\footnotesize (b) Painting Texture $\rightarrow$ Clothing}
		\\
		\includegraphics[width=0.16\linewidth]{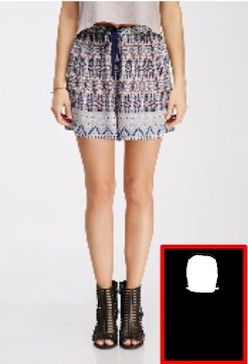}&
		\includegraphics[width=0.16\linewidth]{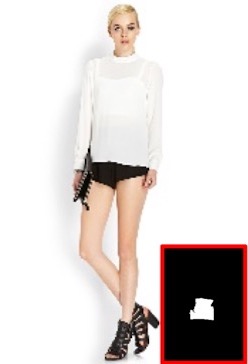}&
		\includegraphics[width=0.16\linewidth]{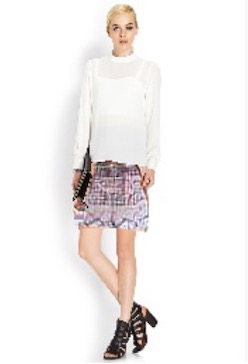}&&&
		\includegraphics[width=0.16\linewidth]{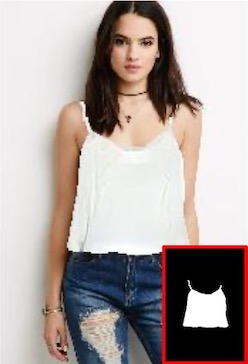}&
		\includegraphics[width=0.16\linewidth]{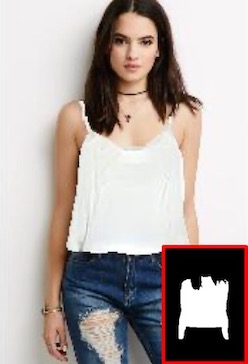}&
		\includegraphics[width=0.16\linewidth]{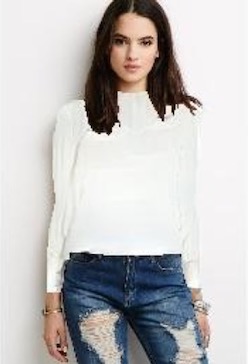}
		\\
		\multicolumn{3}{c}{\footnotesize (c) Virtual Try-on}&&&\multicolumn{3}{c}{\footnotesize (d) Clothing Shape Editing}
		
	\end{tabular}
     \vspace{-0.5em}
	\caption{Virtual Clothing Manipulation.
	}
	\label{fig:cloth}
	\vspace{-1em}
\end{figure}

\renewcommand\arraystretch{0.1}
\begin{figure*}[t]
	\centering
	\setlength{\tabcolsep}{0.05cm}
	\begin{tabular}{ccccccccc}
		\includegraphics[width=0.104\linewidth]{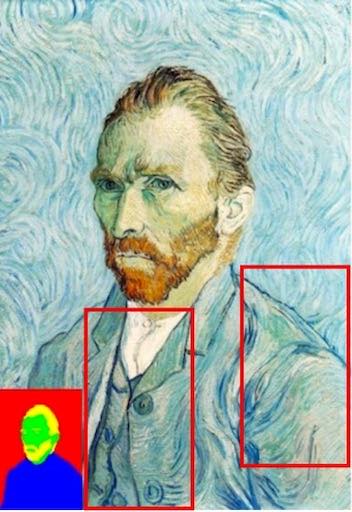}&
		\includegraphics[width=0.104\linewidth]{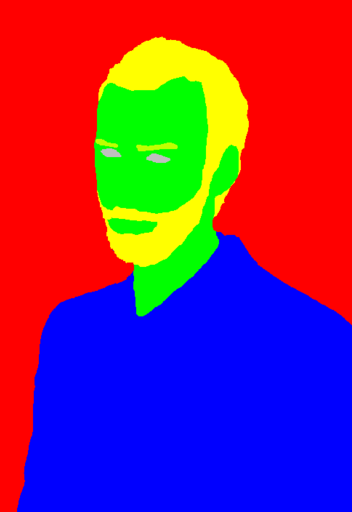}&
		\includegraphics[width=0.104\linewidth]{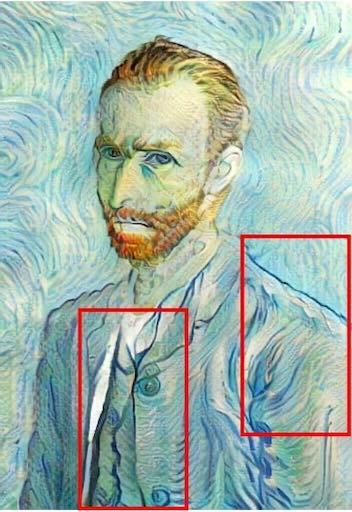}&
		\includegraphics[width=0.104\linewidth]{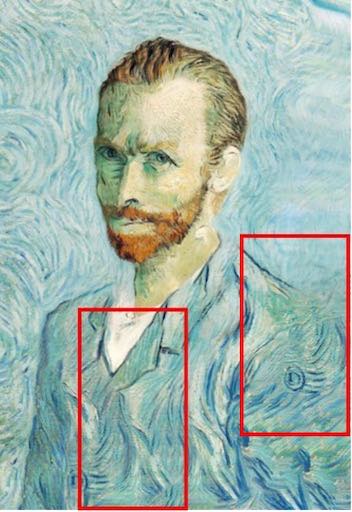}&
		\includegraphics[width=0.104\linewidth]{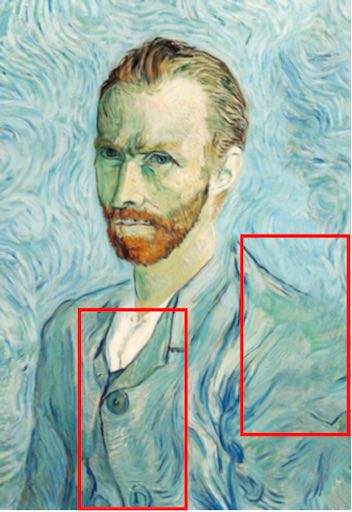}&
		\includegraphics[width=0.104\linewidth]{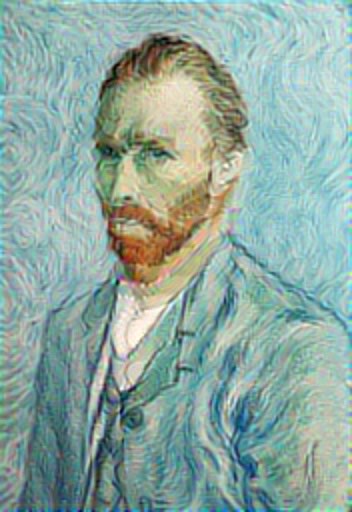}&
		\includegraphics[width=0.104\linewidth]{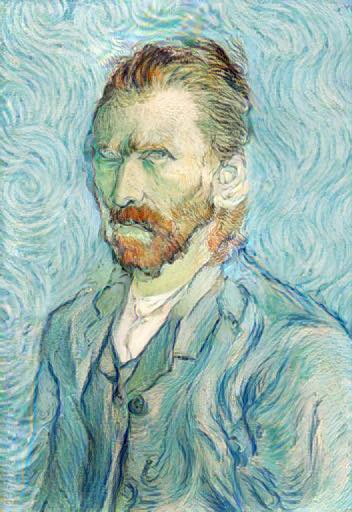}&
		\includegraphics[width=0.104\linewidth]{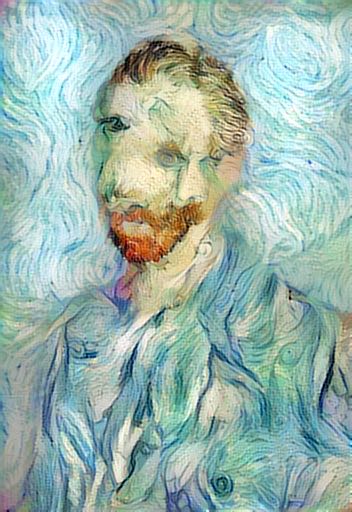}&
		\includegraphics[width=0.104\linewidth]{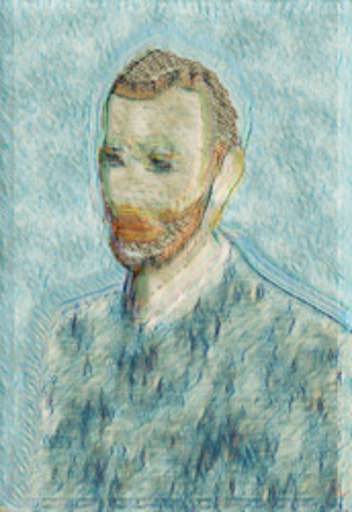}
		\\
		
		\includegraphics[width=0.104\linewidth]{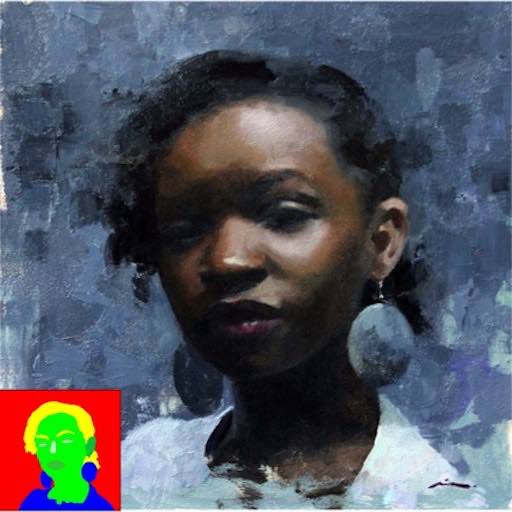}&
		\includegraphics[width=0.104\linewidth]{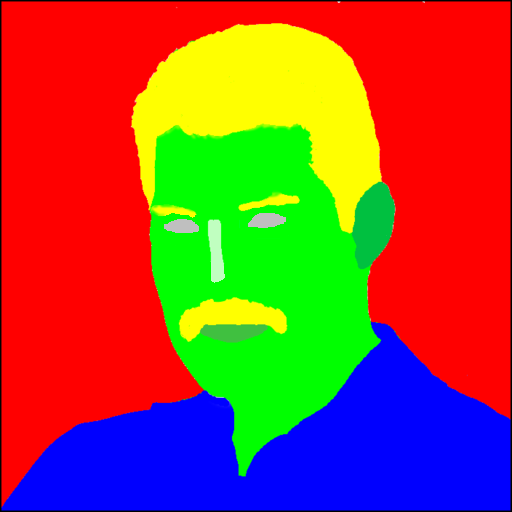}&
		\includegraphics[width=0.104\linewidth]{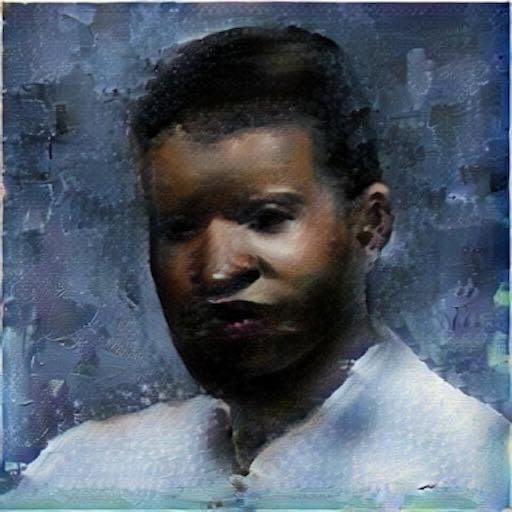}&
		\includegraphics[width=0.104\linewidth]{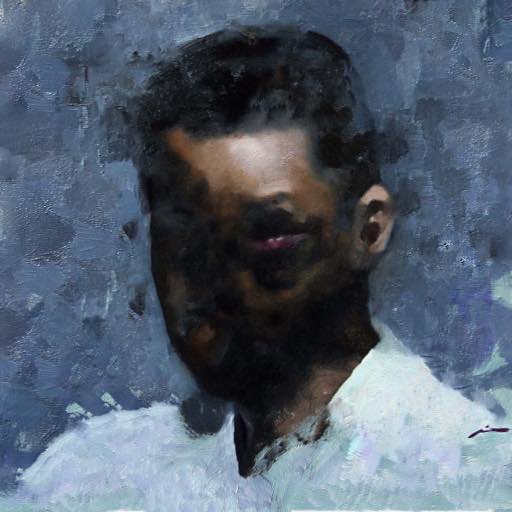}&
		\includegraphics[width=0.104\linewidth]{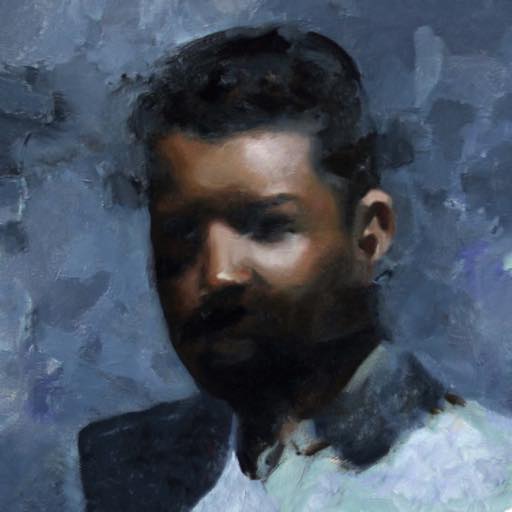}&
		\includegraphics[width=0.104\linewidth]{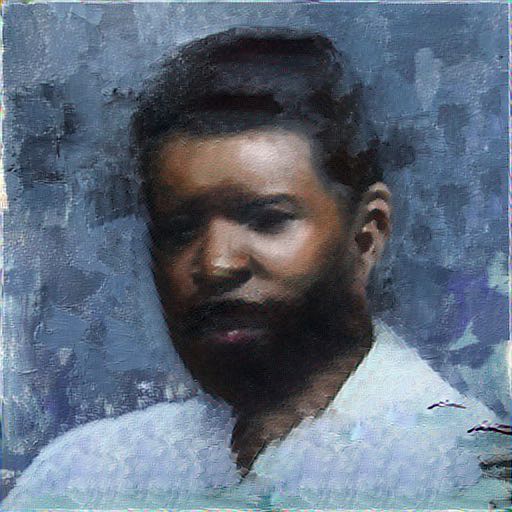}&
		\includegraphics[width=0.104\linewidth]{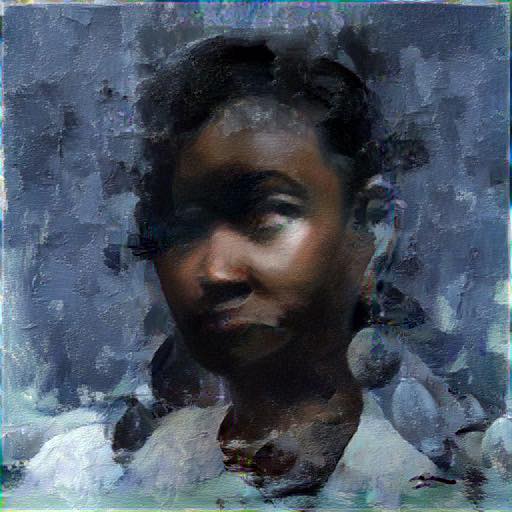}&
		\includegraphics[width=0.104\linewidth]{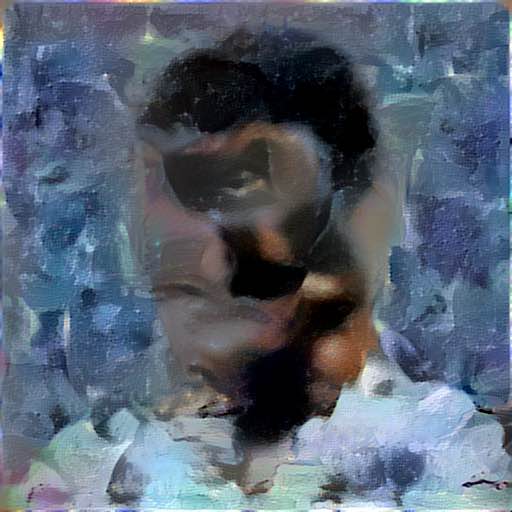}&
		\includegraphics[width=0.104\linewidth]{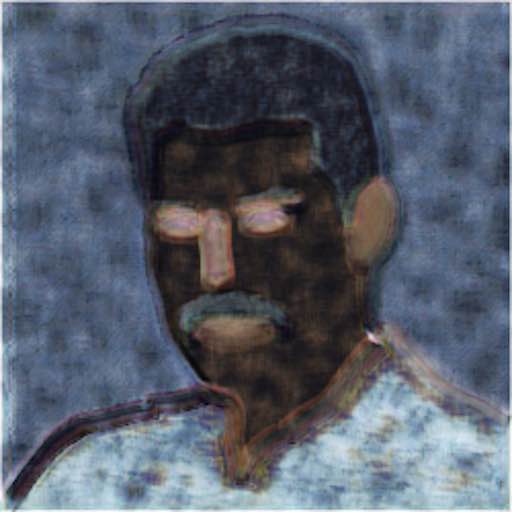}
		\\
		
		\includegraphics[width=0.104\linewidth]{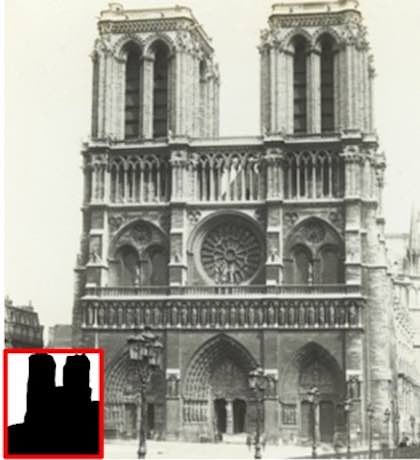}&
		\includegraphics[width=0.088\linewidth]{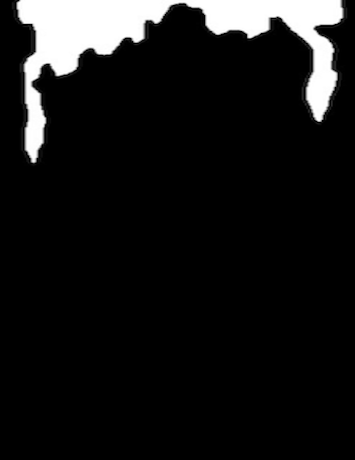}&
		\includegraphics[width=0.088\linewidth]{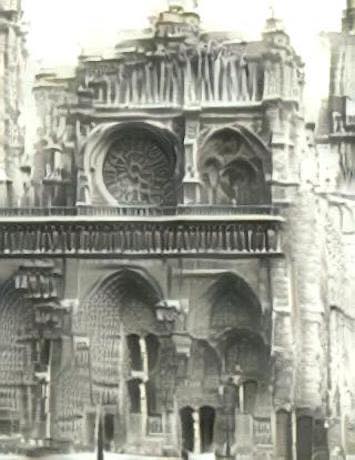}&
		\includegraphics[width=0.088\linewidth]{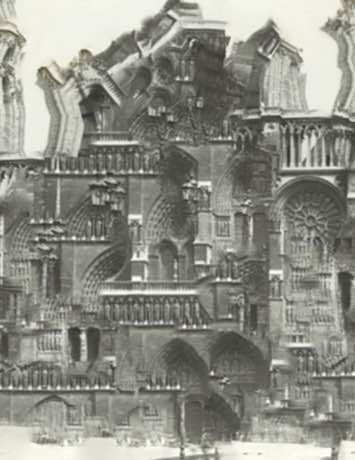}&
		\includegraphics[width=0.088\linewidth]{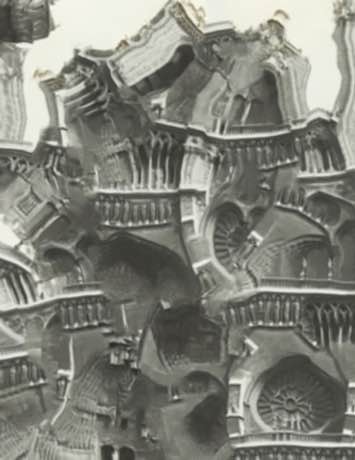}&
		\includegraphics[width=0.088\linewidth]{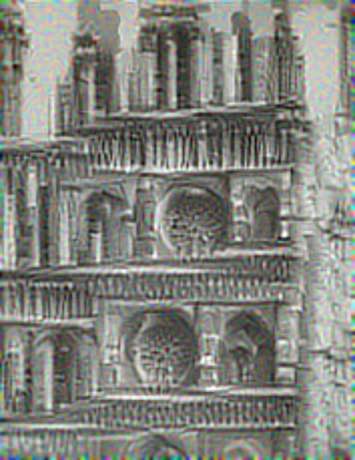}&
		\includegraphics[width=0.088\linewidth]{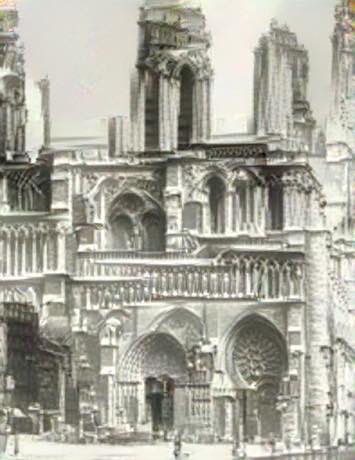}&
		\includegraphics[width=0.088\linewidth]{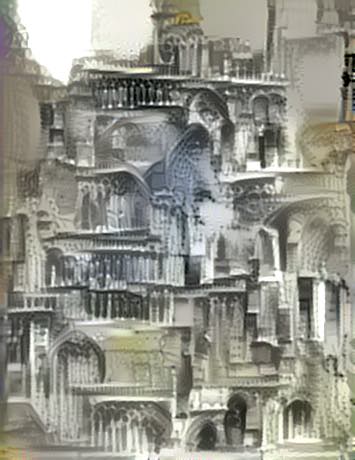}&
		\includegraphics[width=0.088\linewidth]{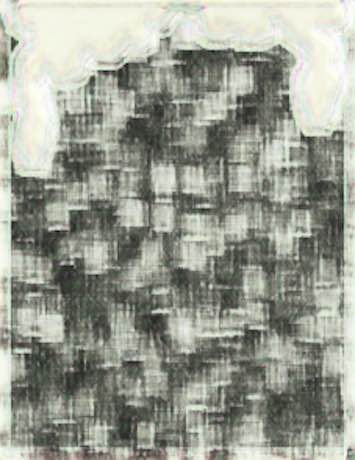}
		\\

		\includegraphics[width=0.104\linewidth]{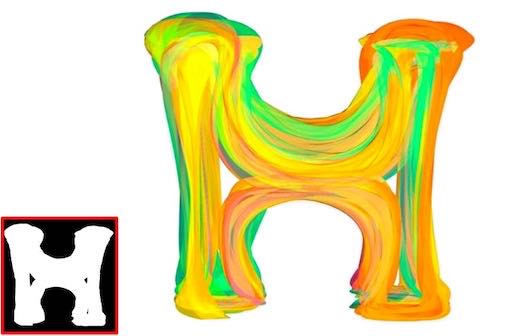}&
		\includegraphics[width=0.104\linewidth]{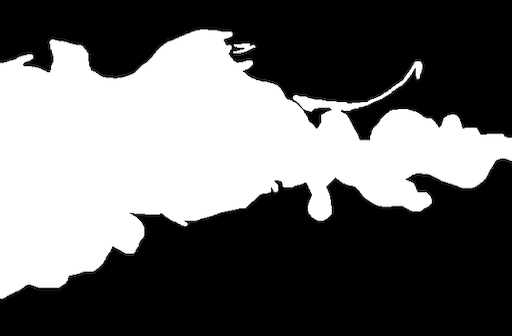}&
		\includegraphics[width=0.104\linewidth]{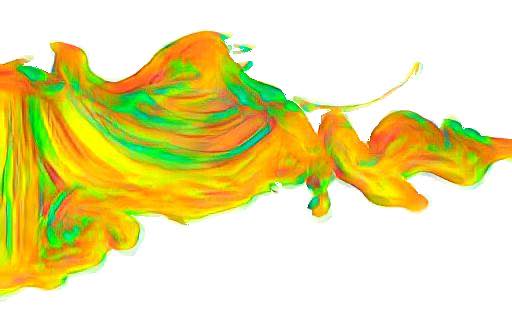}&
		\includegraphics[width=0.104\linewidth]{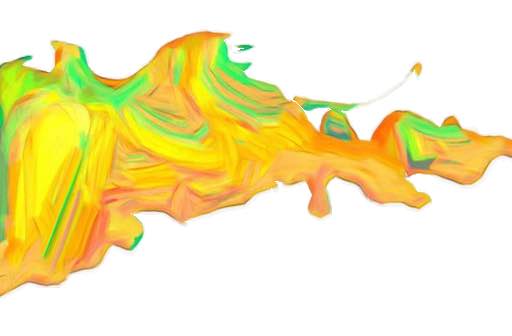}&
		\includegraphics[width=0.104\linewidth]{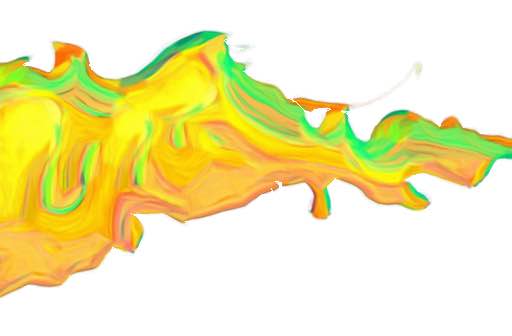}&
		\includegraphics[width=0.104\linewidth]{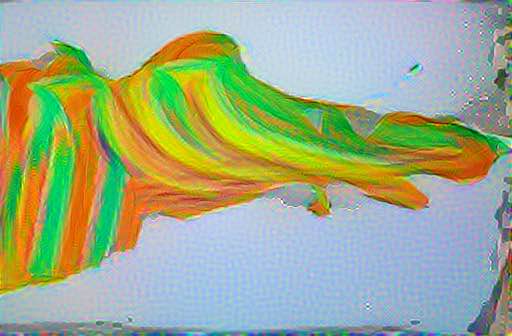}&
		\includegraphics[width=0.104\linewidth]{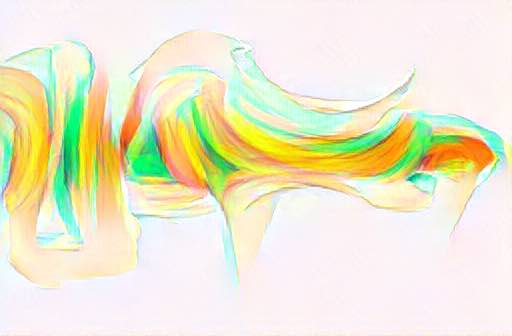}&
		\includegraphics[width=0.104\linewidth]{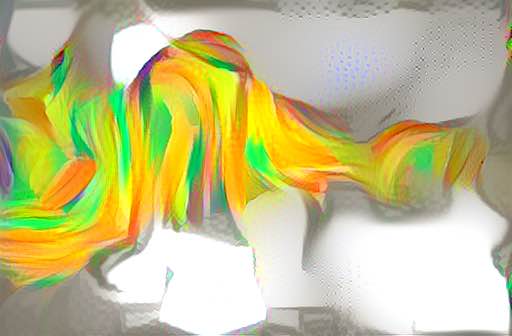}&
		\includegraphics[width=0.104\linewidth]{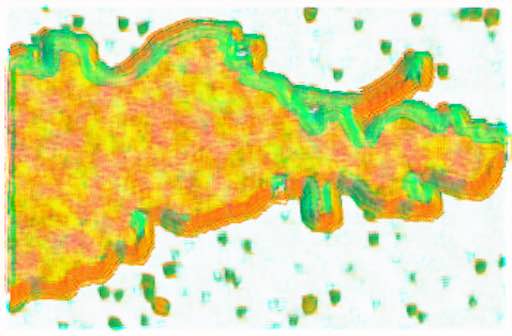}
		\\
		
		\includegraphics[width=0.104\linewidth]{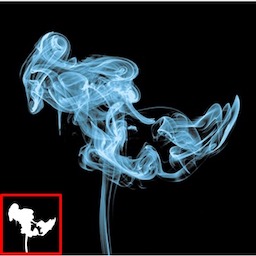}&
		\includegraphics[width=0.104\linewidth]{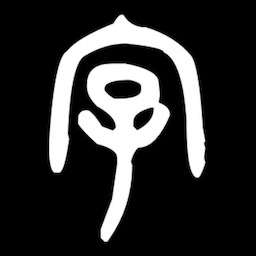}&
		\includegraphics[width=0.104\linewidth]{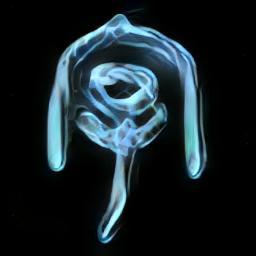}&
		\includegraphics[width=0.104\linewidth]{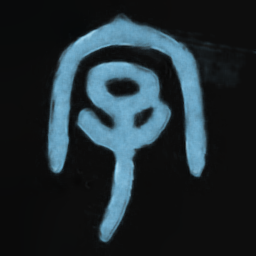}&
		\includegraphics[width=0.104\linewidth]{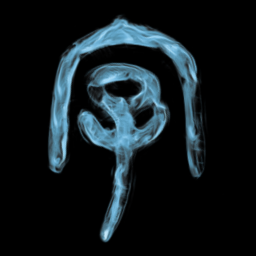}&
		\includegraphics[width=0.104\linewidth]{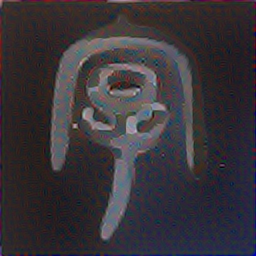}&
		\includegraphics[width=0.104\linewidth]{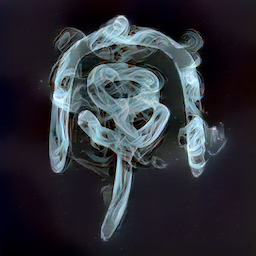}&
		\includegraphics[width=0.104\linewidth]{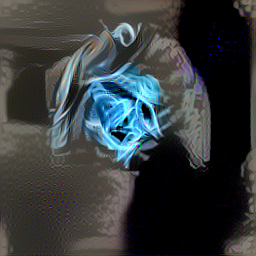}&
		\includegraphics[width=0.104\linewidth]{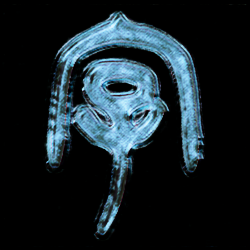}
		\\
		\vspace{0.2em} 
		\\
		\footnotesize ($S_{sem}$, $S_{sty}$) & \footnotesize $T_{sem}$& \footnotesize {\bf Ours} & \footnotesize T-Effect & \footnotesize CFITT& \footnotesize Neural Doodle & \footnotesize STROTSS & \footnotesize Gatys2017 & \footnotesize TuiGAN 
	\end{tabular}
    \vspace{-0.5em}
	\caption{Qualitative comparison with the state-of-the-art universal interactive texture transfer methods. See more in {\em SM}.
	}
	\vspace{-1em} 
	\label{fig:quality}
\end{figure*}

\vspace{-0.5em} 
\section{Experimental Results}
\subsection{Implementation Details}
We adopt concatenation as the default setting to fuse semantic guidance. The hyperparameters that control the semantic-awareness (Eq.~\ref{semaware}) in stage I and stage II are set to $\omega_1=\omega_2=50$ ($\omega_1$ for stage I, $\omega_2$ for stage II. See {\em supplementary material (SM)} for their effects).

\subsection{Applications}
Our framework can be effectively applied to multiple interactive texture transfer tasks, such as doodles-to-artworks, texture pattern editing, text effects transfer, and virtual clothing manipulation (see the examples in Fig.~\ref{fig:teaser},~\ref{fig:doodle},~\ref{fig:pattern},~\ref{fig:text},~\ref{fig:cloth}).

{\bf Doodles-to-artworks.} This task aims to turn the two-bit doodles annotated by users into fine artworks with similar styles as the given exemplar paintings or photographs. Some of our results are shown in Fig.~\ref{fig:doodle}. See more in {\em SM}.

{\bf Texture Pattern Editing.} As illustrated in Fig.~\ref{fig:pattern}, given an exemplar image, users can edit the texture patterns such as path and shape according to their needs. This provides a controllable way to modify the existing patterns.

{\bf Text Effects Transfer.} As shown in Fig.~\ref{fig:text}, our method is also effective for text effects transfer which can migrate the artistic effects of stylized text images or source styles onto arbitrary raw plain texts.

{\bf Virtual Clothing Manipulation.} Manipulating the clothing textures and distributions in a virtual way is an interesting and practical problem that has attracted much attention in recent years~\cite{han2018viton,han2019clothflow,han2019finet}. Existing methods customized for this task usually learn the generation from a large-scale dataset~\cite{liu2016deepfashion}. Unlike them, our framework can also be applied to this task, but it generates the result using only one exemplar image. As shown in Fig.~\ref{fig:cloth}, our method can transfer the clothing or painting textures to other clothing ({\em e.g.}, (a) and (b)), virtually try on target clothing ({\em e.g.}, (c)), or edit the clothing shape ({\em e.g.}, (d)).

\subsection{Comparisons}
\label{cmp}
We compare our method with SOTA universal interactive texture transfer algorithms including two conventional methods (T-Effect~\cite{yang2017awesome} and CFITT~\cite{men2018common}), three neural-based methods (Neural Doodle~\cite{champandard2016semantic}, STROTSS~\cite{kolkin2019style}, and Gatys2017~\cite{gatys2017controlling}), and one GAN-based method (TuiGAN~\cite{lin2020tuigan}). For a fair comparison, we use their default settings except that the content weights of Neural Doodle, STROTSS, and Gatys2017 are set to $0$, as there is no content image corresponding to $T_{sem}$ in our task.

{\bf Qualitative Comparison.} The qualitative results are shown in Fig.~\ref{fig:quality}. Compared with T-Effect and CFITT, our method can synthesize higher quality results with better-preserved structures ({\em e.g.}, the red rectangle areas in the $1^{st}$ row, the face or clothing areas in the $2^{nd}$ row, and the building structures in the $3^{rd}$ row) and more vivid stylization effects ({\em e.g.}, the bottom two rows). For Neural Doodle, as it only bases on high-level features, it~fails to reproduce clear images with low-level details and often introduces pixel noises. Moreover, STROTSS cannot achieve accurate semantic guidance, producing poor results with missing details ({\em e.g.}, the eyes in the $1^{st}$ row) and misaligned structures ({\em e.g.}, the $2^{nd}$ and $3^{rd}$ rows). Gatys2017 matches the global statistics ({\em i.e.}, Gram matrix) for each semantic area, which cannot preserve the local texture structures. TuiGAN is hard to learn the underlying relationship between two images with a large domain gap ({\em e.g.}, semantic map and painting), thus cannot translate the exquisite texture details appropriately.

\renewcommand\arraystretch{1.1}
\begin{table}[t]
	\centering
	\setlength{\tabcolsep}{0.35cm}
	\caption{Quantitative Comparison. $\uparrow$: Higher is better. $\downarrow$: Lower is better.}
	\vspace{-0.5em}
	\begin{threeparttable}[b]
		\begin{tabular}{c|ccc}
			\hline
			\hline
			{\bf \footnotesize Method} &  \footnotesize SSIM $\uparrow$ & \footnotesize LPIPS $\downarrow$ &\footnotesize Style Loss $\downarrow$
			\\
			
			\hline
			\footnotesize T-Effect&\footnotesize 0.276 &\footnotesize 0.449  &\footnotesize 0.740  \\
			\footnotesize CFITT&\footnotesize  0.343  &\footnotesize  0.466 &\footnotesize 0.746\\
			\footnotesize Neural Doodle&  \footnotesize 0.302  &\footnotesize  0.481 &\footnotesize 0.867\\
			\footnotesize STROTSS&  \footnotesize 0.315 &  \footnotesize 0.423  &\footnotesize 0.738\\
			\footnotesize Gatys2017&   \footnotesize 0.200  &   \footnotesize 0.578 &\footnotesize  0.639 \\
			\footnotesize TuiGAN&  \footnotesize 0.247 &   \footnotesize 0.654 &\footnotesize 1.195\\
			\hline
			\footnotesize Ours& \footnotesize \bf 0.372  & \footnotesize \bf  0.384  & \footnotesize \bf  0.497\\
			\hline
			\hline
			
		\end{tabular}
	\end{threeparttable} 
	\vspace{-1em} 
	\label{quantity}
\end{table}

{\bf Quantitative Comparison.} In addition to the visual comparison, we also make a quantitative comparison. We adopt the Structural Similarity Index (SSIM)~\cite{wang2004image} and the Learned Perceptual Image Patch Similarity (LPIPS)~\cite{zhang2018unreasonable} between the stylized source images and the synthesized stylized target images as the metrics to measure the performance of the structure information preservation in texture transfer. Inspired by~\cite{gatys2016image,huang2017arbitrary}, we also use the style loss between the stylized source images and the synthesized stylized target images to assess the ability to transfer the holistic styles of the stylized source images. As Table~\ref{quantity} shows, our method achieves the highest SSIM score and the lowest LPIPS score and style loss, which indicates that the proposed method not only has a stronger ability to preserve more structure information of the stylized source textures, but also can better transfer the holistic styles of the stylized source images.

\renewcommand\arraystretch{1.1}
\begin{table}[t]
	\centering
	\setlength{\tabcolsep}{0.115cm}
	\caption{Execution time comparison. OOM: out of memory.}
	\vspace{-0.5em}
	\begin{threeparttable}[b]
		\begin{tabular}{c|cc|cc}
			\hline
			\hline
			\multirow{2}{*}{{\bf \footnotesize Method}\tnote{1}}
			&\multicolumn{2}{|c}{\footnotesize 256 $\times$ 256 (sec)}&\multicolumn{2}{|c}{\footnotesize 512 $\times$ 512 (sec)}\\
			
			\cline{2-5}& \footnotesize CPU& \footnotesize GPU& \footnotesize CPU& \footnotesize GPU\\
			
			\hline
			\footnotesize T-Effect&\footnotesize 101.52 &\footnotesize - & \footnotesize 241.85 & \footnotesize - \\
			\footnotesize CFITT&\footnotesize 112.05 & \footnotesize - & \footnotesize 572.21 & \footnotesize -  \\
			\footnotesize Neural Doodle&\footnotesize $\sim$2.8$\times 10^3$&\footnotesize 178.78 & \footnotesize $\sim$1.6$\times 10^4$ & \footnotesize $\sim$1.1$\times 10^3$\\
			\footnotesize STROTSS&\footnotesize $\sim$1.4$\times 10^3$&\footnotesize 262.32 & \footnotesize $\sim$3.9$\times 10^3$ & \footnotesize 668.84\\
			\footnotesize Gatys2017&\footnotesize $\sim$2.2$\times 10^3$ & \footnotesize 144.78 & \footnotesize $\sim$1.0$\times 10^4$& \footnotesize 574.81 \\
			\footnotesize TuiGAN&\footnotesize -&\footnotesize $\sim$1.8$\times 10^4$& \footnotesize - & \footnotesize OOM\\
			\hline
			\footnotesize Ours& \footnotesize \bf 2.573 & \footnotesize \bf 0.232 & \footnotesize \bf 12.381 & \footnotesize \bf 0.956 \\
			\hline
			\hline
			
		\end{tabular}
		\begin{tablenotes}
			\tiny
			\item[1] \scriptsize Tested on a 3.3 GHz hexa-core CPU and a 6GB Nvidia 1060 GPU.
		\end{tablenotes}
	\end{threeparttable} 
	\vspace{-2em} 
	\label{time}
\end{table}

{\bf Efficiency.} In Table~\ref{time}, we compare the running time~with the competitors. Compared with conventional methods T-Effect and CFITT on CPU, our method achieves $1$-$2$ orders of magnitude faster in resolution $256\times 256$ and $512\times 512$. Our speed can be further accelerated by using a GPU card, eventually reaching $2$-$5$ orders of magnitude faster than SOTA. Note that TuiGAN needs several hours and much more memory to train a model for each image pair.

{\bf User Study.} We also conduct a user study to evaluate the quality quantitatively. Given unlimited time, 50 users are asked to select the favorite ones from 40 octets of images comprising three inputs ($S_{sty}$, $S_{sem}$, $T_{sem}$), and five randomly shuffled outputs (T-Effect, CFITT, Neural Doodle, STROTSS, and ours\footnote[1]{Gatys2017 and TuiGAN are not compared here because the quality of their results is clearly inferior to ours, as shown in Fig.~\ref{fig:quality}.}). We collect 2000 responses in total. The statistics indicate that our method achieves subjectively preferred results ({\bf 32.1\%}) than T-Effect (23.8\%), CFITT (26.2\%), Neural Doodle (10.3\%), and STROTSS~(7.6\%).

\vspace{-0.6em}
\section{Discussion and Conclusion}
In this paper, we propose a novel neural-based framework, dubbed {\em texture reformer}, for fast and universal interactive texture transfer. A feed-forward multi-view and multi-stage synthesis procedure is imposed to synthesize high-quality results with coherent structures and fine texture details from coarse to fine. Moreover, we also introduce a novel learning-free {\em view-specific texture reformation (VSTR)} operation with a new semantic map guidance strategy to realize more accurate semantic-guided and structure-preserved texture transfer. Experimental results demonstrate the effectiveness of our framework on many texture transfer tasks. And compared with SOTA algorithms, it not only achieves higher quality results but also is 2-5 orders of magnitude faster.

However, the method still suffers from some limitations. The inherent nature of patch alignment used in the VSTR operation may result in a few quality issues. And since our framework does not make patches rotated or scaled, it may not work very well for semantic maps with drastically different shapes and large geometric deformation. Besides, since our method does not use any additional structure guidance or distribution constraint, it may fail to achieve correct texture transfer for patterns with decorative elements. We strongly encourage the readers to refer to the {\em SM} for the comprehensive discussions on our limitations. 

There are also a lot of prior arts that can give enlightenment to improve our method. For example, one may consider integrating the histogram matching techniques in \cite{risser2017stable} into our VSTR or SE operations to improve the performance. To preserve the shading of the source textures, one may consider adding the lighting map constraint like \cite{tsin2001texture} on the intermediate features of our VSTR or SE operations. To achieve geometric texture transfer, one may model a deformation field like~\cite{liu2004near}, or borrow some ideas from recent geometric style transfer approaches~\cite{kim2020deformable,liu2021learning}. Given the simplicity of our framework, we believe there is substantial room for improvement. In addition, the compression and acceleration of the framework to achieve real-time interactive texture transfer is also a practical direction worthy of further exploration.

\section*{Acknowledgements}
This work was supported in part by the projects No. 2020YFC1523101, 19ZDA197, LY21F020005, 2021009, 2019C03137, NSFC project: research on key technologies of art image restoration based on decoupling learning (62172365), MOE Frontier Science Center for Brain Science \& Brain-Machine Integration (Zhejiang University), and Key Scientific Research Base for Digital Conservation of Cave Temples (Zhejiang University), State Administration for Cultural Heritage.

{\small
	\bibliographystyle{aaai22}
	\bibliography{aaai22}
}

\clearpage

\appendix

\section{Supplementary Material}

\subsection{Effects of the semantic hyperparameters}
\label{sec1}
\begin{figure*}[h]
	\centering
	\setlength{\tabcolsep}{0.1cm}
	\begin{tabular}{cccccccc}
		\multirow{1}{*}[0.6in]{$\omega_2=0$}&
		\includegraphics[width=0.115\linewidth]{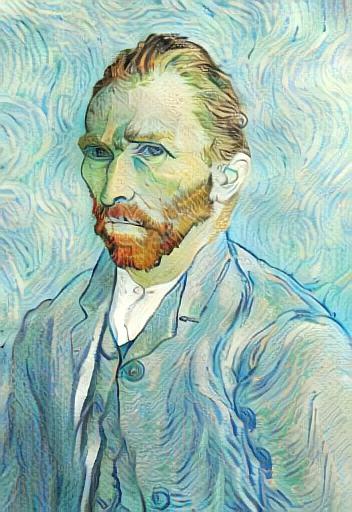}&
		\includegraphics[width=0.115\linewidth]{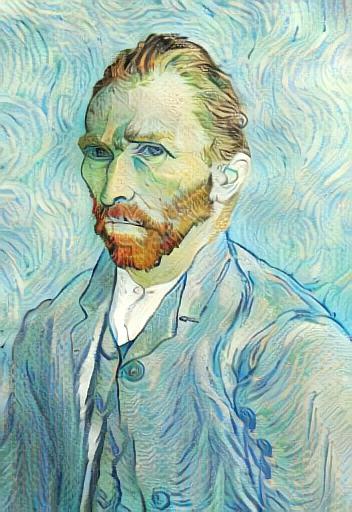}&
		\includegraphics[width=0.115\linewidth]{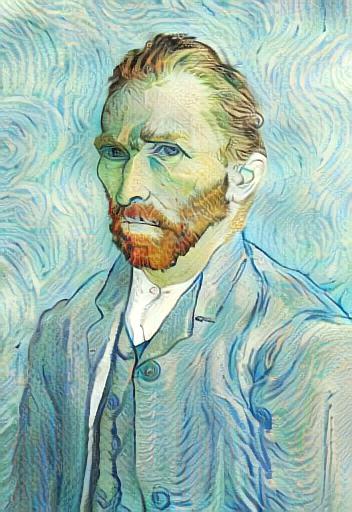}&
		\includegraphics[width=0.115\linewidth]{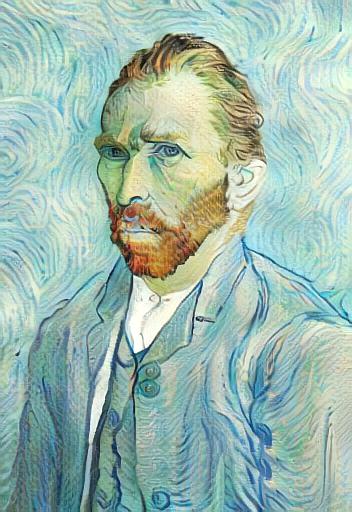}&
		\includegraphics[width=0.115\linewidth]{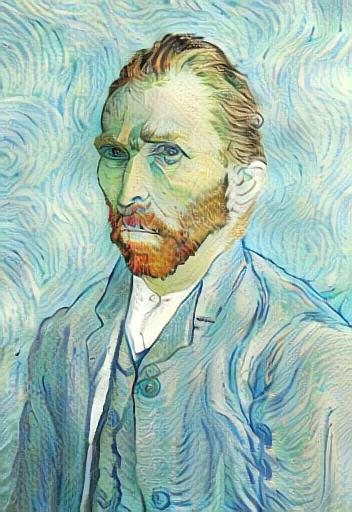}&
		\includegraphics[width=0.115\linewidth]{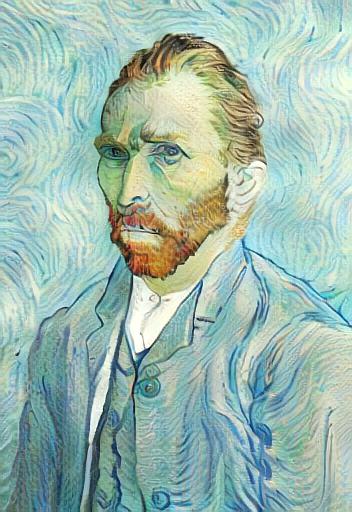}&
		\includegraphics[width=0.115\linewidth]{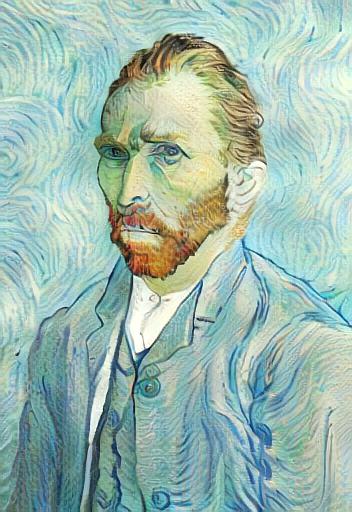}
		\\
		& $\omega_1=0$& $\omega_1=0.1$& $\omega_1=1$& $\omega_1=10$& \bf $\textcolor{red}{\omega_1=50}$& $\omega_1=100$& $\omega_1=1000$
		\\
		
		\multirow{1}{*}[0.6in]{$\omega_1=0$}&
		\includegraphics[width=0.115\linewidth]{fig/hyper/0_0}&
		\includegraphics[width=0.115\linewidth]{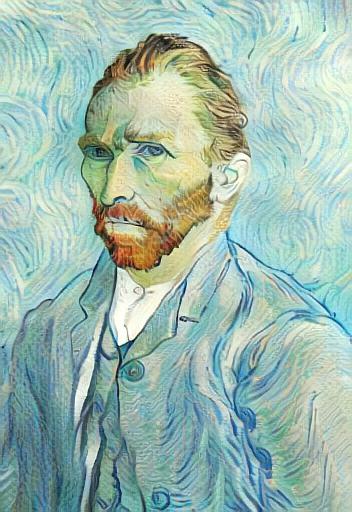}&
		\includegraphics[width=0.115\linewidth]{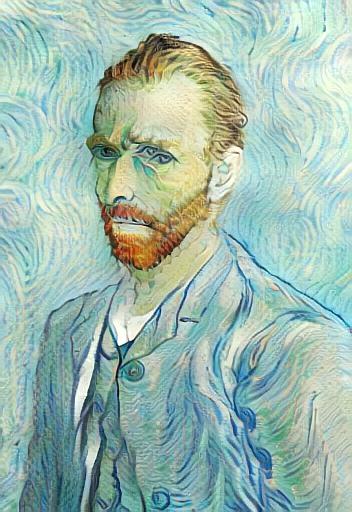}&
		\includegraphics[width=0.115\linewidth]{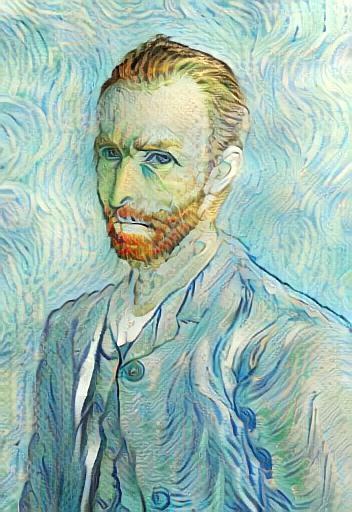}&
		\includegraphics[width=0.115\linewidth]{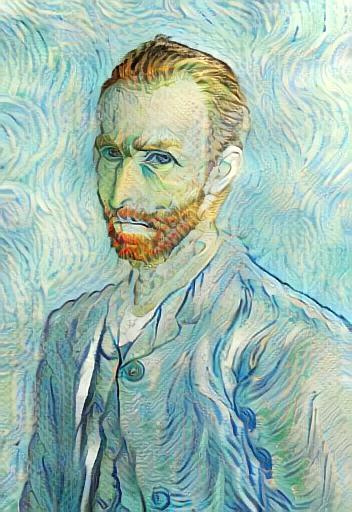}&
		\includegraphics[width=0.115\linewidth]{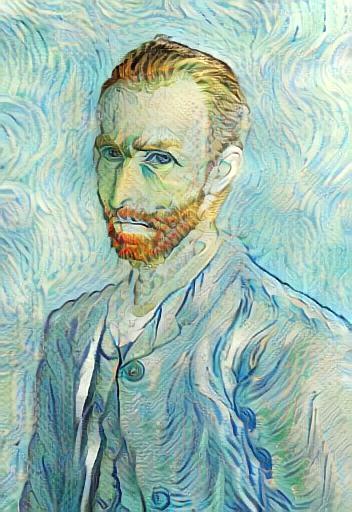}&
		\includegraphics[width=0.115\linewidth]{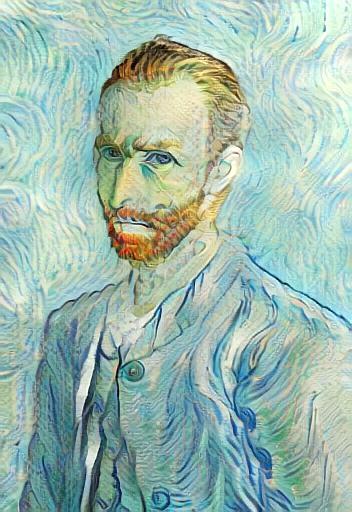}
		\\
		& $\omega_2=0$& $\omega_2=0.1$& $\omega_2=1$& $\omega_2=10$& \bf $\textcolor{red}{\omega_2=50}$& $\omega_2=100$& $\omega_2=1000$
		\\
		
		\multirow{1}{*}[0.6in]{$\omega_1=50$}&
		\includegraphics[width=0.115\linewidth]{fig/hyper/50_0}&
		\includegraphics[width=0.115\linewidth]{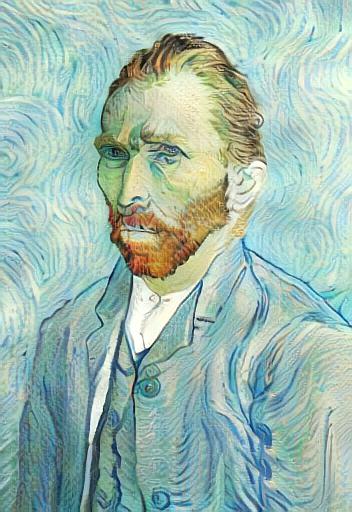}&
		\includegraphics[width=0.115\linewidth]{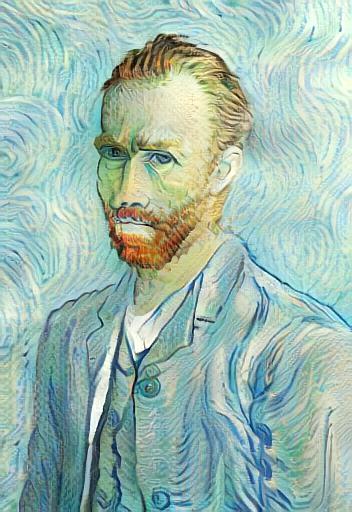}&
		\includegraphics[width=0.115\linewidth]{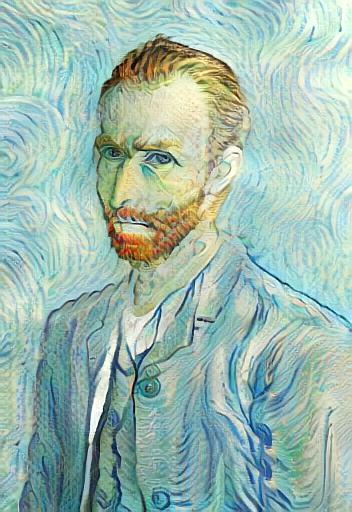}&
		\includegraphics[width=0.115\linewidth]{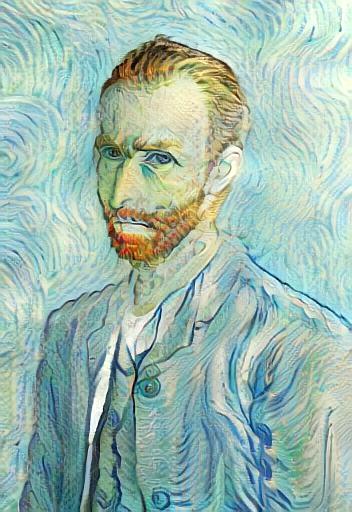}&
		\includegraphics[width=0.115\linewidth]{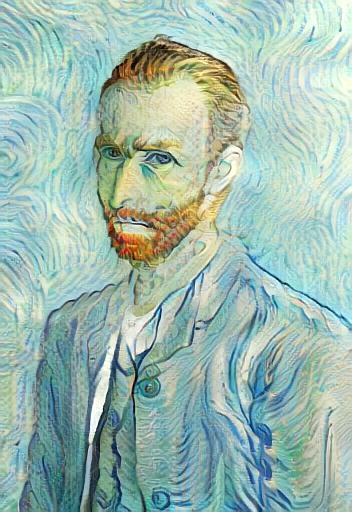}&
		\includegraphics[width=0.115\linewidth]{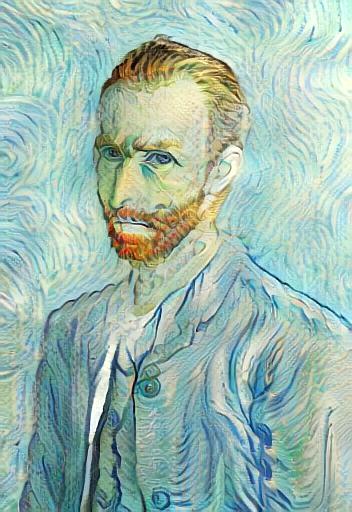}
		\\
		& $\omega_2=0$& $\omega_2=0.1$& $\omega_2=1$& $\omega_2=10$& \bf $\textcolor{red}{\omega_2=50}$& $\omega_2=100$& $\omega_2=1000$

	\end{tabular}
	\vspace{1em}
	\caption{Effects of the hyperparameters ($\omega_1$ and $\omega_2$) to control the semantic-awareness. The default values are marked in red.
	}
	\label{fig:hyper}
\end{figure*}

Semantic hyperparameters $\omega_1$ and $\omega_2$ control the semantic-awareness in stage I and stage II, respectively. As shown in Fig.~\ref{fig:hyper}, when $\omega_1$ and $\omega_2$ decrease, the algorithm reverts to its semantically unaware version that ignores the semantic maps provided. Due to the different views used, different stages may concentrate on different perspectives to align with the semantic maps. Specifically, stage I ($1^{st}$ row) is more inclined to align with the semantic maps from a global perspective, e.g., adjusting the position of the portrait or rectifying some large structures in the clothing. By contrast, stage II ($2^{nd}$ row) can align more detailed structures, such as faces, with the semantic maps. By using stage I and stage II together ($3^{rd}$ row), the result can be more faithful to the semantic maps in both large structures and small details.

\subsection{Additional interactive texture transfer results}
\label{sec2}
\begin{figure*}[t]
	\centering
	\setlength{\tabcolsep}{0.05cm}
	\begin{tabular}{cccc}
		\includegraphics[width=0.215\linewidth]{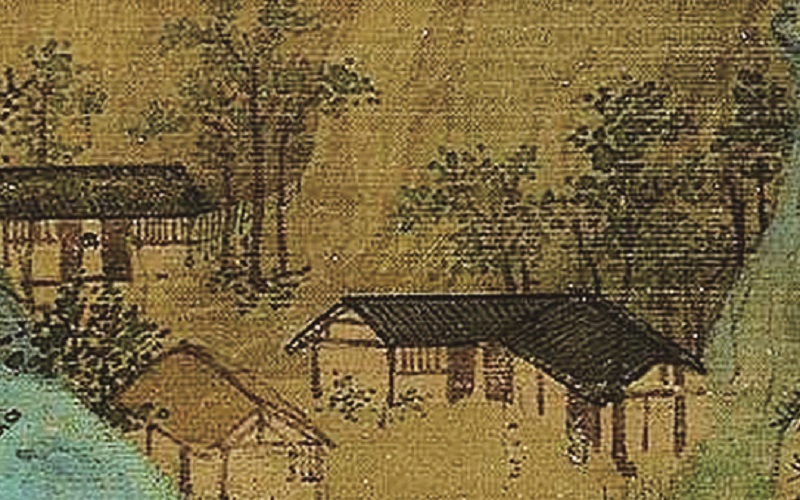}&
		\includegraphics[width=0.215\linewidth]{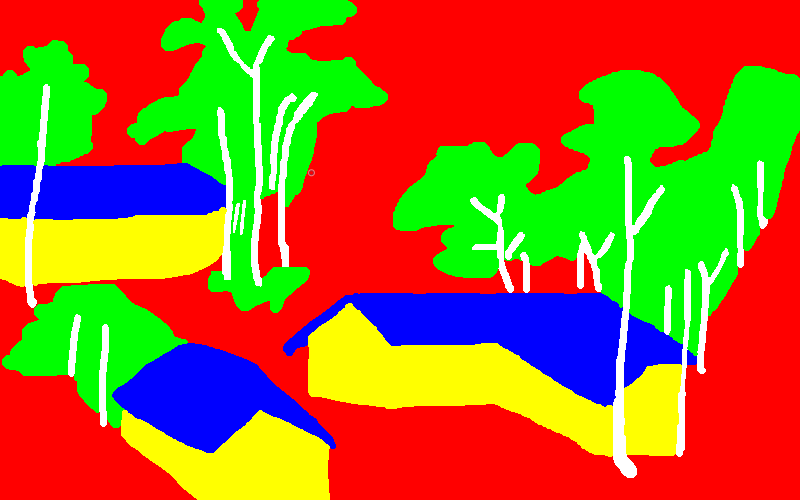}&
		\includegraphics[width=0.215\linewidth]{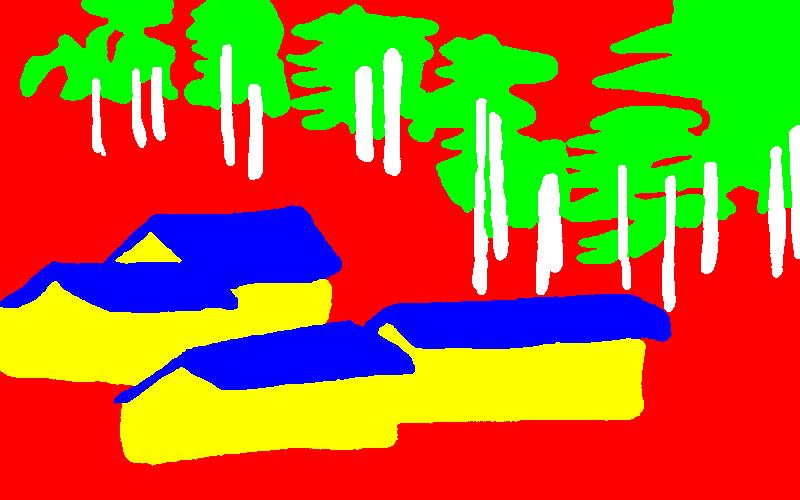}&
		\includegraphics[width=0.215\linewidth]{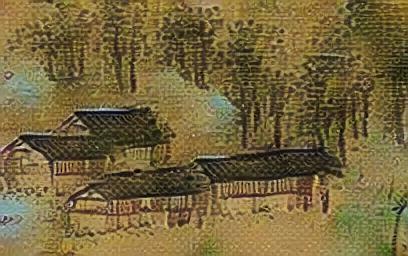}
		\\
		\includegraphics[width=0.215\linewidth]{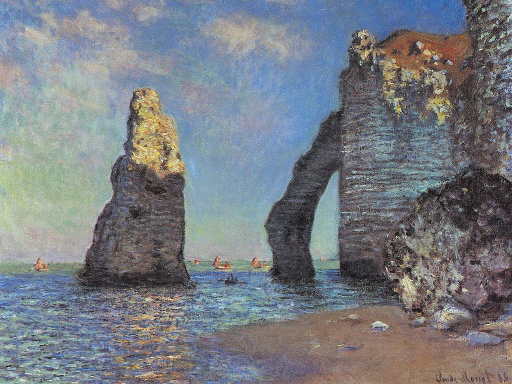}&
		\includegraphics[width=0.215\linewidth]{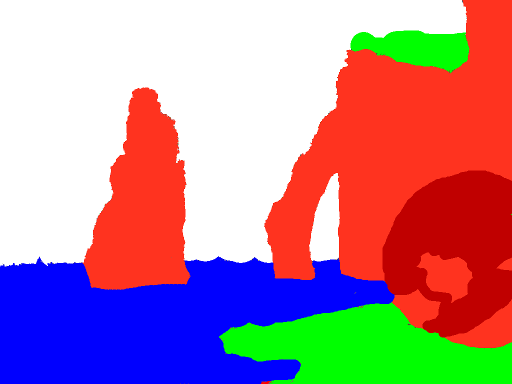}&
		\includegraphics[width=0.215\linewidth]{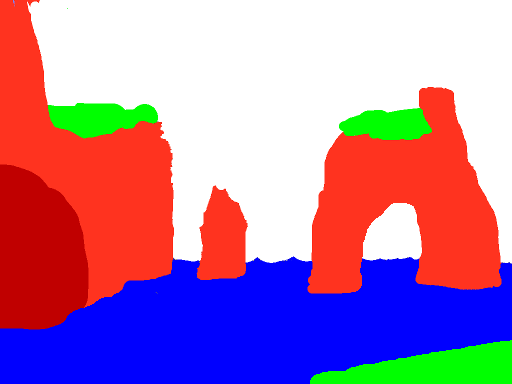}&
		\includegraphics[width=0.215\linewidth]{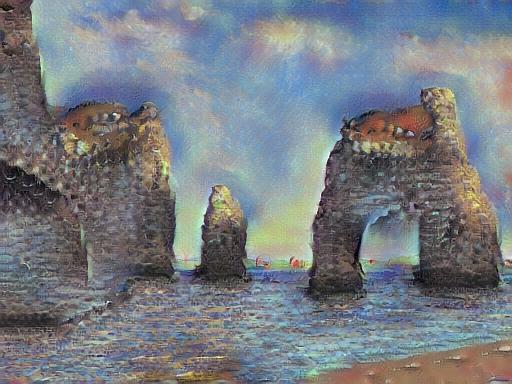}
		\\
		\includegraphics[width=0.215\linewidth]{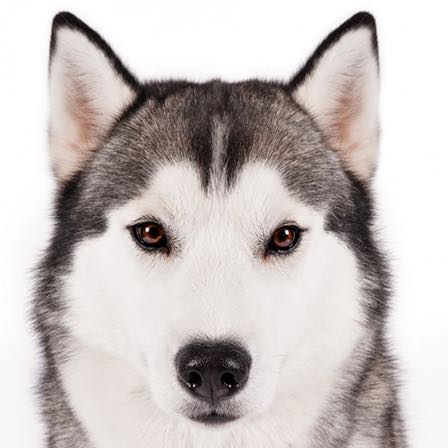}&
		\includegraphics[width=0.215\linewidth]{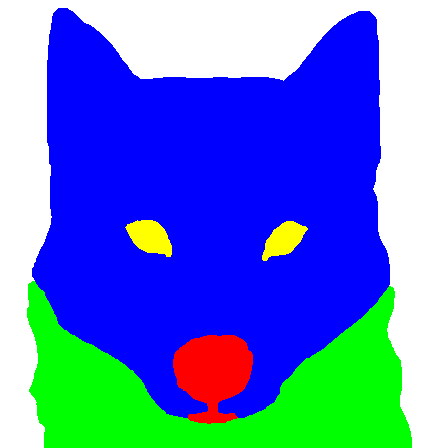}&
		\includegraphics[width=0.215\linewidth]{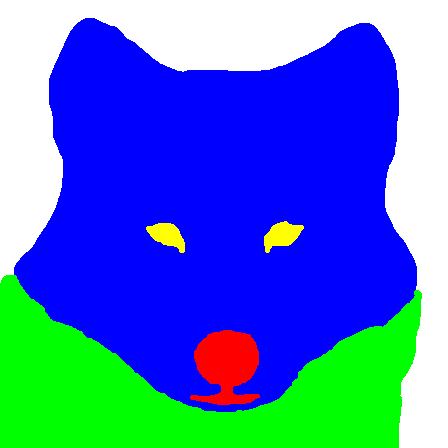}&
		\includegraphics[width=0.215\linewidth]{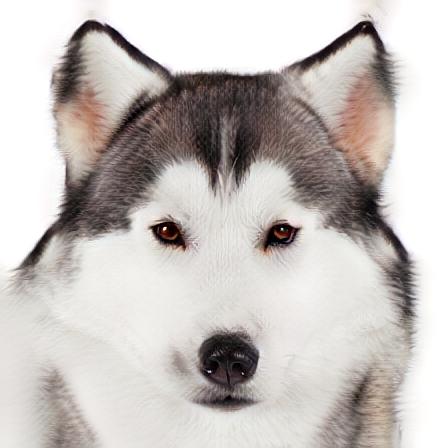}
		\\
		\includegraphics[width=0.215\linewidth]{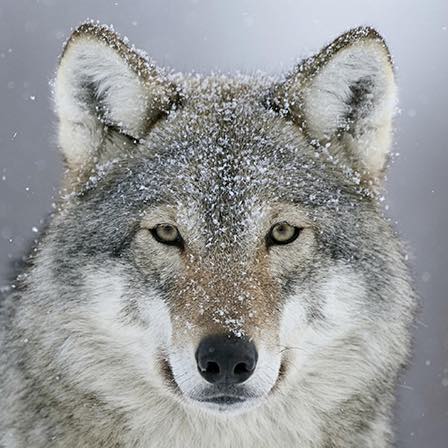}&
		\includegraphics[width=0.215\linewidth]{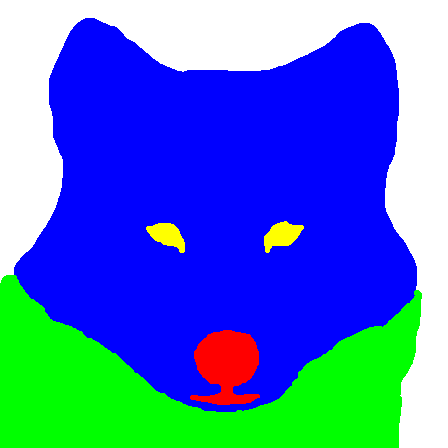}&
		\includegraphics[width=0.215\linewidth]{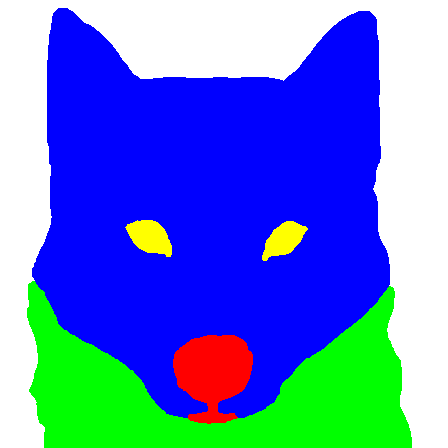}&
		\includegraphics[width=0.215\linewidth]{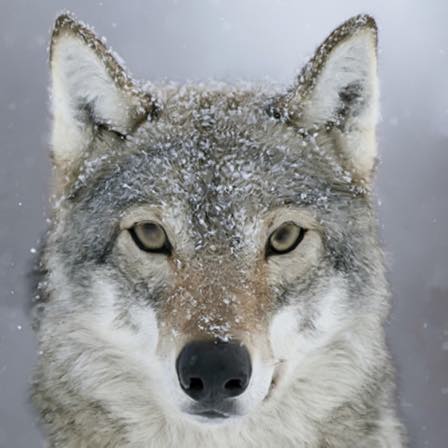}
		\\
		\includegraphics[width=0.215\linewidth]{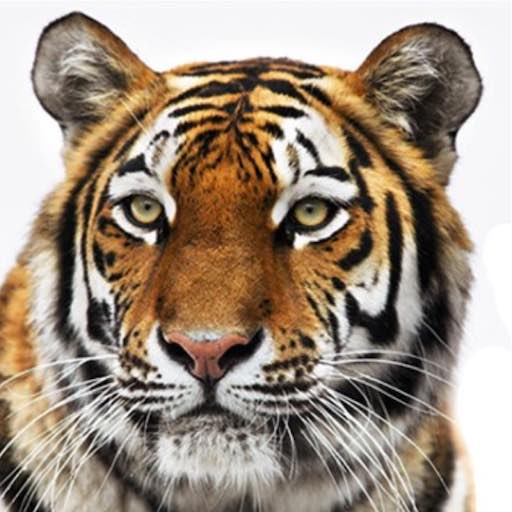}&
		\includegraphics[width=0.215\linewidth]{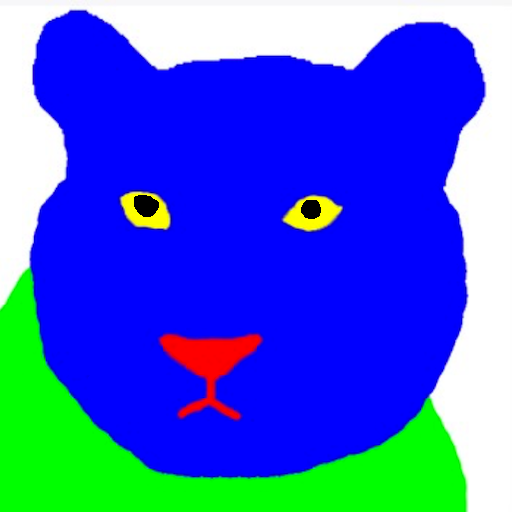}&
		\includegraphics[width=0.215\linewidth]{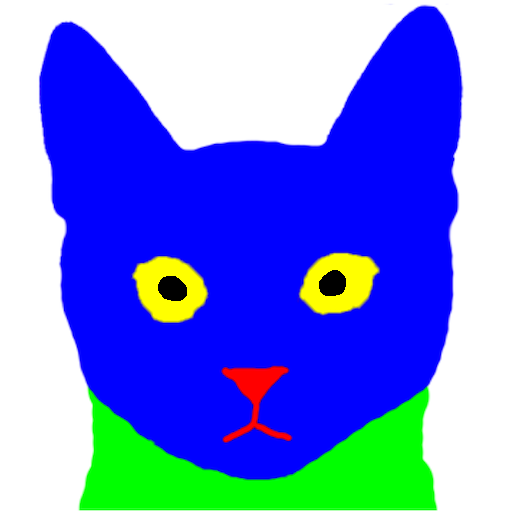}&
		\includegraphics[width=0.215\linewidth]{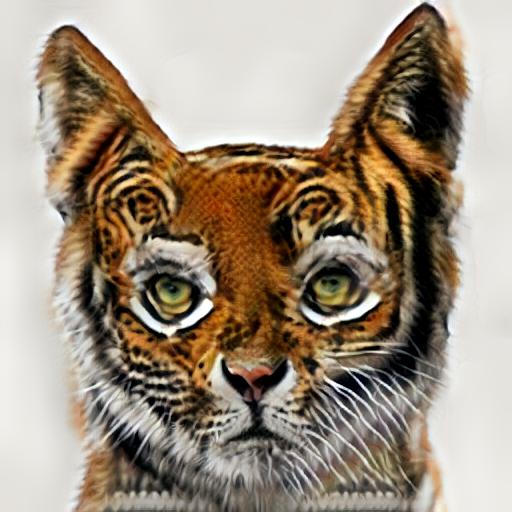}
		\\
		Input (source) & Input (source semantic) & Input (target semantic) & Output (target) 
		
	\end{tabular}
	\vspace{1em}
	\caption{Doodles-to-artworks transfer results. Image courtesy of \cite{wang2020glstylenet,champandard2016semantic,liao2017visual}.
	}
	\label{fig:doodle1}
\end{figure*}

\begin{figure*}[b]
	\centering
	\setlength{\tabcolsep}{0.05cm}
	\begin{tabular}{cccccc}
		&
		\includegraphics[width=0.145\linewidth]{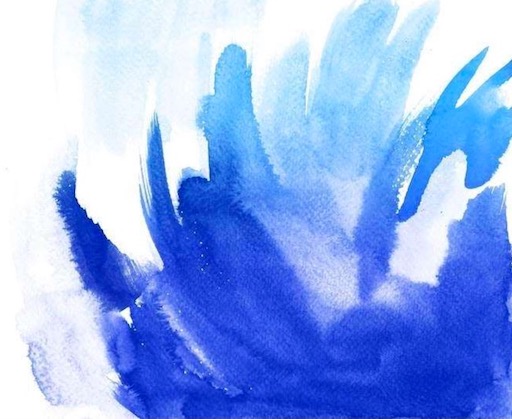}&
		\includegraphics[width=0.155\linewidth]{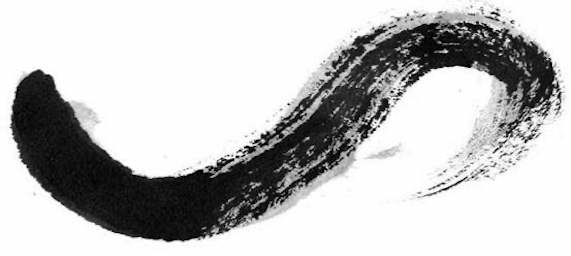}&
		\includegraphics[width=0.155\linewidth]{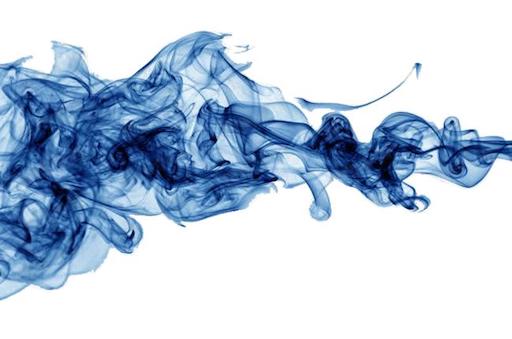}&
		\includegraphics[width=0.155\linewidth]{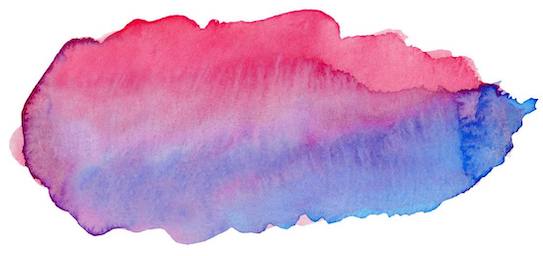}&
		\includegraphics[width=0.125\linewidth]{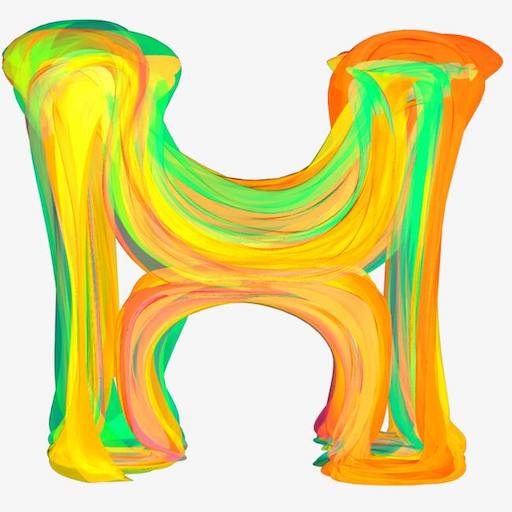}
		\\
		\\
		& Input (source 1) & Input (source 2) & Input (source 3) & Input (source 4) & Input (source 5)
		\\
		\\
		\hline
		\\
		Input (paths)& Output 1 & Output 2 & Output 3 & Output 4 &Output 5
		\\
		\\
		
		\includegraphics[width=0.155\linewidth]{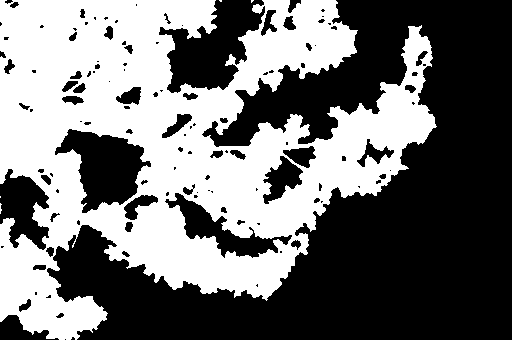}&
		\includegraphics[width=0.155\linewidth]{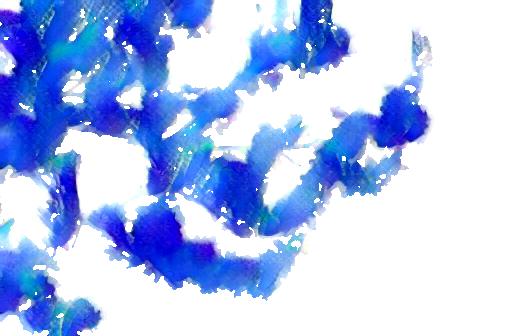}&
		\includegraphics[width=0.155\linewidth]{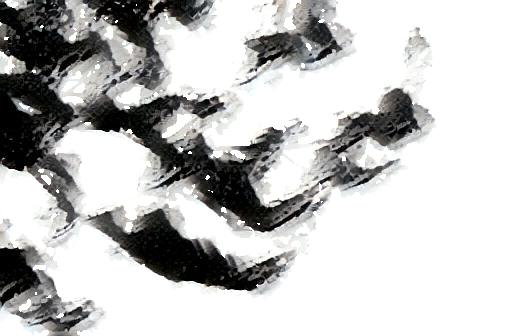}&
		\includegraphics[width=0.155\linewidth]{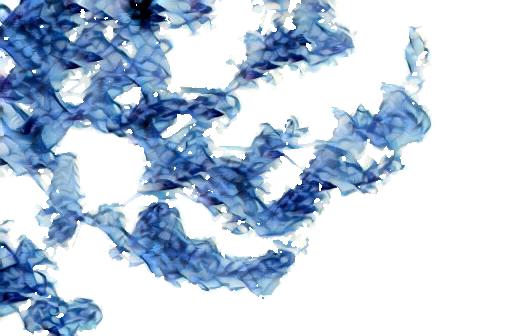}&
		\includegraphics[width=0.155\linewidth]{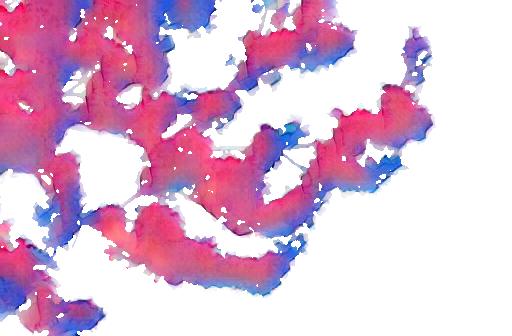}&
		\includegraphics[width=0.155\linewidth]{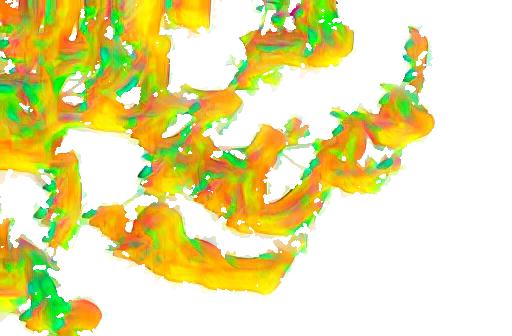}
		
		\\
		\includegraphics[width=0.155\linewidth]{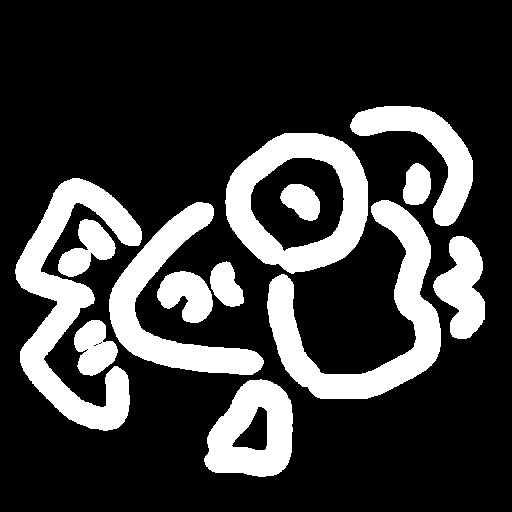}&
		\includegraphics[width=0.155\linewidth]{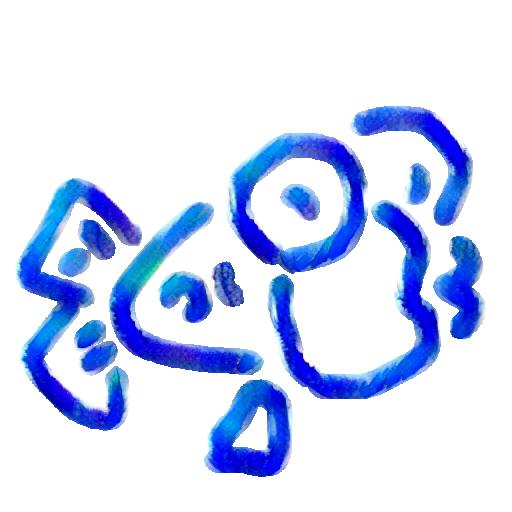}&
		\includegraphics[width=0.155\linewidth]{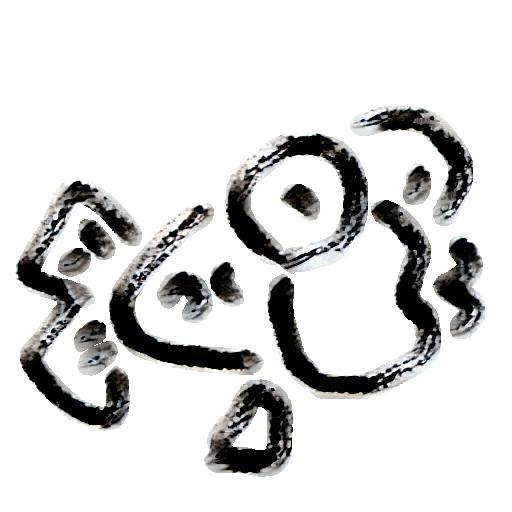}&
		\includegraphics[width=0.155\linewidth]{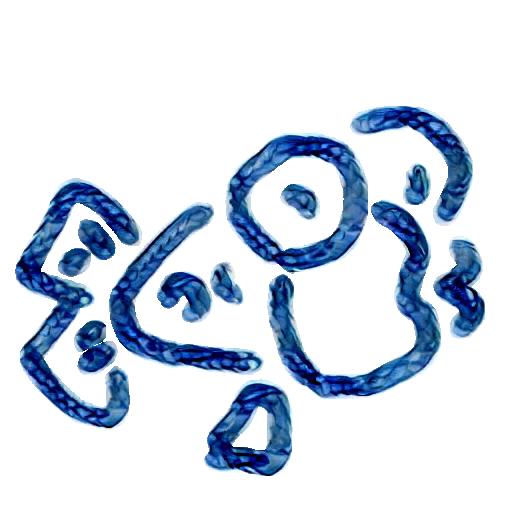}&
		\includegraphics[width=0.155\linewidth]{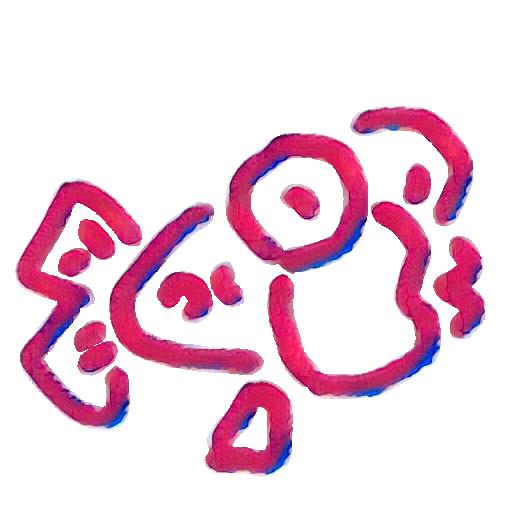}&
		\includegraphics[width=0.155\linewidth]{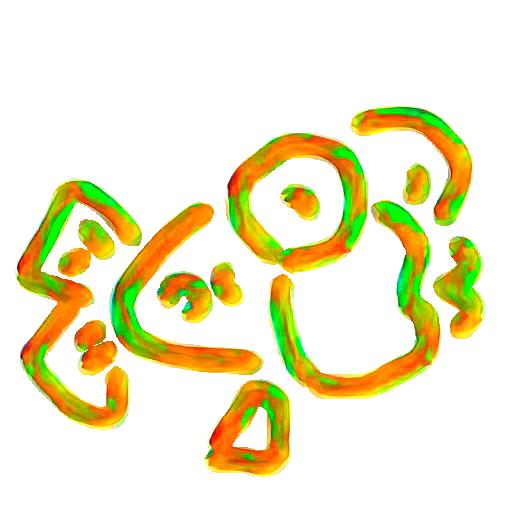}
		
		\\
		\includegraphics[width=0.155\linewidth]{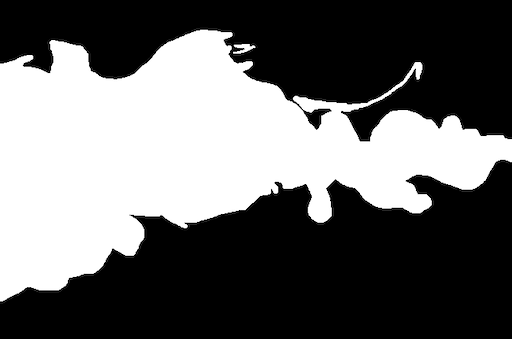}&
		\includegraphics[width=0.155\linewidth]{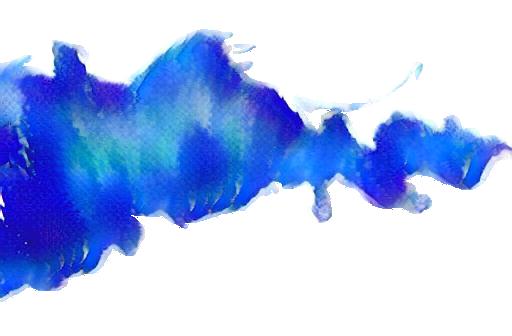}&
		\includegraphics[width=0.155\linewidth]{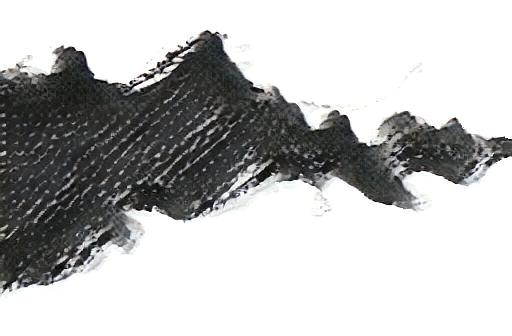}&
		\includegraphics[width=0.155\linewidth]{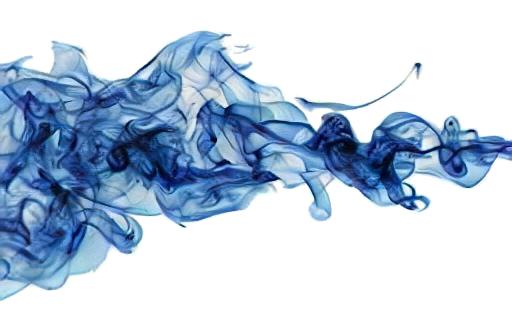}&
		\includegraphics[width=0.155\linewidth]{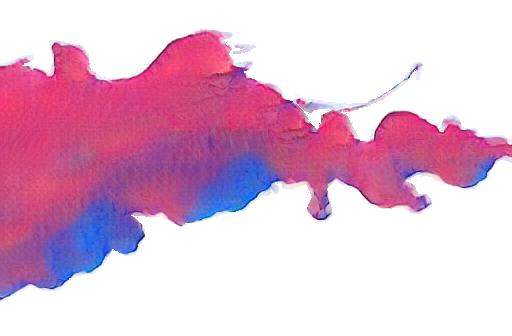}&
		\includegraphics[width=0.155\linewidth]{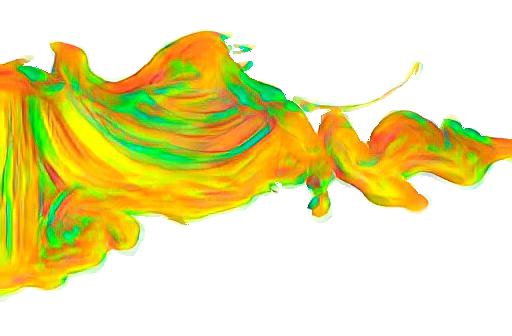}
		
		\\
		\includegraphics[width=0.155\linewidth]{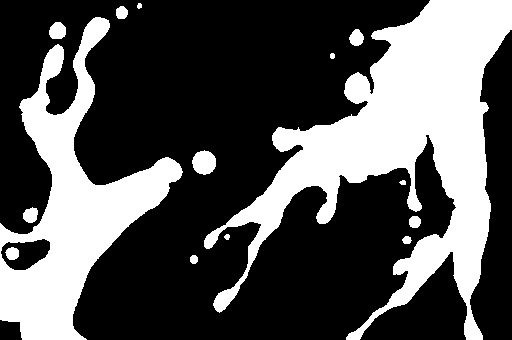}&
		\includegraphics[width=0.155\linewidth]{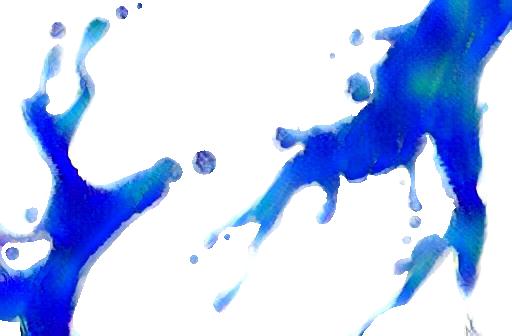}&
		\includegraphics[width=0.155\linewidth]{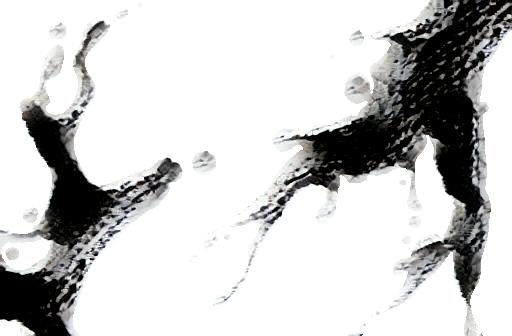}&
		\includegraphics[width=0.155\linewidth]{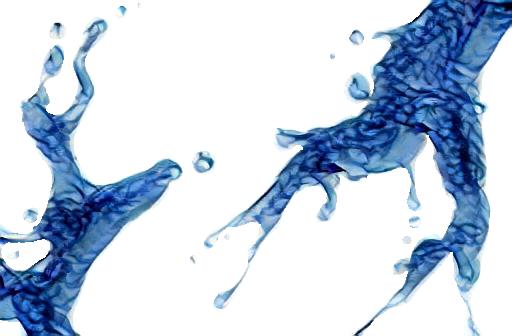}&
		\includegraphics[width=0.155\linewidth]{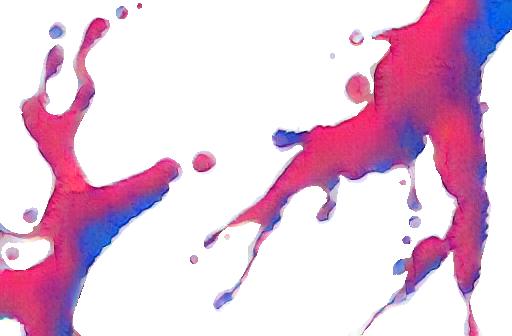}&
		\includegraphics[width=0.155\linewidth]{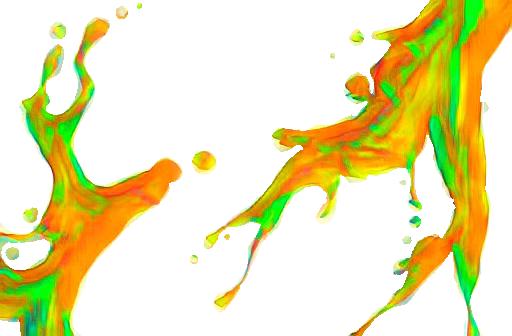}
		
		\\
		\includegraphics[width=0.155\linewidth]{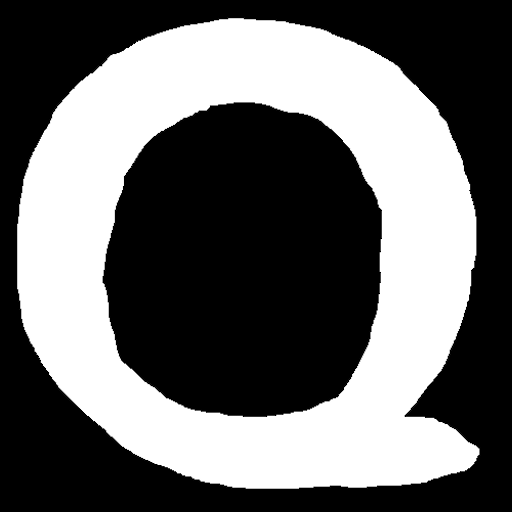}&
		\includegraphics[width=0.155\linewidth]{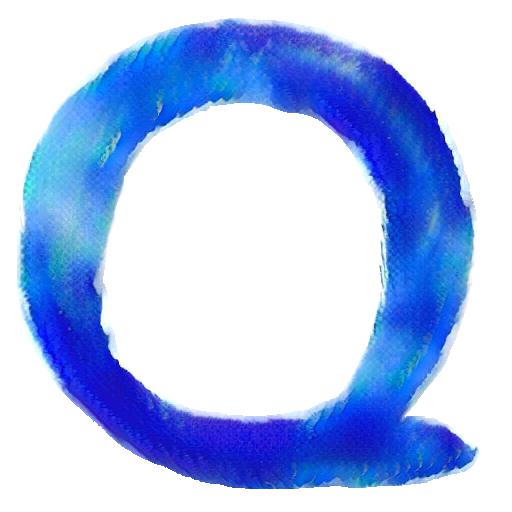}&
		\includegraphics[width=0.155\linewidth]{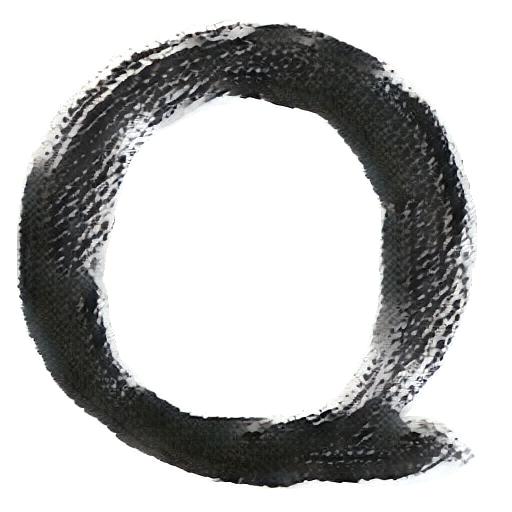}&
		\includegraphics[width=0.155\linewidth]{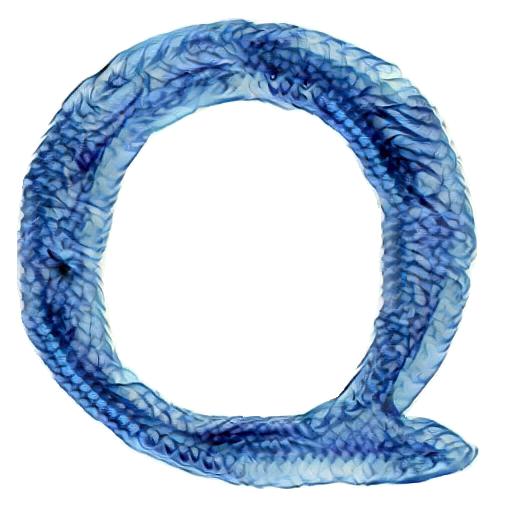}&
		\includegraphics[width=0.155\linewidth]{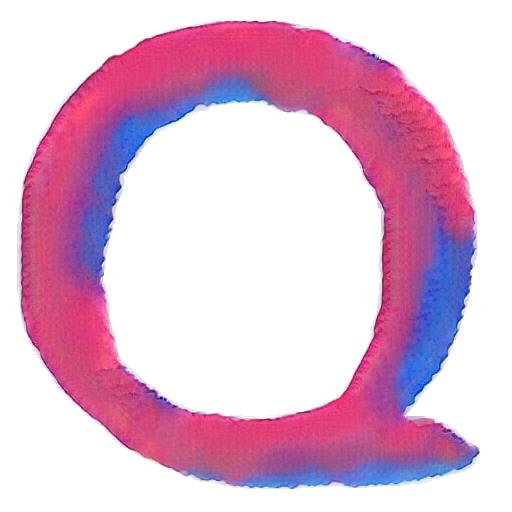}&
		\includegraphics[width=0.155\linewidth]{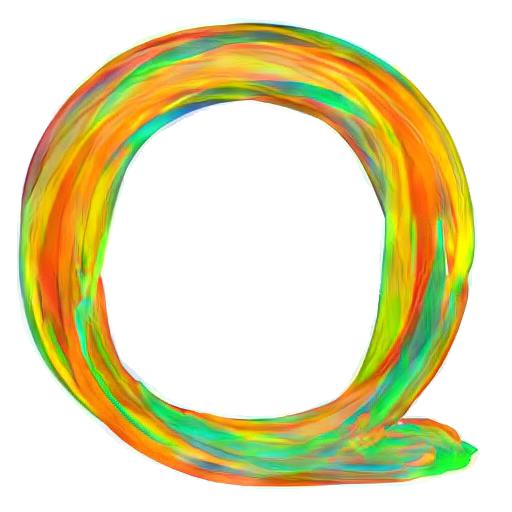}
		
		\\
		\includegraphics[width=0.155\linewidth]{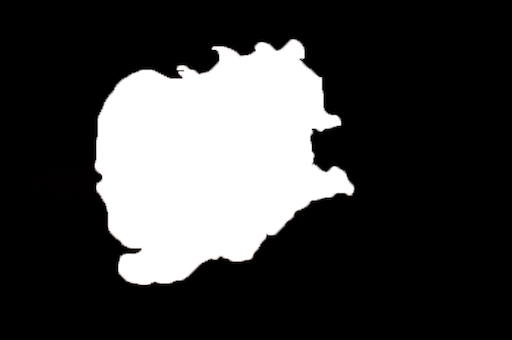}&
		\includegraphics[width=0.155\linewidth]{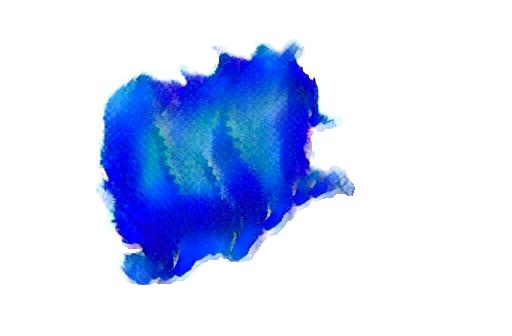}&
		\includegraphics[width=0.155\linewidth]{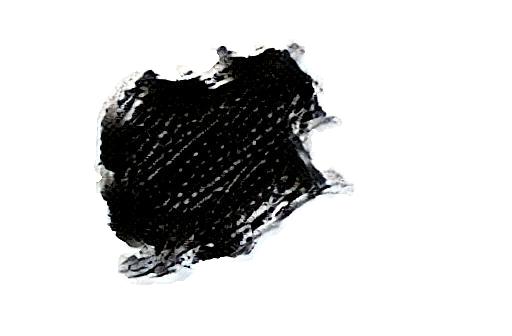}&
		\includegraphics[width=0.155\linewidth]{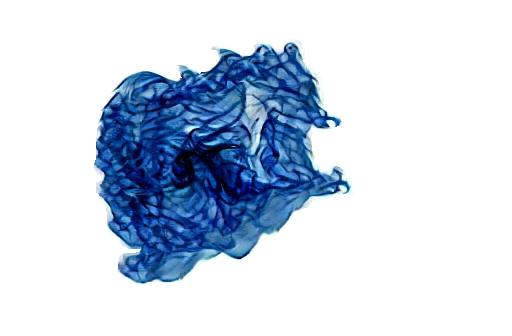}&
		\includegraphics[width=0.155\linewidth]{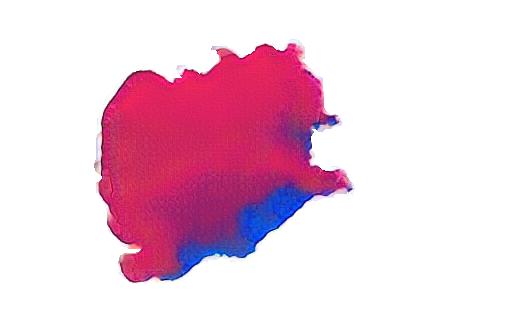}&
		\includegraphics[width=0.155\linewidth]{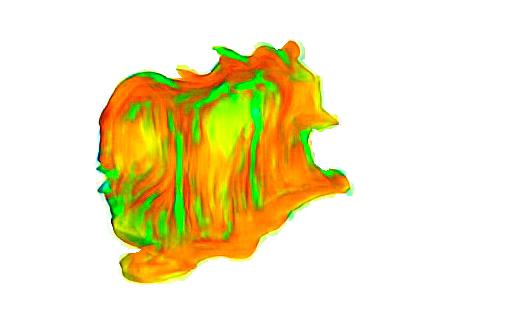}
		
		\\
		\includegraphics[width=0.155\linewidth]{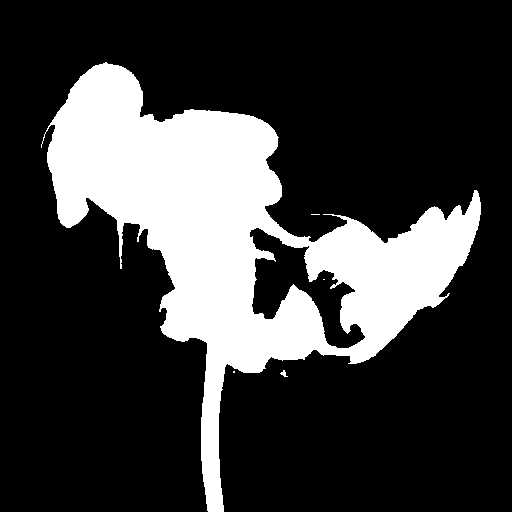}&
		\includegraphics[width=0.155\linewidth]{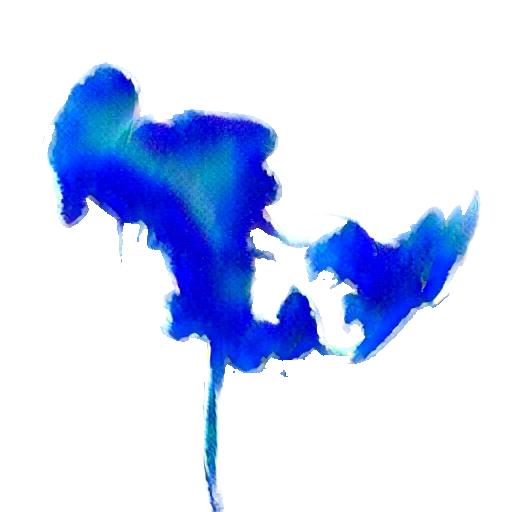}&
		\includegraphics[width=0.155\linewidth]{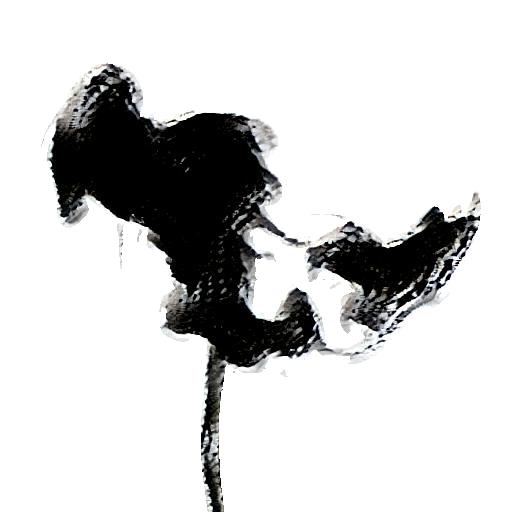}&
		\includegraphics[width=0.155\linewidth]{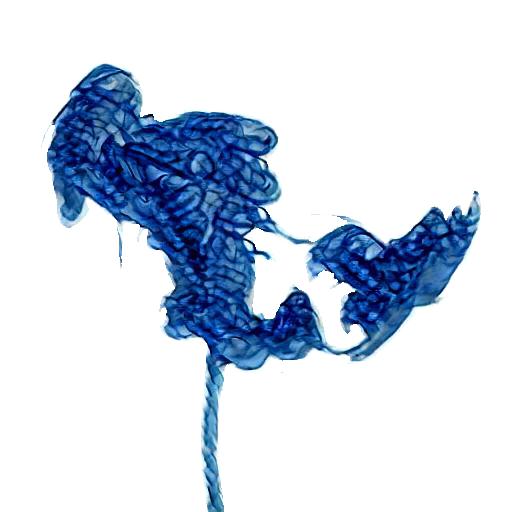}&
		\includegraphics[width=0.155\linewidth]{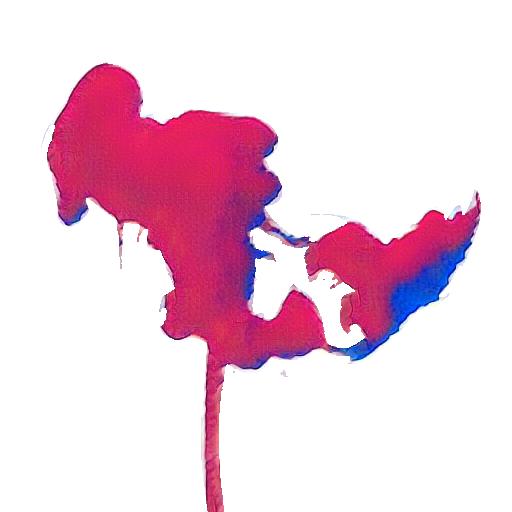}&
		\includegraphics[width=0.155\linewidth]{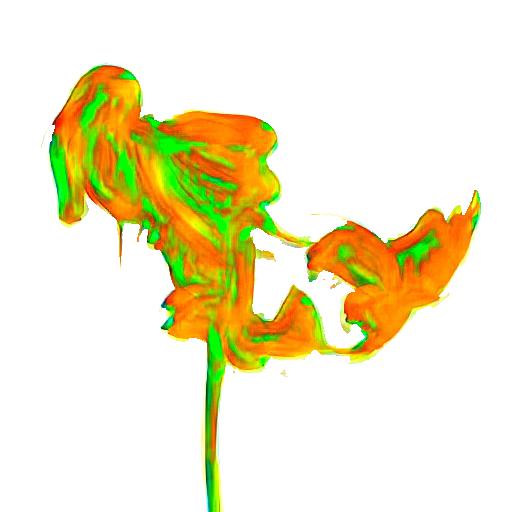}

	\end{tabular}
	\vspace{1em}
	\caption{Texture pattern editing results. Image courtesy of \cite{men2018common,yang2019controllable,luan2017deep}.
	}
	\label{fig:pattern1}
\end{figure*}

\begin{figure*}[t]
	\centering
	\setlength{\tabcolsep}{0.2cm}
	\begin{tabular}{ccc}
		Input (source) & Input (plain text) & Output (target) 
		\\
		\\
		\includegraphics[width=0.305\linewidth]{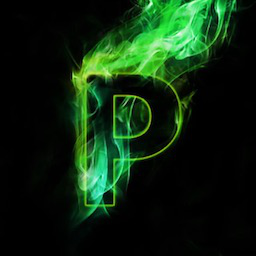}&
		\includegraphics[width=0.305\linewidth]{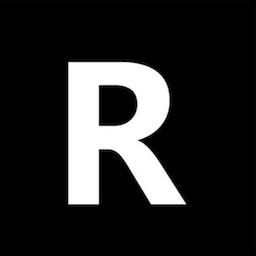}&
		\includegraphics[width=0.305\linewidth]{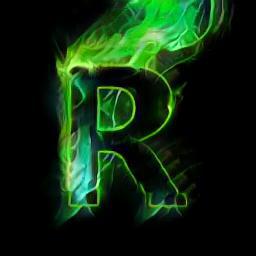}
		\\
		&
		\includegraphics[width=0.305\linewidth]{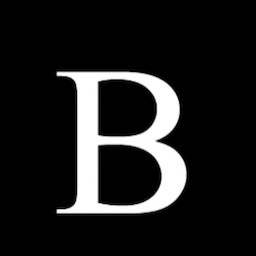}&
		\includegraphics[width=0.305\linewidth]{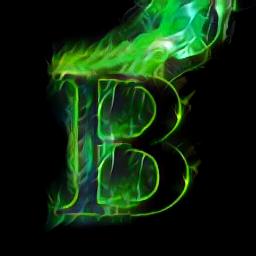}
		\\
		&
		\includegraphics[width=0.305\linewidth]{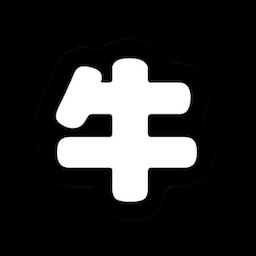}&
		\includegraphics[width=0.305\linewidth]{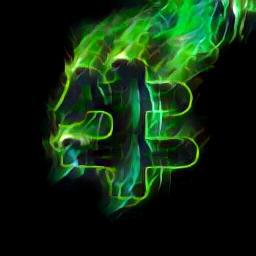}

	\end{tabular}
	\vspace{1em}
	\caption{Text effects transfer results. Image courtesy of \cite{yang2017awesome}.
	}
	\label{fig:text1}
\end{figure*}

\begin{figure*}[t]
	\centering
	\setlength{\tabcolsep}{0.2cm}
	\begin{tabular}{ccc}
		Input (source) & Input (plain text) & Output (target) 
		\\
		\\
		\includegraphics[width=0.305\linewidth]{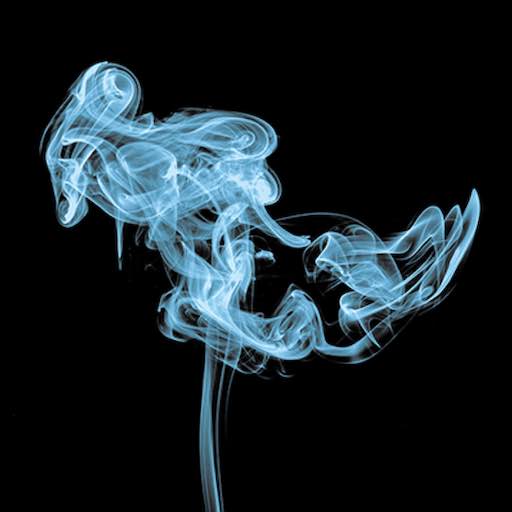}&
		\includegraphics[width=0.305\linewidth]{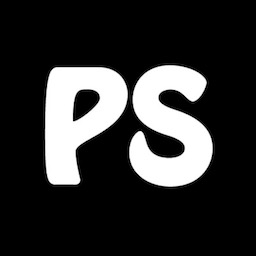}&
		\includegraphics[width=0.305\linewidth]{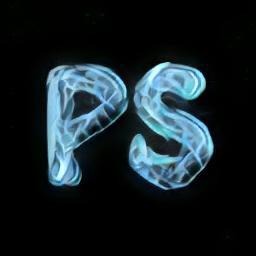}
		\\
		&
		\includegraphics[width=0.305\linewidth]{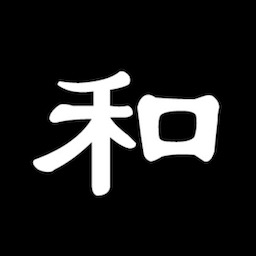}&
		\includegraphics[width=0.305\linewidth]{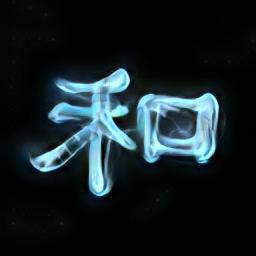}
		\\
		&
		\includegraphics[width=0.305\linewidth]{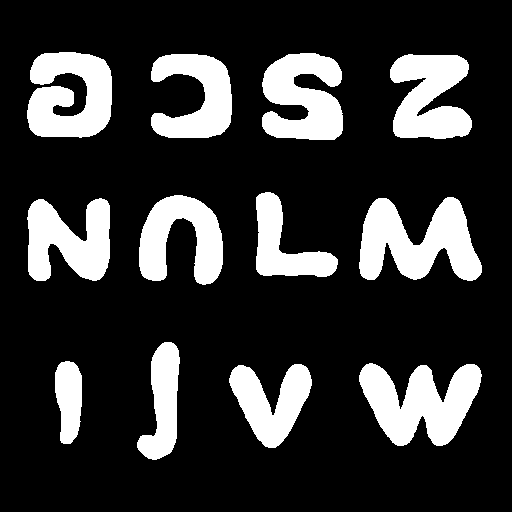}&
		\includegraphics[width=0.305\linewidth]{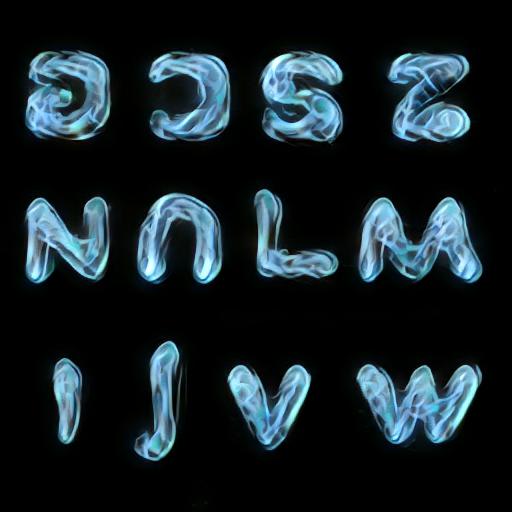}

	\end{tabular}
	\vspace{1em}
	\caption{Text effects transfer results. Image courtesy of \cite{yang2017awesome,yang2019controllable,lu2014decobrush}.
	}
	\label{fig:text2}
\end{figure*}

\begin{figure*}[b]
	\centering
	\setlength{\tabcolsep}{0.05cm}
	\begin{tabular}{cccccc}
		&
		\includegraphics[width=0.135\linewidth]{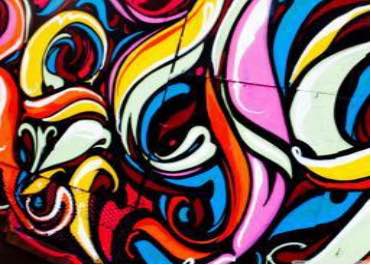}&
		\includegraphics[width=0.145\linewidth]{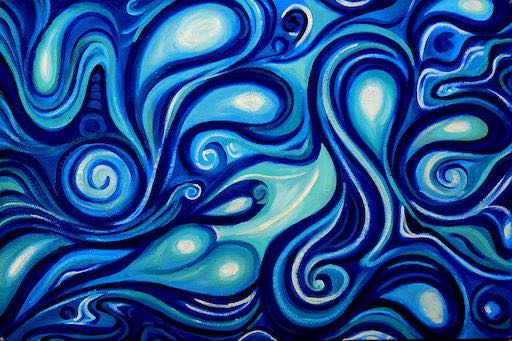}&
		\includegraphics[width=0.095\linewidth]{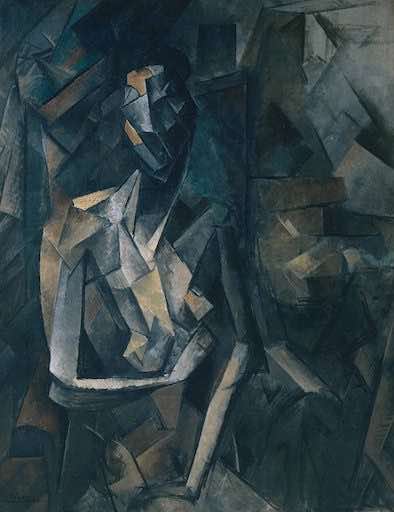}&
		\includegraphics[width=0.100\linewidth]{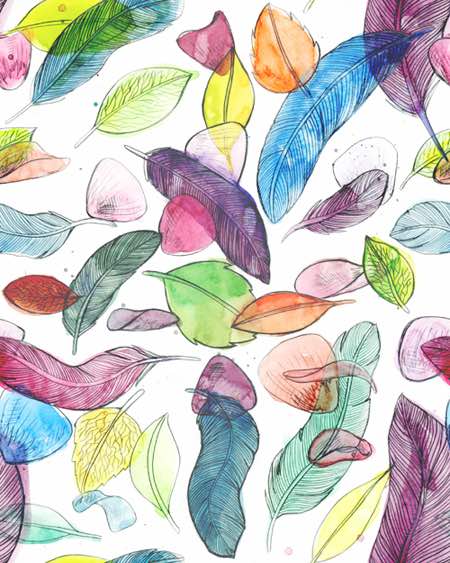}&
		\includegraphics[width=0.125\linewidth]{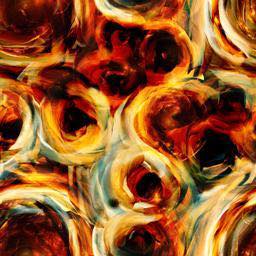}
		\\
		\\
		& Input (source 1) & Input (source 2) & Input (source 3) & Input (source 4) & Input (source 5)
		\\
		\\
		\hline
		\\
		Manipulated & Output 1 & Output 2 & Output 3 & Output 4 &Output 5
		\\
		
		\includegraphics[width=0.155\linewidth]{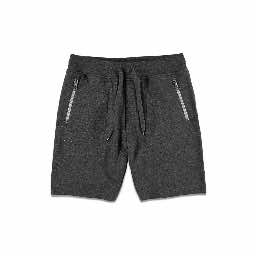}&
		\includegraphics[width=0.155\linewidth]{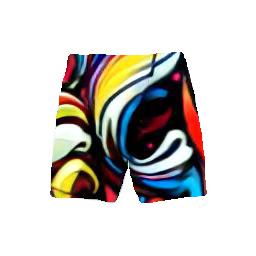}&
		\includegraphics[width=0.155\linewidth]{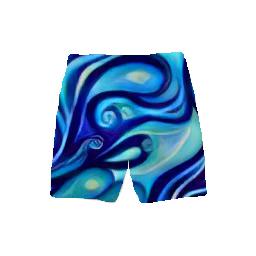}&
		\includegraphics[width=0.155\linewidth]{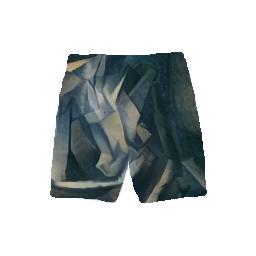}&
		\includegraphics[width=0.155\linewidth]{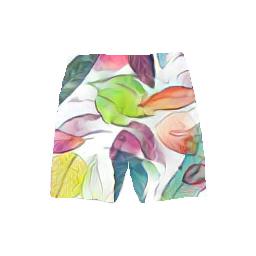}&
		\includegraphics[width=0.155\linewidth]{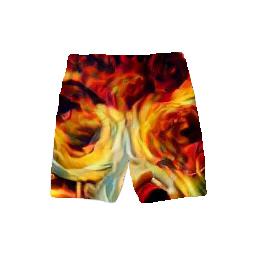}
		
		\\
		\includegraphics[width=0.155\linewidth]{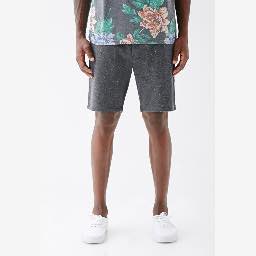}&
		\includegraphics[width=0.155\linewidth]{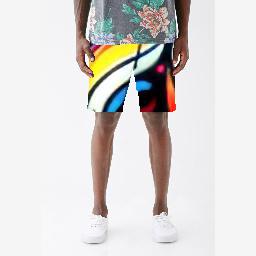}&
		\includegraphics[width=0.155\linewidth]{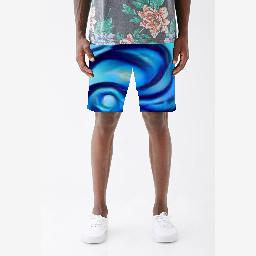}&
		\includegraphics[width=0.155\linewidth]{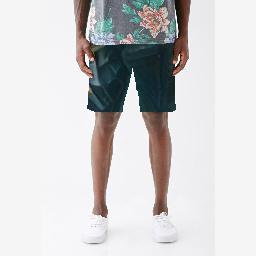}&
		\includegraphics[width=0.155\linewidth]{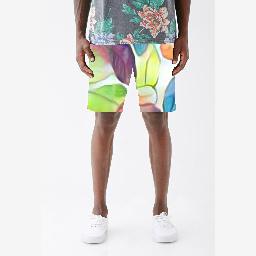}&
		\includegraphics[width=0.155\linewidth]{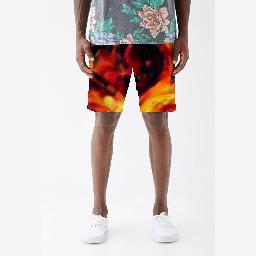}
		
		\\
		\includegraphics[width=0.155\linewidth]{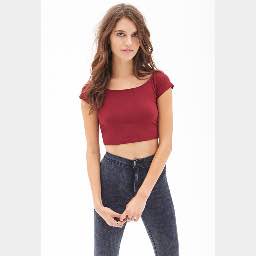}&
		\includegraphics[width=0.155\linewidth]{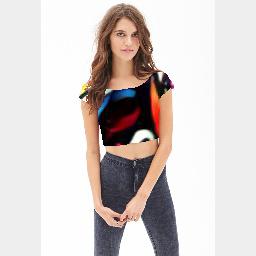}&
		\includegraphics[width=0.155\linewidth]{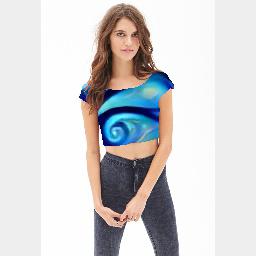}&
		\includegraphics[width=0.155\linewidth]{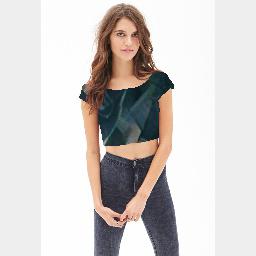}&
		\includegraphics[width=0.155\linewidth]{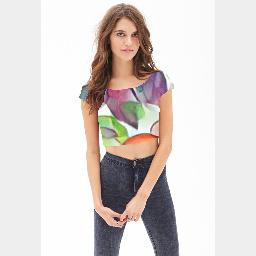}&
		\includegraphics[width=0.155\linewidth]{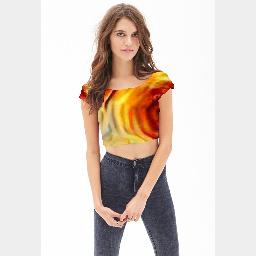}
		
		\\
		\includegraphics[width=0.155\linewidth]{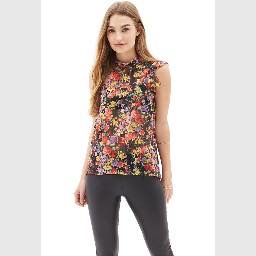}&
		\includegraphics[width=0.155\linewidth]{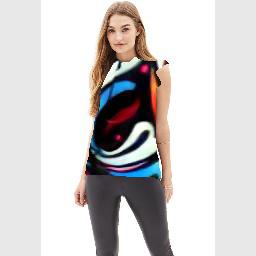}&
		\includegraphics[width=0.155\linewidth]{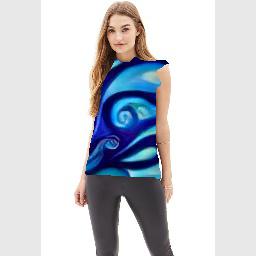}&
		\includegraphics[width=0.155\linewidth]{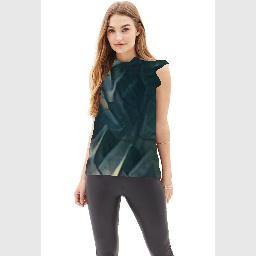}&
		\includegraphics[width=0.155\linewidth]{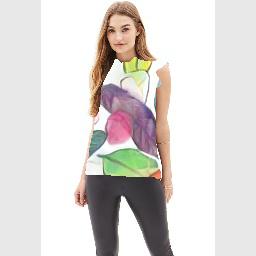}&
		\includegraphics[width=0.155\linewidth]{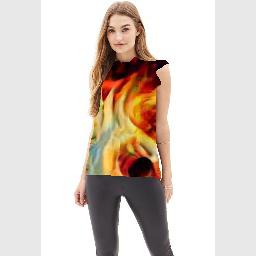}
		
		\\
		\includegraphics[width=0.155\linewidth]{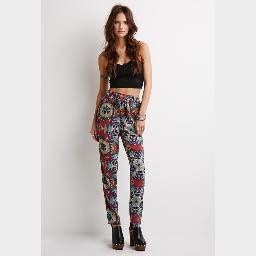}&
		\includegraphics[width=0.155\linewidth]{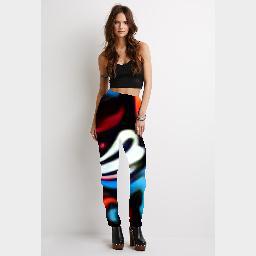}&
		\includegraphics[width=0.155\linewidth]{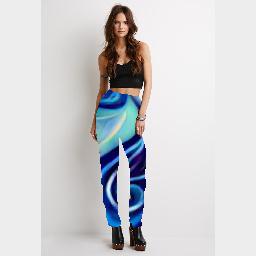}&
		\includegraphics[width=0.155\linewidth]{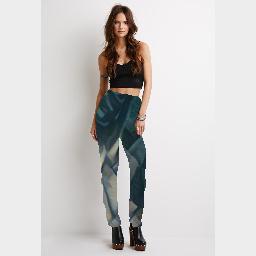}&
		\includegraphics[width=0.155\linewidth]{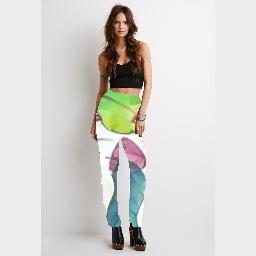}&
		\includegraphics[width=0.155\linewidth]{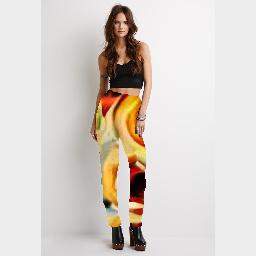}
		
		\\
		\includegraphics[width=0.155\linewidth]{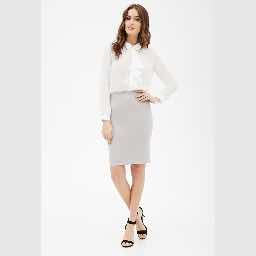}&
		\includegraphics[width=0.155\linewidth]{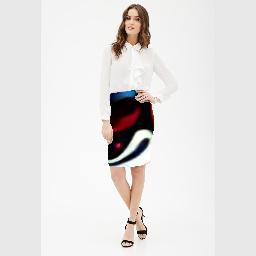}&
		\includegraphics[width=0.155\linewidth]{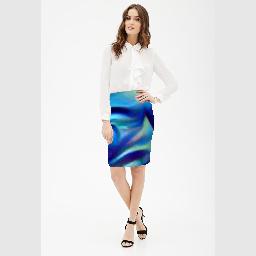}&
		\includegraphics[width=0.155\linewidth]{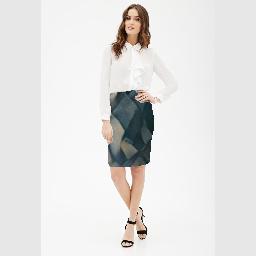}&
		\includegraphics[width=0.155\linewidth]{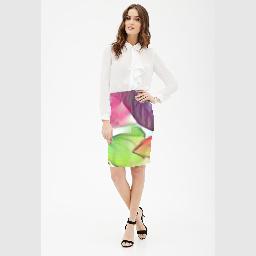}&
		\includegraphics[width=0.155\linewidth]{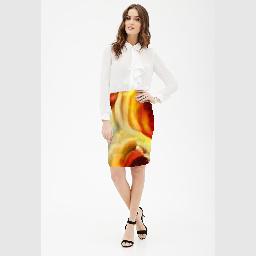}

	\end{tabular}
	\vspace{1em}
	\caption{Virtual clothing manipulation results. Image courtesy of \cite{liu2016deepfashion}.
	}
	\label{fig:cloth1}
\end{figure*}

\begin{figure*}[b]
	\centering
	\setlength{\tabcolsep}{0.05cm}
	\begin{tabular}{cccccc}
		&
		\includegraphics[width=0.156\linewidth]{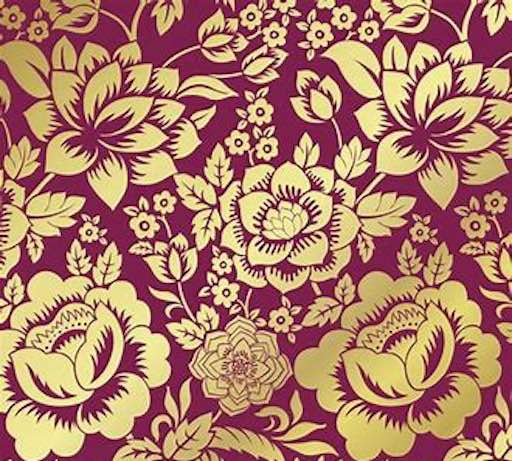}&
		\includegraphics[width=0.155\linewidth]{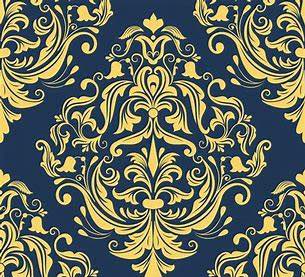}&
		\includegraphics[width=0.141\linewidth]{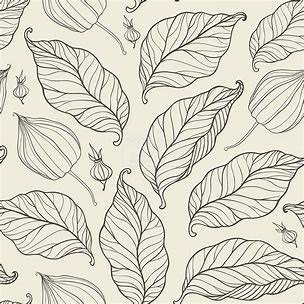}&
		\includegraphics[width=0.145\linewidth]{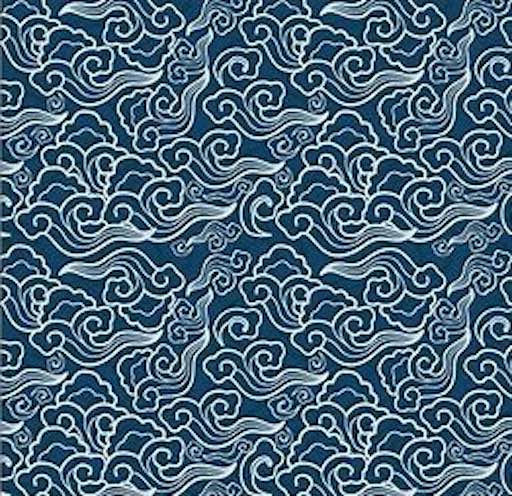}&
		\includegraphics[width=0.141\linewidth]{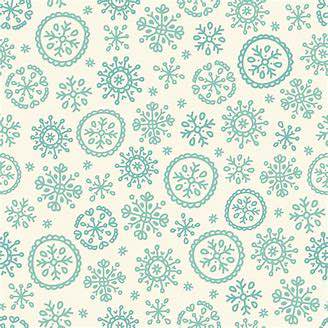}
		\\
		\\
		& Input (source 1) & Input (source 2) & Input (source 3) & Input (source 4) & Input (source 5)
		\\
		\\
		\hline
		\\
		Manipulated & Output 1 & Output 2 & Output 3 & Output 4 &Output 5
		\\
		
		\includegraphics[width=0.155\linewidth]{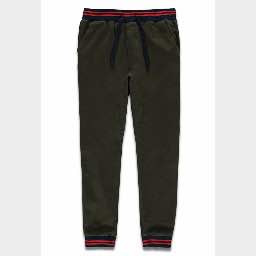}&
		\includegraphics[width=0.155\linewidth]{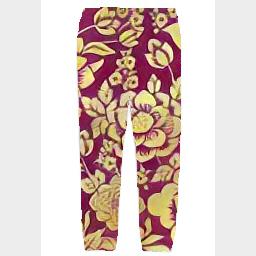}&
		\includegraphics[width=0.155\linewidth]{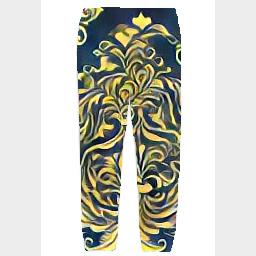}&
		\includegraphics[width=0.155\linewidth]{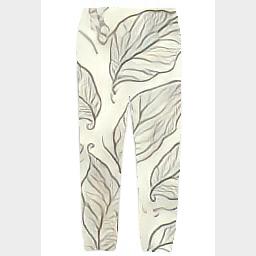}&
		\includegraphics[width=0.155\linewidth]{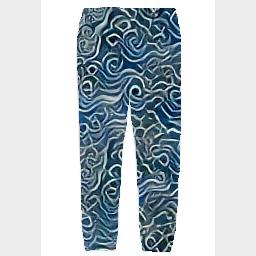}&
		\includegraphics[width=0.155\linewidth]{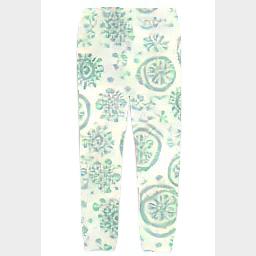}
		
		\\
		\includegraphics[width=0.155\linewidth]{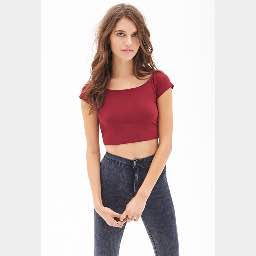}&
		\includegraphics[width=0.155\linewidth]{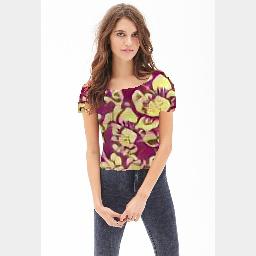}&
		\includegraphics[width=0.155\linewidth]{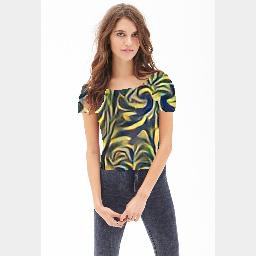}&
		\includegraphics[width=0.155\linewidth]{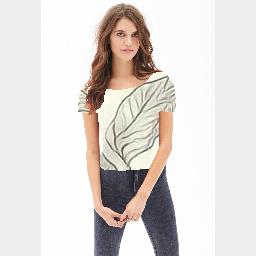}&
		\includegraphics[width=0.155\linewidth]{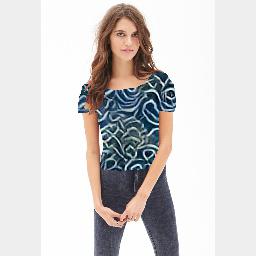}&
		\includegraphics[width=0.155\linewidth]{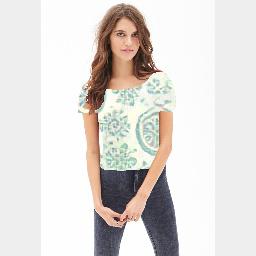}
		
		\\
		\includegraphics[width=0.155\linewidth]{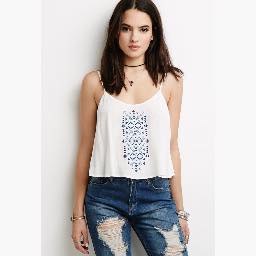}&
		\includegraphics[width=0.155\linewidth]{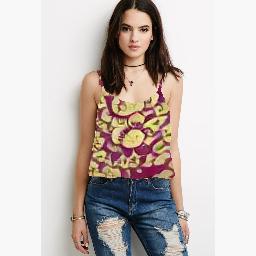}&
		\includegraphics[width=0.155\linewidth]{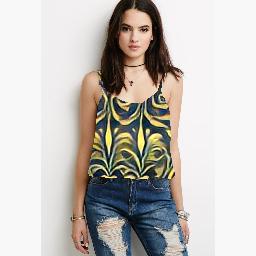}&
		\includegraphics[width=0.155\linewidth]{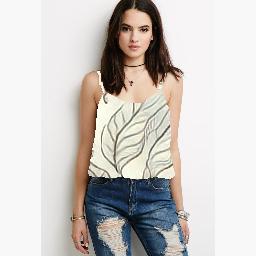}&
		\includegraphics[width=0.155\linewidth]{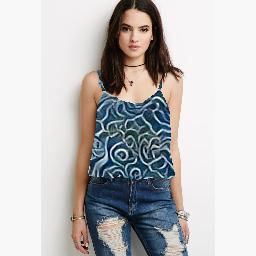}&
		\includegraphics[width=0.155\linewidth]{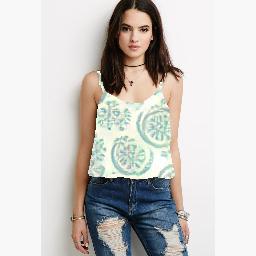}
		
		\\
		\includegraphics[width=0.155\linewidth]{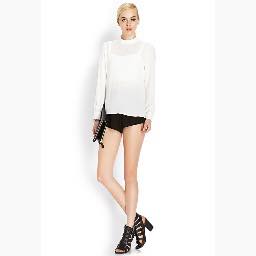}&
		\includegraphics[width=0.155\linewidth]{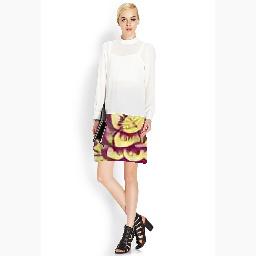}&
		\includegraphics[width=0.155\linewidth]{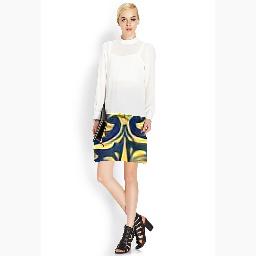}&
		\includegraphics[width=0.155\linewidth]{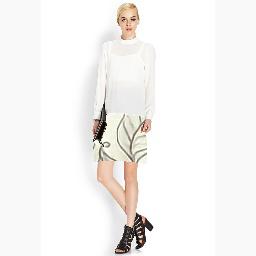}&
		\includegraphics[width=0.155\linewidth]{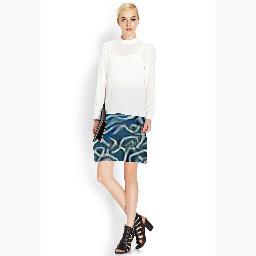}&
		\includegraphics[width=0.155\linewidth]{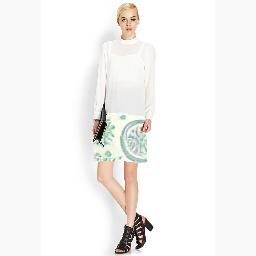}
		
		\\
		\includegraphics[width=0.155\linewidth]{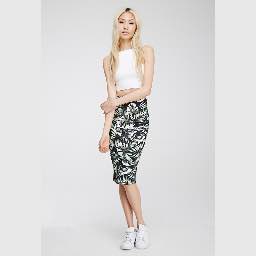}&
		\includegraphics[width=0.155\linewidth]{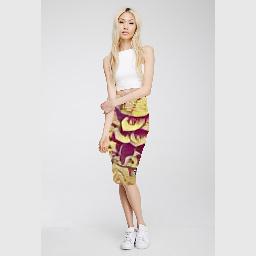}&
		\includegraphics[width=0.155\linewidth]{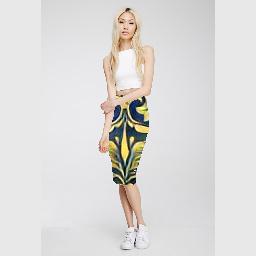}&
		\includegraphics[width=0.155\linewidth]{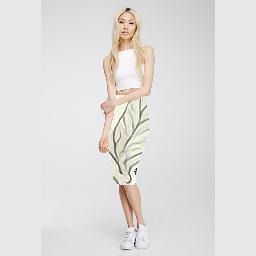}&
		\includegraphics[width=0.155\linewidth]{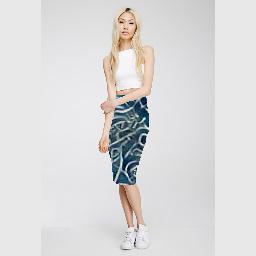}&
		\includegraphics[width=0.155\linewidth]{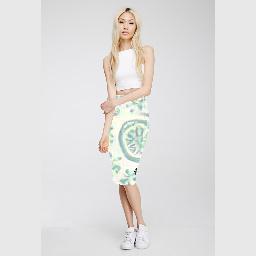}
		
		\\
		\includegraphics[width=0.155\linewidth]{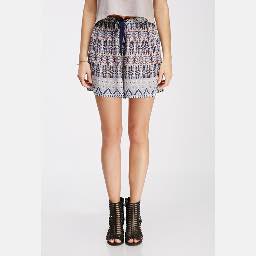}&
		\includegraphics[width=0.155\linewidth]{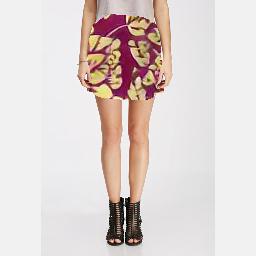}&
		\includegraphics[width=0.155\linewidth]{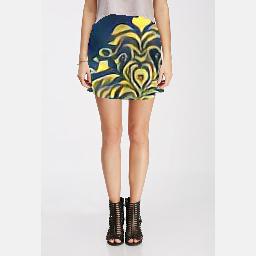}&
		\includegraphics[width=0.155\linewidth]{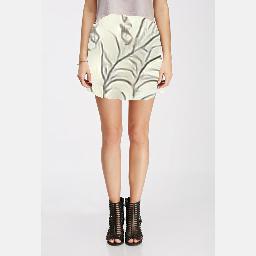}&
		\includegraphics[width=0.155\linewidth]{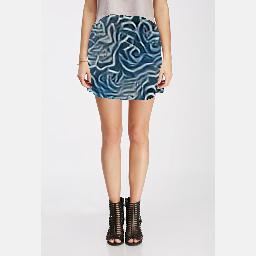}&
		\includegraphics[width=0.155\linewidth]{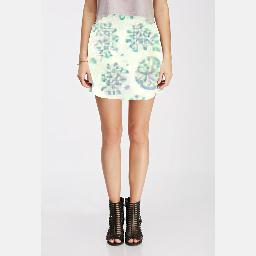}

	\end{tabular}
	\vspace{1em}
	\caption{Virtual clothing manipulation results. Image courtesy of \cite{liu2016deepfashion}.
	}
	\label{fig:cloth2}
\end{figure*}

In this part, we show some additional results generated by the proposed framework in different interactive texture transfer applications: doodles-to-artworks (Fig.~\ref{fig:doodle1}), texture pattern editing (Fig.~\ref{fig:pattern1}), text effects transfer (Fig.~\ref{fig:text1},~\ref{fig:text2}), and virtual clothing manipulation (Fig.~\ref{fig:cloth1},~\ref{fig:cloth2}).

\subsection{Additional comparisons with state-of-the-art algorithms}
\label{sec3}

\begin{figure*}[h]
	\centering
	\setlength{\tabcolsep}{0.2cm}
	\begin{tabular}{ccc}
		\includegraphics[width=0.300\linewidth]{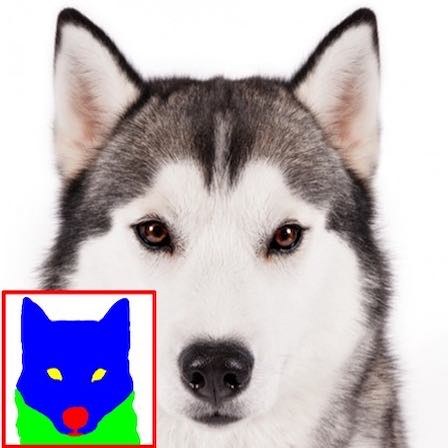}&
		\includegraphics[width=0.300\linewidth]{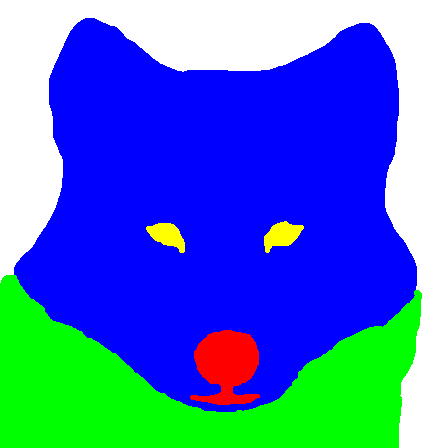}&
		\includegraphics[width=0.300\linewidth]{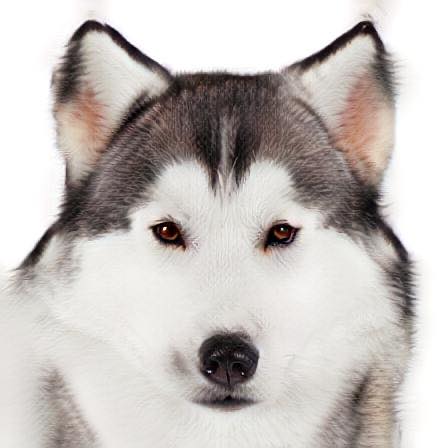}
		\\
		\\
		Source ($S_{sty}$, $S_{sem}$) & Target ($T_{sem}$) & {\bf Our Method}
		\\
		\\
		\includegraphics[width=0.300\linewidth]{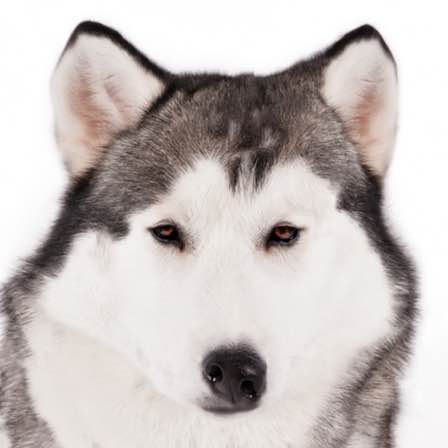}&
		\includegraphics[width=0.300\linewidth]{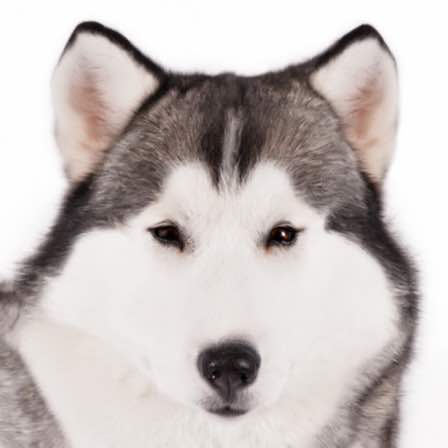}&
		\includegraphics[width=0.300\linewidth]{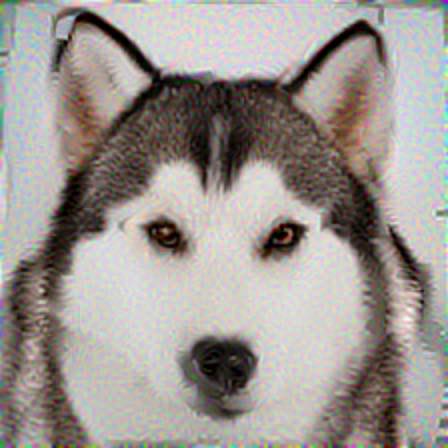}
		\\
		T-Effect & CFITT & Neural Doodle
		\\
		\includegraphics[width=0.300\linewidth]{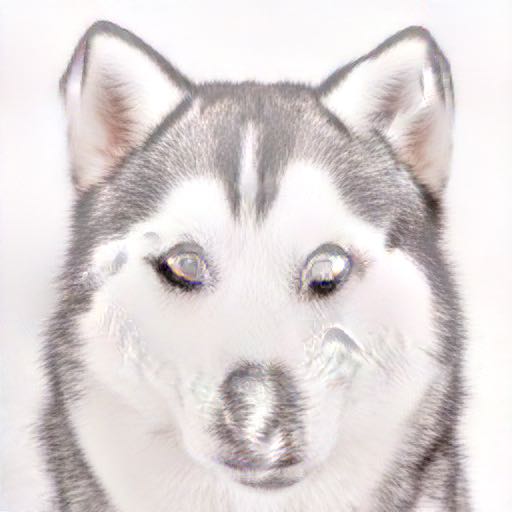}&
		\includegraphics[width=0.300\linewidth]{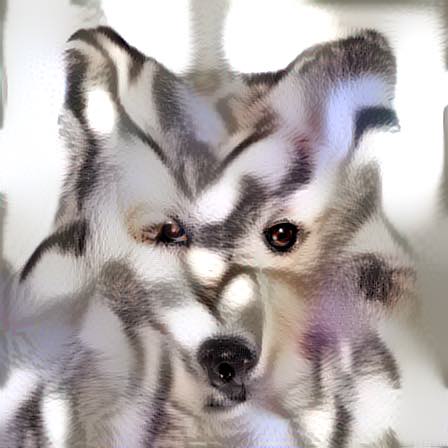}&
		\includegraphics[width=0.300\linewidth]{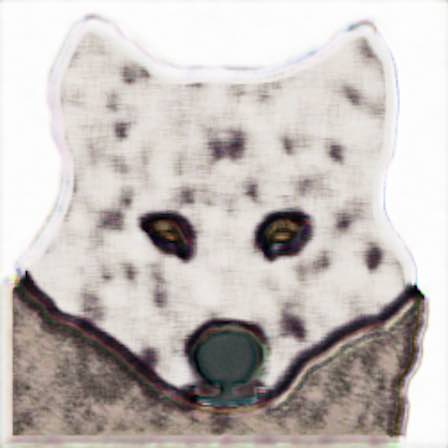}
		\\
		STROTSS & Gatys2017 & TuiGAN

	\end{tabular}
	\vspace{1em}
	\caption{Comparison with state-of-the-art methods on the doodles-to-artworks task.
	}
	\label{fig:cmp1}
\end{figure*}

\begin{figure*}[t]
	\centering
	\setlength{\tabcolsep}{0.2cm}
	\begin{tabular}{ccc}
		\includegraphics[width=0.300\linewidth]{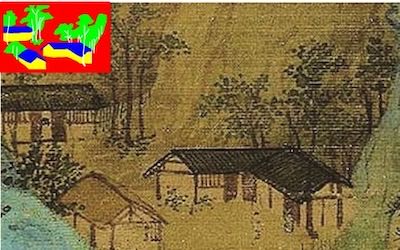}&
		\includegraphics[width=0.300\linewidth]{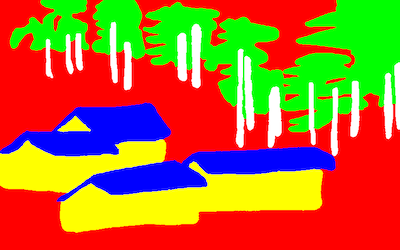}&
		\includegraphics[width=0.300\linewidth]{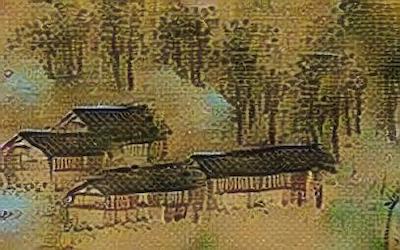}
		\\
		\\
		Source ($S_{sty}$, $S_{sem}$) & Target ($T_{sem}$) & {\bf Our Method}
		\\
		\\
		\includegraphics[width=0.300\linewidth]{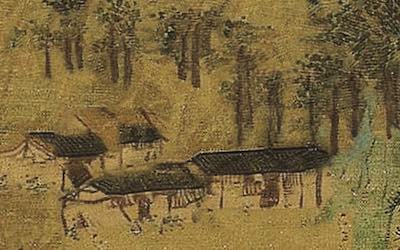}&
		\includegraphics[width=0.300\linewidth]{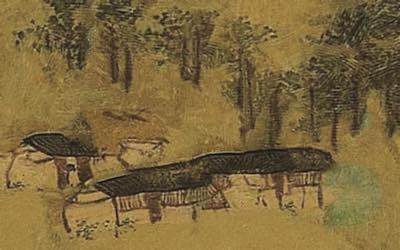}&
		\includegraphics[width=0.300\linewidth]{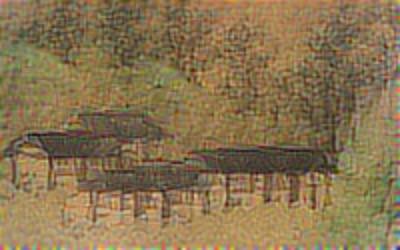}
		\\
		T-Effect & CFITT & Neural Doodle
		\\
		\includegraphics[width=0.300\linewidth]{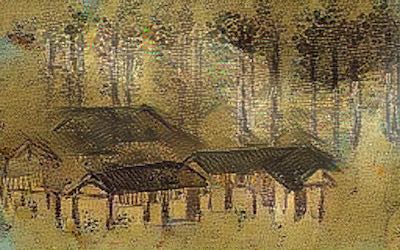}&
		\includegraphics[width=0.300\linewidth]{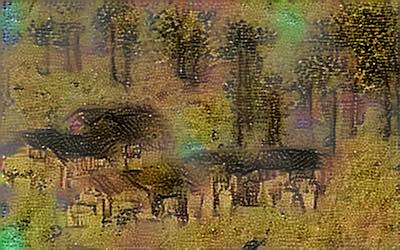}&
		\includegraphics[width=0.300\linewidth]{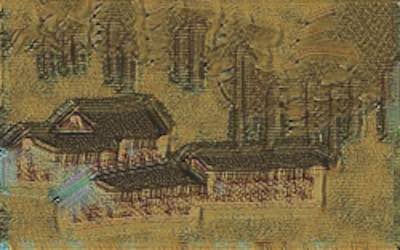}
		\\
		STROTSS & Gatys2017 & TuiGAN

	\end{tabular}
	\vspace{1em}
	\caption{Comparison with state-of-the-art methods on the doodles-to-artworks task.
	}
	\label{fig:cmp2}
\end{figure*}

\begin{figure*}[t]
	\centering
	\setlength{\tabcolsep}{0.2cm}
	\begin{tabular}{ccc}
		\includegraphics[width=0.300\linewidth]{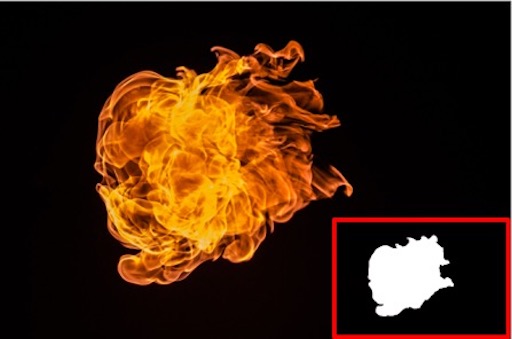}&
		\includegraphics[width=0.300\linewidth]{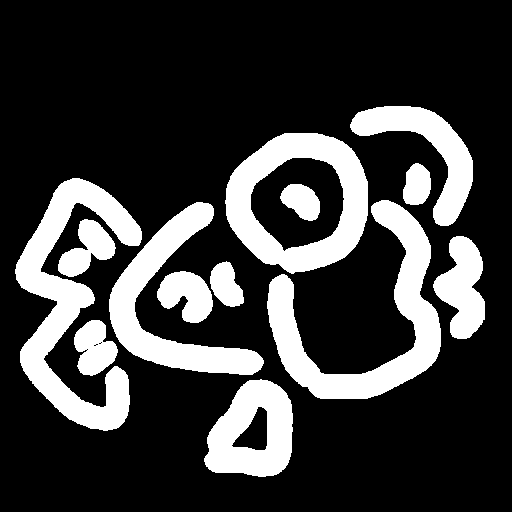}&
		\includegraphics[width=0.300\linewidth]{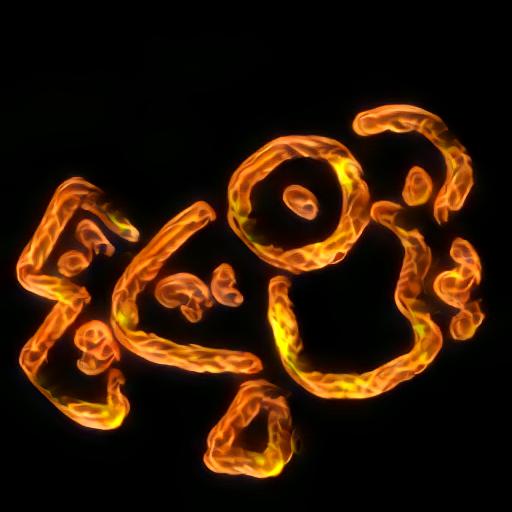}
		\\
		\\
		Source ($S_{sty}$, $S_{sem}$) & Target ($T_{sem}$) & {\bf Our Method}
		\\
		\\
		\includegraphics[width=0.300\linewidth]{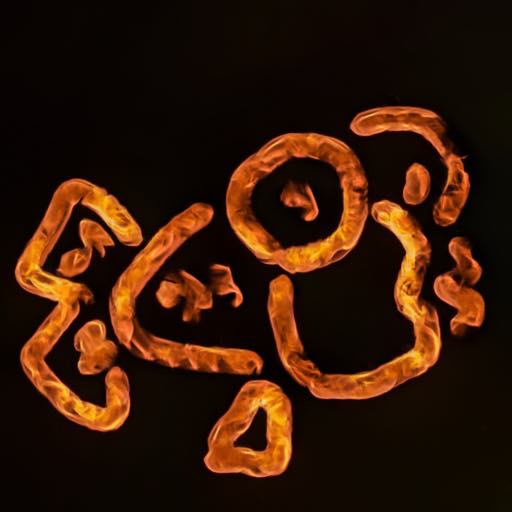}&
		\includegraphics[width=0.300\linewidth]{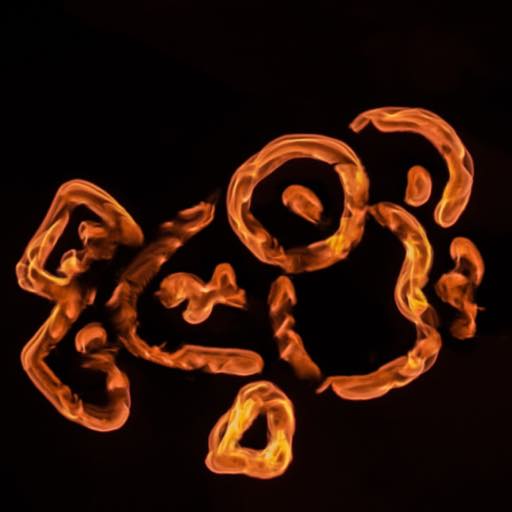}&
		\includegraphics[width=0.300\linewidth]{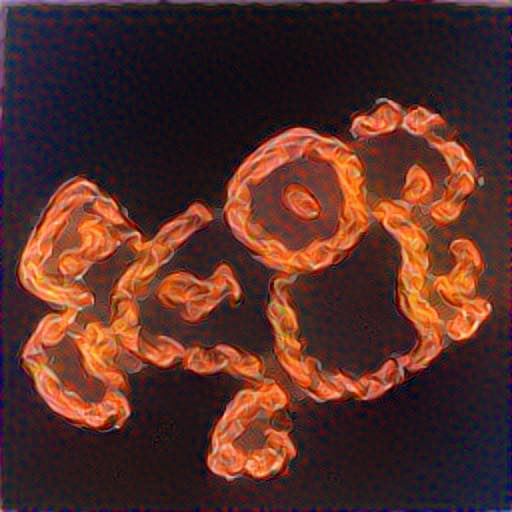}
		\\
		T-Effect & CFITT & Neural Doodle
		\\
		\includegraphics[width=0.300\linewidth]{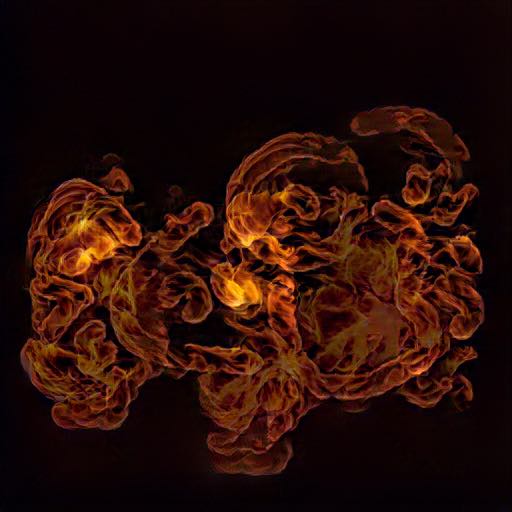}&
		\includegraphics[width=0.300\linewidth]{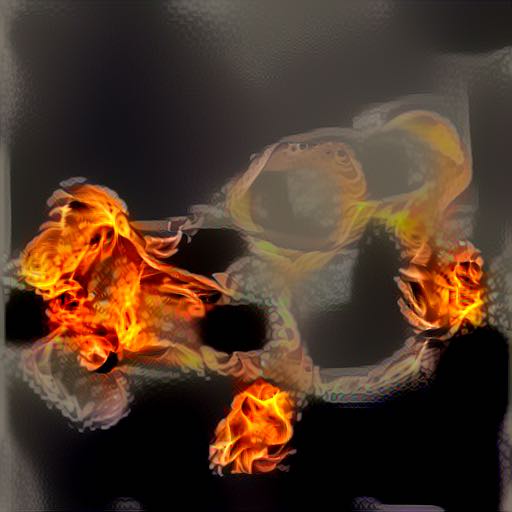}&
		\includegraphics[width=0.300\linewidth]{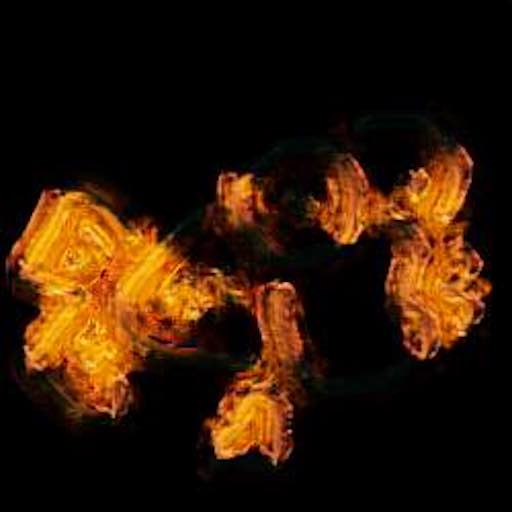}
		\\
		STROTSS & Gatys2017 & TuiGAN

	\end{tabular}
	\vspace{1em}
	\caption{Comparison with state-of-the-art methods on the texture pattern editing task.
	}
	\label{fig:cmp3}
\end{figure*}

\begin{figure*}[t]
	\centering
	\setlength{\tabcolsep}{0.2cm}
	\begin{tabular}{ccc}
		\includegraphics[width=0.300\linewidth]{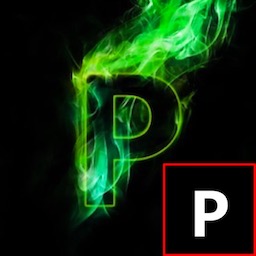}&
		\includegraphics[width=0.300\linewidth]{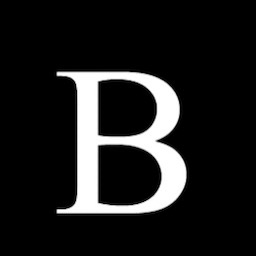}&
		\includegraphics[width=0.300\linewidth]{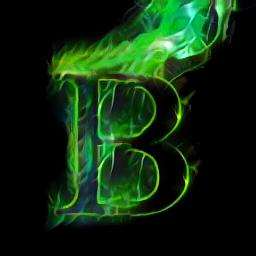}
		\\
		\\
		Source ($S_{sty}$, $S_{sem}$) & Target ($T_{sem}$) & {\bf Our Method}
		\\
		\\
		\includegraphics[width=0.300\linewidth]{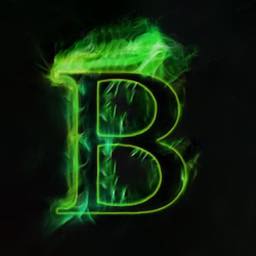}&
		\includegraphics[width=0.300\linewidth]{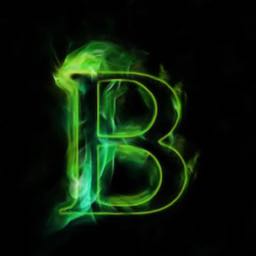}&
		\includegraphics[width=0.300\linewidth]{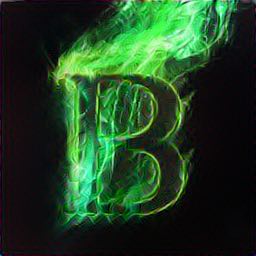}
		\\
		T-Effect & CFITT & Neural Doodle
		\\
		\includegraphics[width=0.300\linewidth]{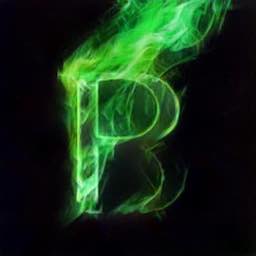}&
		\includegraphics[width=0.300\linewidth]{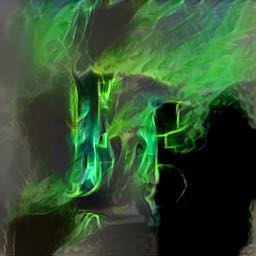}&
		\includegraphics[width=0.300\linewidth]{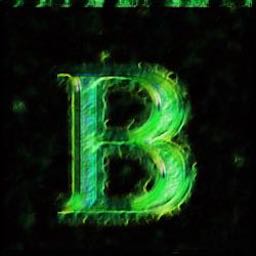}
		\\
		STROTSS & Gatys2017 & TuiGAN

	\end{tabular}
	\vspace{1em}
	\caption{Comparison with state-of-the-art methods on the text effects transfer task.
	}
	\label{fig:cmp4}
\end{figure*}

\begin{figure*}[t]
	\centering
	\setlength{\tabcolsep}{0.2cm}
	\begin{tabular}{ccc}
		\includegraphics[width=0.300\linewidth]{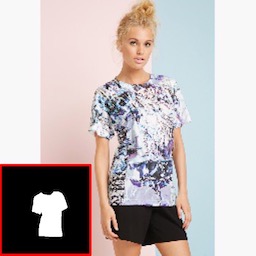}&
		\includegraphics[width=0.300\linewidth]{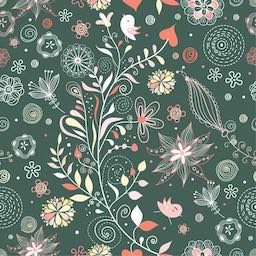}&
		\includegraphics[width=0.300\linewidth]{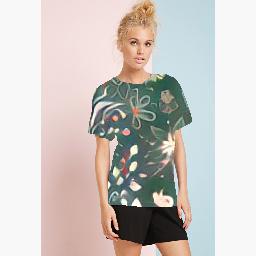}
		\\
		\\
		Manipulated image \& $T_{sem}$ & Source stylized image $S_{sty}$ & {\bf Our Method}
		\\
		\\
		\includegraphics[width=0.300\linewidth]{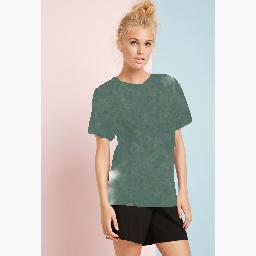}&
		\includegraphics[width=0.300\linewidth]{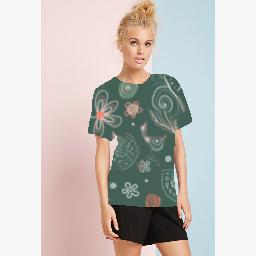}&
		\includegraphics[width=0.300\linewidth]{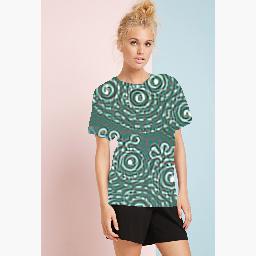}
		\\
		T-Effect & CFITT & Neural Doodle
		\\
		\includegraphics[width=0.300\linewidth]{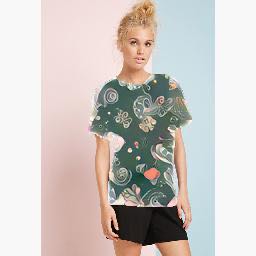}&
		\includegraphics[width=0.300\linewidth]{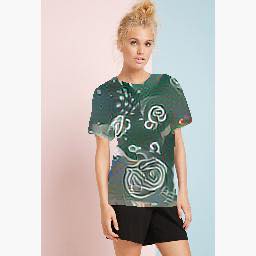}&
		\includegraphics[width=0.300\linewidth]{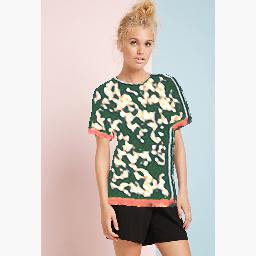}
		\\
		STROTSS & Gatys2017  & TuiGAN

	\end{tabular}
	\vspace{1em}
	\caption{Comparison with state-of-the-art methods on the virtual clothing manipulation task.
	}
	\label{fig:cmp5}
\end{figure*}

We provide additional comparison results of our method and the state-of-the-art universal interactive texture transfer algorithms (T-Effect~\cite{yang2017awesome}, CFITT~\cite{men2018common}, Neural Doodle~\cite{champandard2016semantic}, STROTSS~\cite{kolkin2019style}, Gatys2017~\cite{gatys2017controlling}, and TuiGAN~\cite{lin2020tuigan}) in different scenarios. As shown in Fig.~\ref{fig:cmp1},~\ref{fig:cmp2},~\ref{fig:cmp3},~\ref{fig:cmp4},~\ref{fig:cmp5}, our approach is capable of synthesizing higher quality results with better-preserved structures and more vivid stylization effects.

\subsection{Comparison results with baseline methods}
\label{sec4}
\begin{figure*}[h]
	\centering
	\setlength{\tabcolsep}{0.03cm}
	\begin{tabular}{ccccc}
		\includegraphics[width=0.195\linewidth]{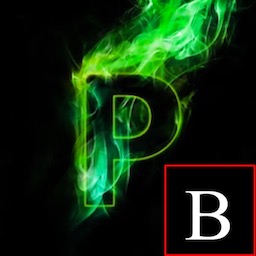}&
		\includegraphics[width=0.195\linewidth]{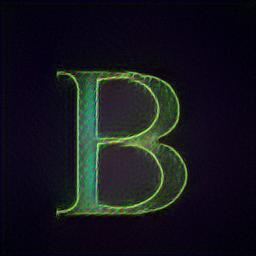}&
		\includegraphics[width=0.195\linewidth]{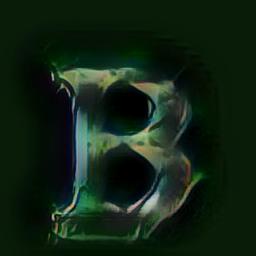}&
		\includegraphics[width=0.195\linewidth]{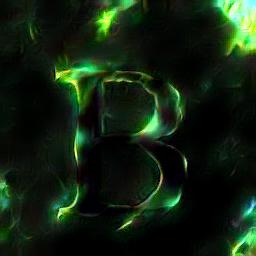}&
		\includegraphics[width=0.195\linewidth]{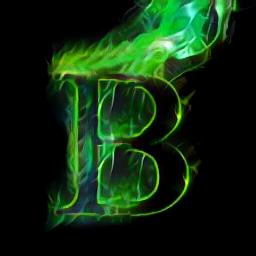}
		\\
		
		\includegraphics[width=0.195\linewidth]{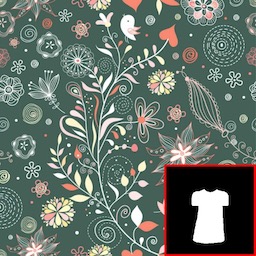}&
		\includegraphics[width=0.195\linewidth]{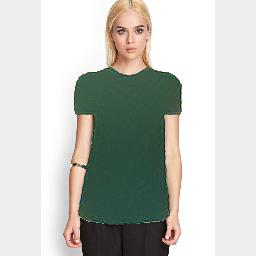}&
		\includegraphics[width=0.195\linewidth]{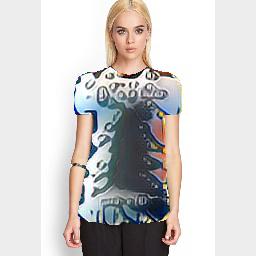}&
		\includegraphics[width=0.195\linewidth]{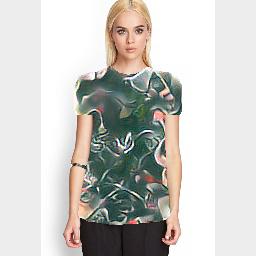}&
		\includegraphics[width=0.195\linewidth]{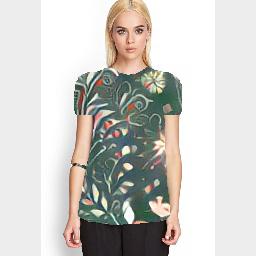}
		\\
		
		($S_{sty}$, $T_{sem}$)& Style-Swap & AdaIN & WCT& {\bf Ours} 
		\\
		
	\end{tabular}
	\vspace{1em} 
	\caption{Comparison results with baseline methods (Style-Swap~\cite{chen2016fast}, AdaIN~\cite{huang2017arbitrary}, and WCT~\cite{li2017universal}). For fairness, all baselines are tailored to be interactive by introducing semantic maps. As is evident, Style-Swap and AdaIN can only produce less stylized results due to the single-level stylization (see the top row). By exploiting a multi-level mechanism, WCT and ours achieve much more vivid stylization. However, since WCT is designed to match the global statistics (i.e., covariance), it cannot transfer local textures and often introduces distorted patterns. In addition, the three baselines all fail to preserve the local texture structures (e.g., the patterns in the bottom row), while ours not only retains the complete local patterns but also transfers the vivid global texture effects, thanks to the proposed view-specific texture reformation (VSTR) operation and multi-view and multi-stage synthesis procedure (see more analyses in our main paper).
	}
	\label{fig:baselinecmp}
\end{figure*}

The main innovation of our work lies in the fast and universal interactive texture transfer framework. Since some of its components are based on Style-Swap~\cite{chen2016fast}, AdaIN~\cite{huang2017arbitrary}, and WCT~\cite{li2017universal}, it would be helpful to compare against them to validate the superiority of our framework. The comparison results are shown in Fig.~\ref{fig:baselinecmp}, and the analyses are provided in the caption.

\subsection{Limitations and discussions}
\label{sec5}

\begin{figure*}[h]
	\centering
	\setlength{\tabcolsep}{0.05cm}
	\begin{tabular}{ccp{0.01cm}|p{0.01cm}ccp{0.01cm}|p{0.01cm}cc}
		Input& Output &&& Input& Output  &&& Input& Output 
		\\
		\includegraphics[width=0.15\linewidth]{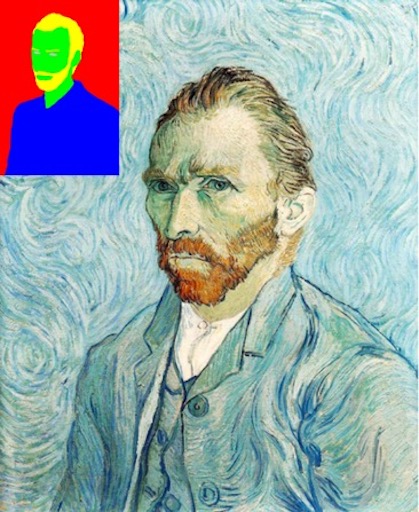}&
		\includegraphics[width=0.126\linewidth]{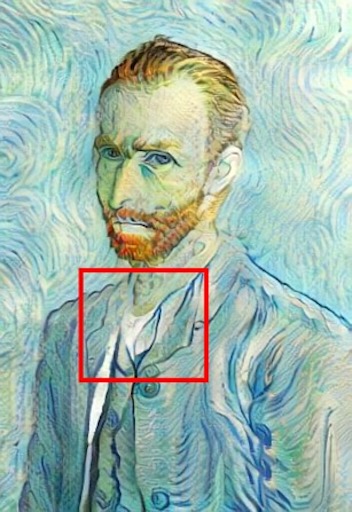}&&&
		\includegraphics[width=0.18\linewidth]{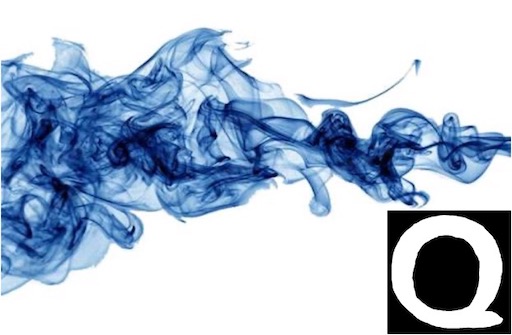}&
		\includegraphics[width=0.15\linewidth]{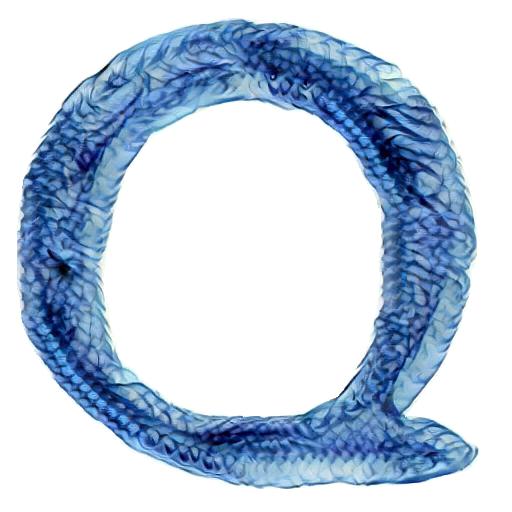}&&&
		\includegraphics[width=0.15\linewidth]{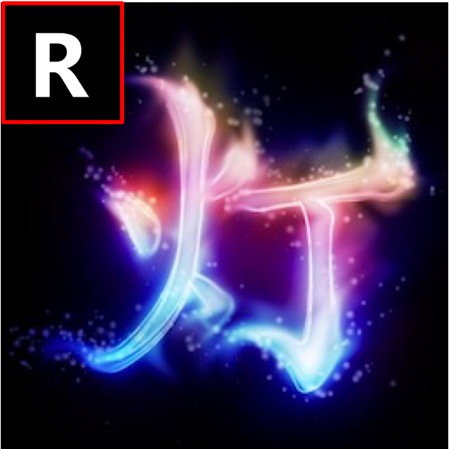}&
		\includegraphics[width=0.15\linewidth]{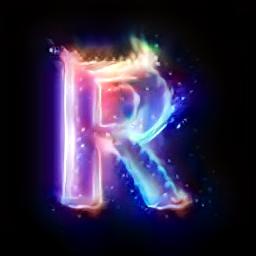}
		\\
		\\
		($S_{sty}$, $T_{sem}$) & (a) artifacts &&&
		($S_{sty}$, $T_{sem}$) & (b) repeating patterns &&&
		($S_{sty}$, $T_{sem}$) & (c) color blending
		\\
		\\
	\end{tabular}
	\caption{ {\bf Limitation of type 1: quality issues.} Some of the results generated by our method may suffer from a few quality issues, e.g., (a) noticeable artifacts, (b) smaller repeating patterns, and (c) color blending. These issues may be caused by the inherent nature of patch alignment in the VSTR we used in local refinement stage II (we removed the global alignment stage I, but these effects were not mitigated). As a well-known problem, the patch alignment is prone to introduce wash-out artifacts due to the repetitive use of the same patches~\cite{gu2018arbitrary,sheng2018avatar}, and sometimes may cause color blending problem due to the averaging of overlapping patches~\cite{chen2016fast}. Nevertheless, we still utilize it in our framework due to its flexibility and efficiency. The further improvement can be achieved by improving the VSTR operation, or integrating additional guidances like~\cite{men2018common}, or more straightforwardly, specifying more elaborate semantic maps.
	}
	\label{fig:limit1}
\end{figure*}

\begin{figure*}[h]
	\centering
	\setlength{\tabcolsep}{0.05cm}
	\begin{tabular}{ccp{0.01cm}|p{0.01cm}ccccc}
		Input & Input &&& \multicolumn{5}{c}{Outputs}
		\\
		\includegraphics[width=0.13\linewidth]{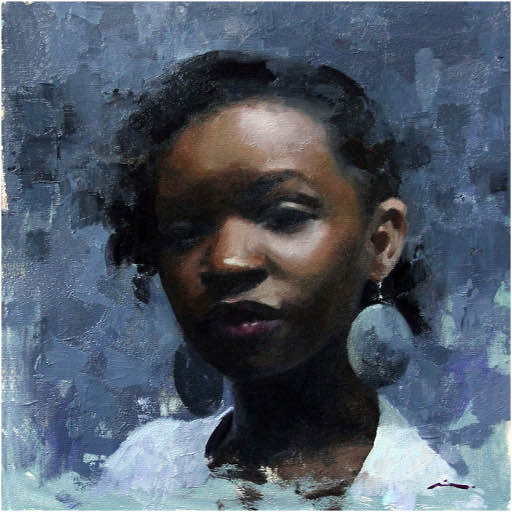}&
		\includegraphics[width=0.13\linewidth]{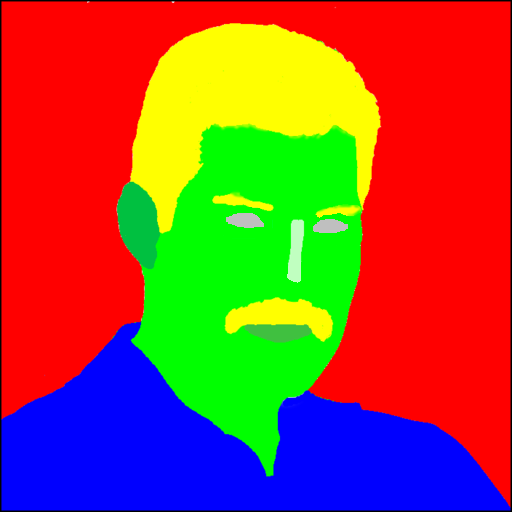}&&&
		\includegraphics[width=0.13\linewidth]{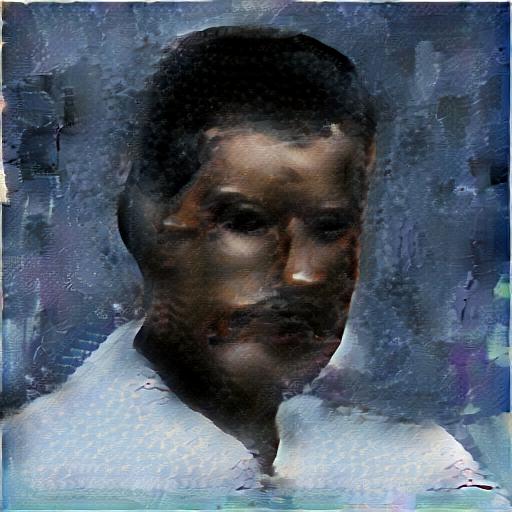}&
		\includegraphics[width=0.13\linewidth]{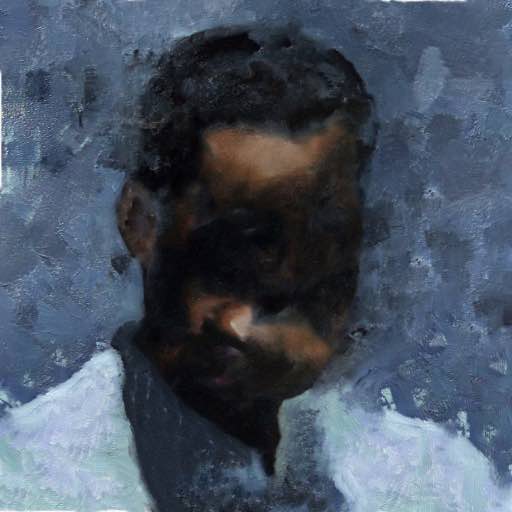}&
		\includegraphics[width=0.13\linewidth]{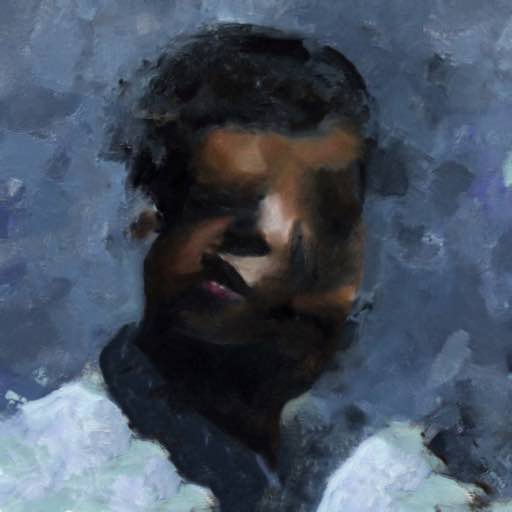}&
		\includegraphics[width=0.13\linewidth]{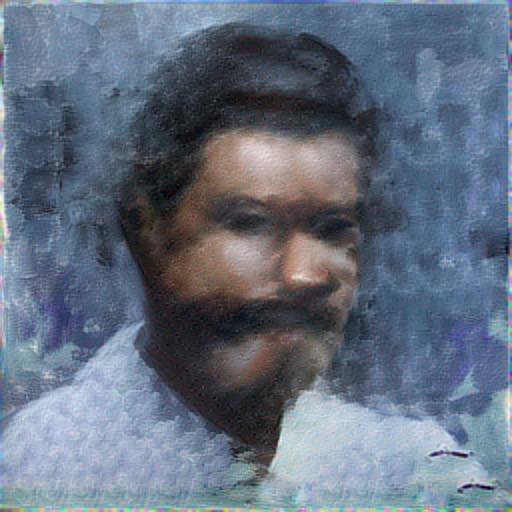}&
		\includegraphics[width=0.13\linewidth]{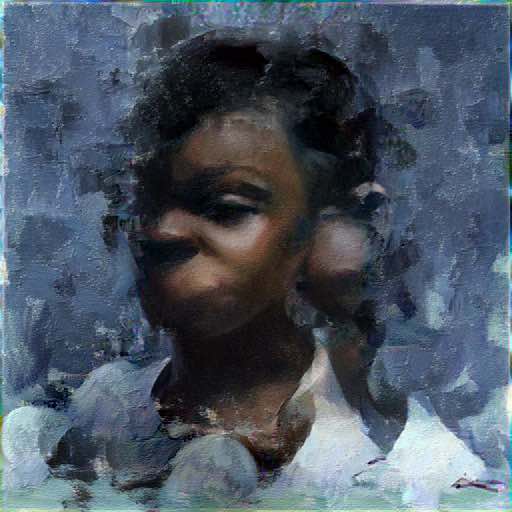}
		\\
		\includegraphics[width=0.13\linewidth]{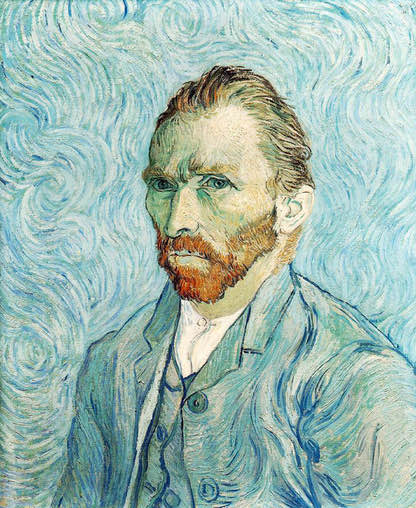}&
		\includegraphics[width=0.108\linewidth]{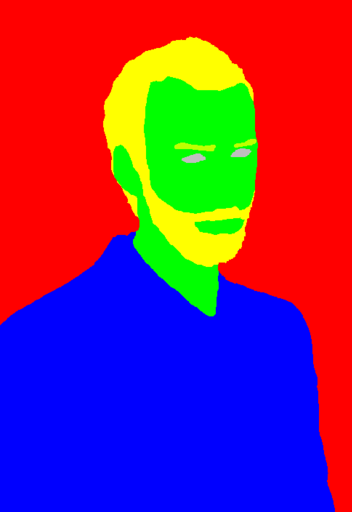}&&&
		\includegraphics[width=0.108\linewidth]{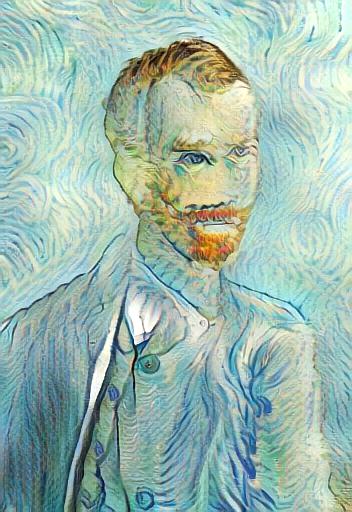}&
		\includegraphics[width=0.108\linewidth]{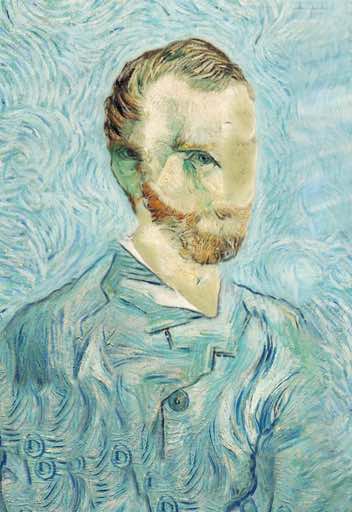}&
		\includegraphics[width=0.108\linewidth]{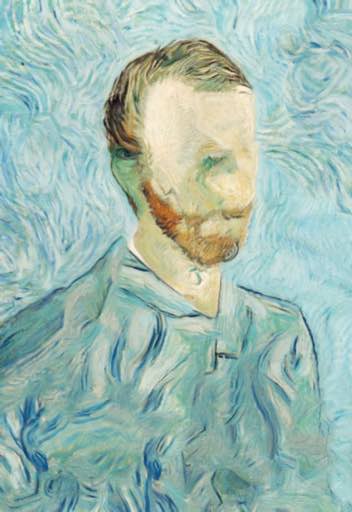}&
		\includegraphics[width=0.108\linewidth]{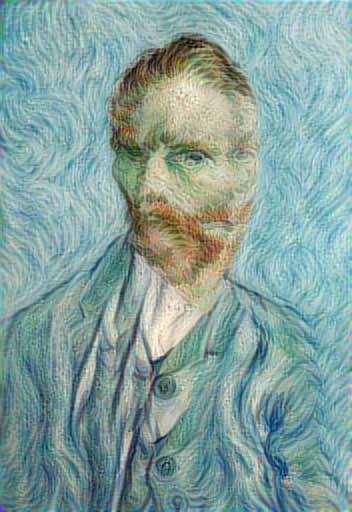}&
		\includegraphics[width=0.108\linewidth]{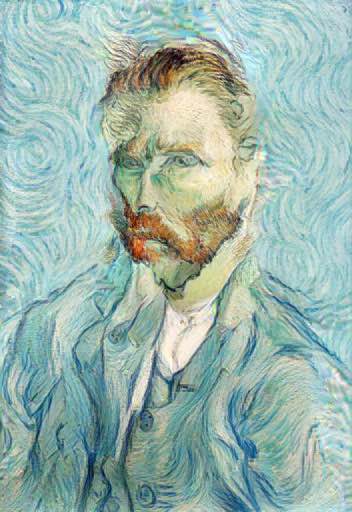}
		\\
		\\
		$S_{sty}$  & $T_{sem}$ &&& {\bf Ours} & \small T-Effect & \small CFITT & \small Neural Doodle & \small STROTSS
		\\
		\\
		
	\end{tabular}
	\caption{ {\bf Limitation of type 2: semantic maps with drastically different shapes.} Since our framework does not make patches rotated or scaled, it may not work very well for semantic maps with drastically different shapes (e.g., flipped poses). Nonetheless, as we can see, our method still outperforms SOTA in this situation. The issues may be solved by allowing patches to be rotated and scaled in our framework.
	}
	\label{fig:limit2}
\end{figure*}

\begin{figure*}[h]
	\centering
	\setlength{\tabcolsep}{0.1cm}
	\begin{tabular}{cccc}
		Input & Input & Input & Output
		\\
		\includegraphics[width=0.225\linewidth]{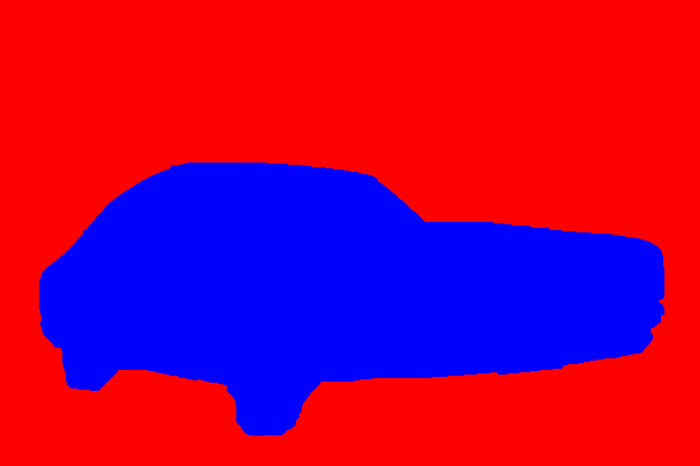}&
		\includegraphics[width=0.225\linewidth]{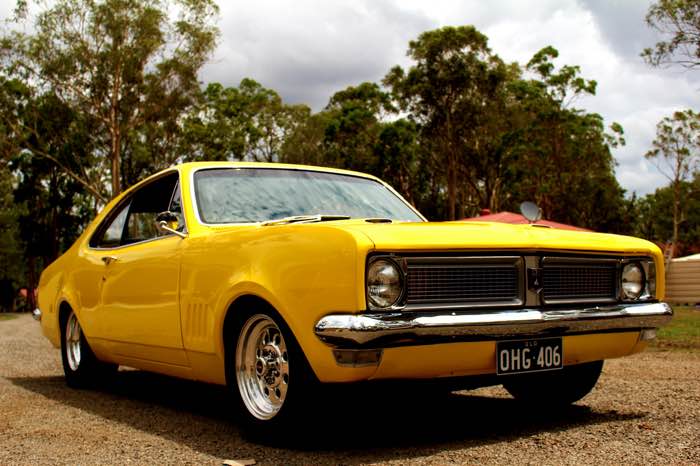}&
		\includegraphics[width=0.225\linewidth]{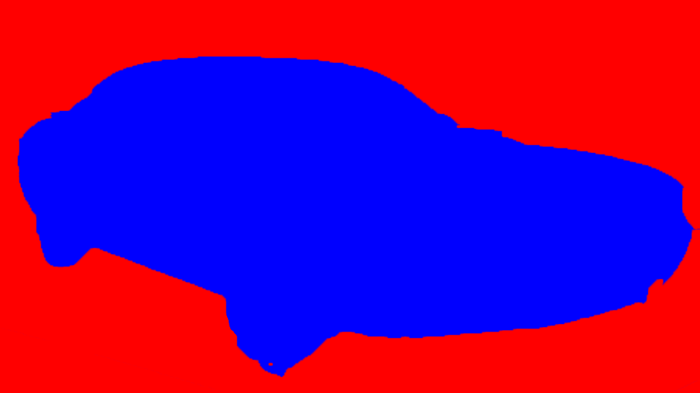}&
		\includegraphics[width=0.225\linewidth]{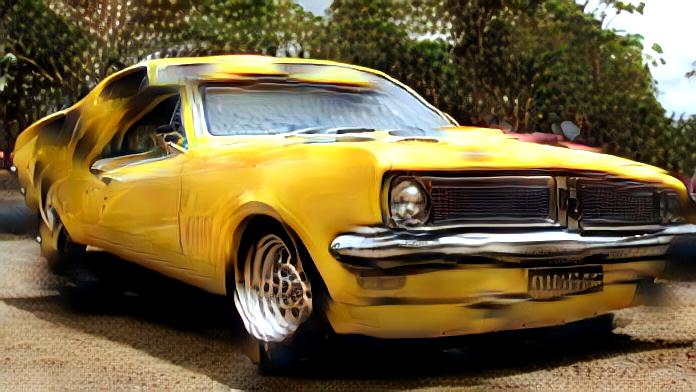}
		\\
		\\
		$S_{sem}$  & $S_{sty}$ &  $T_{sem}$ & $T_{sty}$ 
		\\
		\\

	\end{tabular}
	\caption{ {\bf Limitation of type 3: geometric texture transfer.} Our method cannot produce geometric texture transfer for images with strict structure requirements (e.g., the structures of the car shell and tires), because our VSTR does not allow patches to be either rotated or scaled. This issue may be addressed by allowing patches to be rotated and scaled in our framework, or using some geometric transformations like thin-plate spline interpolation (TPS) in the pipeline. Moreover, one may also model a deformation field like~\cite{liu2004near}, or borrow some ideas from recent geometric style transfer approaches~\cite{kim2020deformable,liu2021learning} to achieve geometric texture transfer.
	}
	\label{fig:limit3}
\end{figure*}

\begin{figure*}[h]
	\centering
	\setlength{\tabcolsep}{0.1cm}
	\begin{tabular}{cccc}
		Input & Input & Input & Output
		\\
		\includegraphics[width=0.225\linewidth]{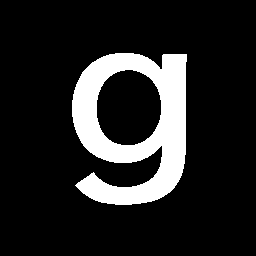}&
		\includegraphics[width=0.225\linewidth]{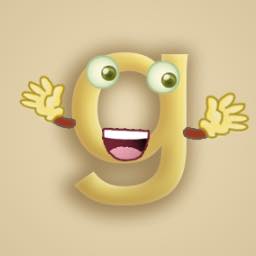}&
		\includegraphics[width=0.225\linewidth]{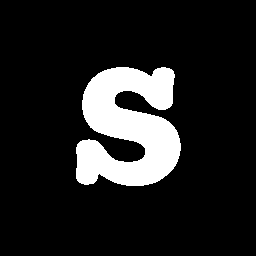}&
		\includegraphics[width=0.225\linewidth]{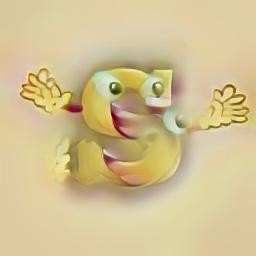}
		\\
		\includegraphics[width=0.225\linewidth]{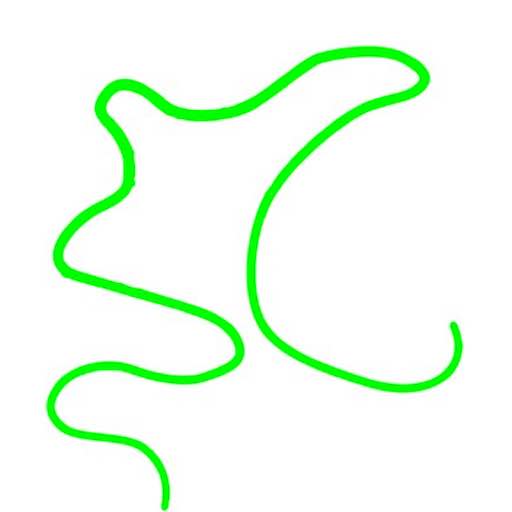}&
		\includegraphics[width=0.225\linewidth]{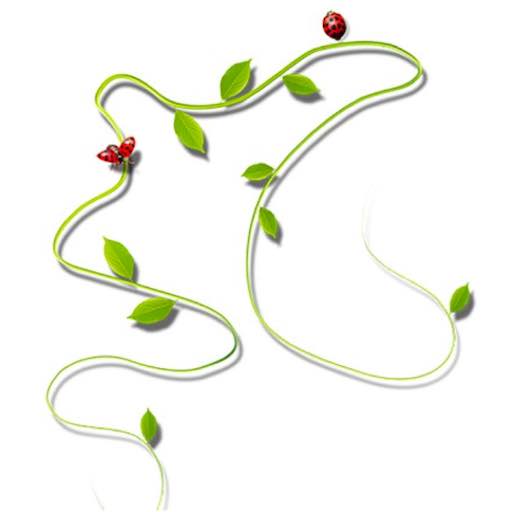}&
		\includegraphics[width=0.225\linewidth]{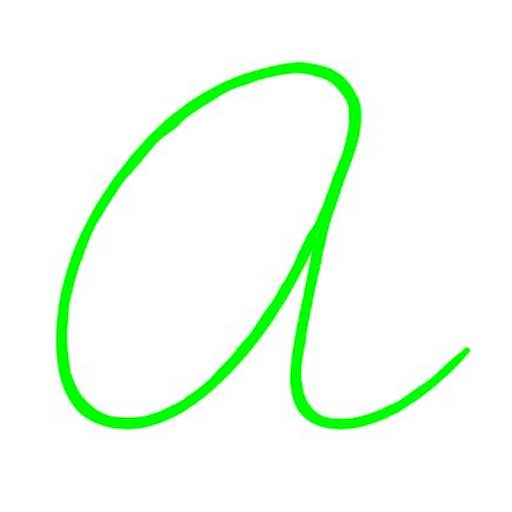}&
		\includegraphics[width=0.225\linewidth]{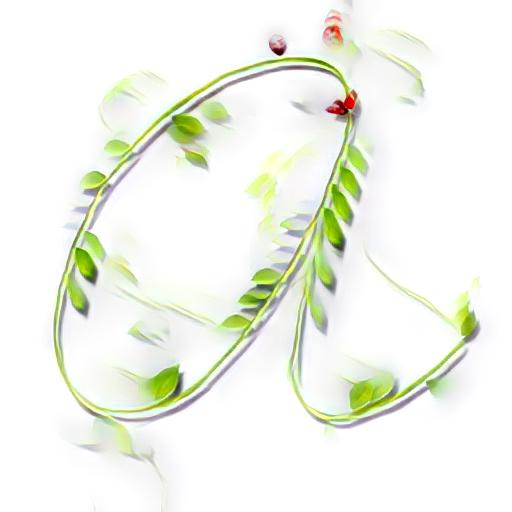}
		\\
		\\
		$S_{sem}$  & $S_{sty}$ & $T_{sem}$ & $T_{sty}$ 
		\\
		\\

	\end{tabular}
	\caption{ {\bf Limitation of type 4: patterns with decorative elements.} Since our method does not use any additional structure guidance or distribution constraint, it may fail to achieve correct texture transfer for patterns with decorative elements (e.g., the eyes, mouth, and hands in the top row) when given extremely simple semantic maps. This may be solved by integrating additional guidances or distribution constraints like~\cite{men2018common} and~\cite{yang2017awesome}, or simply specifying more elaborate semantic maps.
	}
	\label{fig:limit4}
\end{figure*}

In this work, since our focus is a novel fast and universal solution for interactive texture transfer, {\em we prioritize flexibility and efficiency over quality}. Although our results have shown improvement over existing works, and the user study has also verified our superiority against SOTA on quality, the results are imperfect, and there are still some limitations. Given the simplicity of our proposed framework, we believe there is substantial room for improvement. In Fig.~\ref{fig:limit1},~\ref{fig:limit2},~\ref{fig:limit3},~\ref{fig:limit4}, we provide some typical limitations of our method, then analyze the reasons behind them and discuss the possible solutions to address them. The further improvement of our framework we leave as future work.

\subsection{How to balance the performance of different stages?}
In our implementation code (\url{https://github.com/EndyWon/Texture-Reformer}), we provide some interfaces to help users to balance the performance of each stage. Users can control the performance by varying the patch size, changing the stage separation, adjusting the semantic weight, or interpolating between features (like the ways applied in~\cite{huang2017arbitrary} and~\cite{li2017universal}). There is no optimal setting that can perfectly deal with all cases. Our default setting provides a general solution that can be applied to most cases. However, since our framework is fast, modular, and learning-free, users can easily explore the optimal settings for any specific case.

\renewcommand\arraystretch{0.6}

\renewcommand\arraystretch{0.6}

\renewcommand\arraystretch{0.6}

\renewcommand\arraystretch{0.6}

\renewcommand\arraystretch{0.6}

\renewcommand\arraystretch{0.6}

\renewcommand\arraystretch{0.6}

\renewcommand\arraystretch{0.6}

\renewcommand\arraystretch{0.6}

\renewcommand\arraystretch{0.6}

\renewcommand\arraystretch{0.6}

\renewcommand\arraystretch{0.6}

\renewcommand\arraystretch{0.6}
\clearpage

\renewcommand\arraystretch{0.6}

\renewcommand\arraystretch{0.6}

\renewcommand\arraystretch{0.6}

\renewcommand\arraystretch{0.6}

\end{document}